\documentclass[10pt,twocolumn]{article} 
\usepackage{simpleConference}
\usepackage{times}
\usepackage{graphicx}
\usepackage{amssymb}
\usepackage{subfigure}
\usepackage{multirow}
\usepackage{diagbox}
\usepackage{booktabs}

\begin{document}

\title{Construction of all-in-focus images assisted by depth sensing}

\author{Hang Liu\footnotemark[1], Hengyu Li\footnotemark[1], Jun Luo\footnotemark[1], Shaorong Xie\footnotemark[1], Yu Sun\footnotemark[2]}

\maketitle
\thispagestyle{empty}

\footnotetext[1]{Hang Liu, Hengyu Li, Jun Luo, Shaorong Xie are with the School of Mechatronic Engineering and Automation, Shanghai University, China. Hengyu Li is the corresponding author. \href{scholar.hang@gmail.com}{scholar.hang@gmail.com} \href{lihengyu@shu.edu.cn}{lihengyu@shu.edu.cn} }
\footnotetext[2]{Yu Sun is with the Department of Mechanical and Industrial Engineering, University of
	Toronto, Canada, and the Shanghai University. \href{sun@mie.utoronto.ca}{sun@mie.utoronto.ca}}

\begin{abstract}
Multi-focus image fusion is a technique for obtaining an all-in-focus image in which all objects are in focus to extend the limited depth of field (DoF) of an imaging system. Different from traditional RGB-based methods, this paper presents a new multi-focus image fusion method assisted by depth sensing. In this work, a depth sensor is used together with a color camera to capture images of a scene. A graph-based segmentation algorithm is used to segment the depth map from the depth sensor, and the segmented regions are used to guide a focus algorithm to locate in-focus image blocks from among multi-focus source images to construct the reference all-in-focus image. Five test scenes and six evaluation metrics were used to compare the proposed method and representative state-of-the-art algorithms. Experimental results quantitatively demonstrate that this method outperforms existing methods in both speed and quality (in terms of comprehensive fusion metrics). The generated images can potentially be used as reference all-in-focus images.
\end{abstract}

\section{Introduction}
The depth of field (DoF) of an imaging system is limited. With a fixed focus setting, only objects in a particular depth range appear focused in the captured source image, whereas objects in other depth ranges are defocused and blurred. An all-in-focus image in which all objects are in focus has many applications, such as digital photography \cite{Li2017Pixel}, medical imaging \cite{Motta2015All}, and microscopic imaging \cite{Nguyen2012Real, Liu2011Extended}. A number of all-in-focus imaging methods have been proposed, which can be grouped into two categories: point spread function (PSF)-based methods and RGB-based multi-focus image fusion methods.

The PSF-based methods obtain an all-in-focus image by estimating the PSF of the imaging system and restoring an all-in-focus image based on the estimated PSF. A partially focused image can be modeled as an all-in-focus image convolved with a PSF. Deconvolution methods first estimate the PSF and then deconvolve with this PSF to restore an all-in-focus image. The PSF of a partially focused image is non-uniform because the farther an object is from the DoF of an imaging system, the larger is the extent of blurriness of the object in an image. One type of deconvolution method directly estimates the non-uniform PSF of an imaging system using specially designed cameras \cite{Bishop2011The} or a camera with a specially designed lattice-focal lens \cite{Gario20094D}. Instead of estimating the non-uniform PSF, the other type of deconvolution method first constructs an image with uniform blur and then estimates a uniform PSF. The image with uniform blur can be obtained by scanning the focus positions \cite{Liu2011Extended, Iwai2015Extended} or moving lens or image detector \cite{kuthirummal2011flexible} during a single detector exposure. The wave-front coding technique is another approach to obtain a uniform blur image by adding a suitable phase mask to the aperture plane and making the optical transfer function of the imaging system defocus invariant \cite{Dowski1995Extended, Yang2007Optimized, Zhao2008Optimized, Cossairt2010Diffusion}. The deconvolution methods enable single-shot extended DoF imaging. However, deconvolution ringing artifacts can appear in the resulting image, and high frequencies can be captured with lower fidelity \cite{kuthirummal2011flexible}.

In RGB-based multi-focus image fusion methods, in-focus image blocks are distinguished from among multiple multi-focus source images that are captured using different focus settings, to construct an all-in-focus image. Existing multi-focus image fusion algorithms include multi-scale transform \cite{Zhao2013Multi, Li2011Performance}, feature space transform \cite{Wan2013Multifocus, Liang2012Image}, spatial domain methods \cite{Li2013ImageFusion, Li2013ImageMatting, Mckee1996Everywhere, Ohba2003Microscopic, Motta2015All}, pulse coupled neural network \cite{Huang2007Multi, Xiao2008Image}, and deep convolutional neural network \cite{Liu2017Multi}.

In multi-focus image fusion, one challenge is to obtain a reference all-in-focus image, which better reflects the ground truth, for other methods to directly compared to. Due to the lack of reference images, a number of metrics were defined for indirectly comparing the performance across multi-focus image fusion methods. As discussed in \cite{Liu2015Multi}, various metrics, such as information theory-based metrics, image feature-based metrics, image structural similarity-based metrics, and human perception-based metrics \cite{Liu2012Objective} were developed because they all represent different aspects of the quality of an all-in-focus image.

In order to obtain a reference all-in-focus image, if the distances between all objects and the camera are known, the in-focus image blocks can be directly determined by choosing those objects whose distances are within the DoF of the camera. This is enabled by the advent and rapid advances of depth sensors (e.g., Microsoft Kinect and ZED stereo camera) which provide a convenient approach for accurately determining the distances of objects in a scene.

In this paper, a fast multi-focus image fusion method assisted by depth sensing is reported. Instead of distinguishing in-focus image blocks from among multi-focus source images, a graph-based depth map segmentation algorithm is proposed to directly obtain in-focus image block regions by segmenting the depth map. The distances of objects in each segmented in-focus image block region are confined to be within the DoF of the camera such that all objects in the region appear focused in a multi-focus source image. These regions are used to guide the focus algorithm to locate an in-focus image for each region from among multi-focus source images to construct an all-in-focus image. Experimental results quantitatively demonstrate that this method outperforms existing methods in both speed and quality (in terms of fusion metrics); thus, the generated images can potentially be used as reference all-in-focus images. The proposed method is not dependent on a specific depth sensor and can be implemented with structured light-based depth sensors (e.g., Microsoft Kinect v1), time of flight-based depth sensors (e.g., Microsoft Kinect v2), stereo cameras (e.g., ZED stereo camera), and laser scanners.

\section{Multi-focus image fusion system}
\label{sec2}
\begin{figure}[!t]
	\centering
	\includegraphics[width=0.35\textwidth]{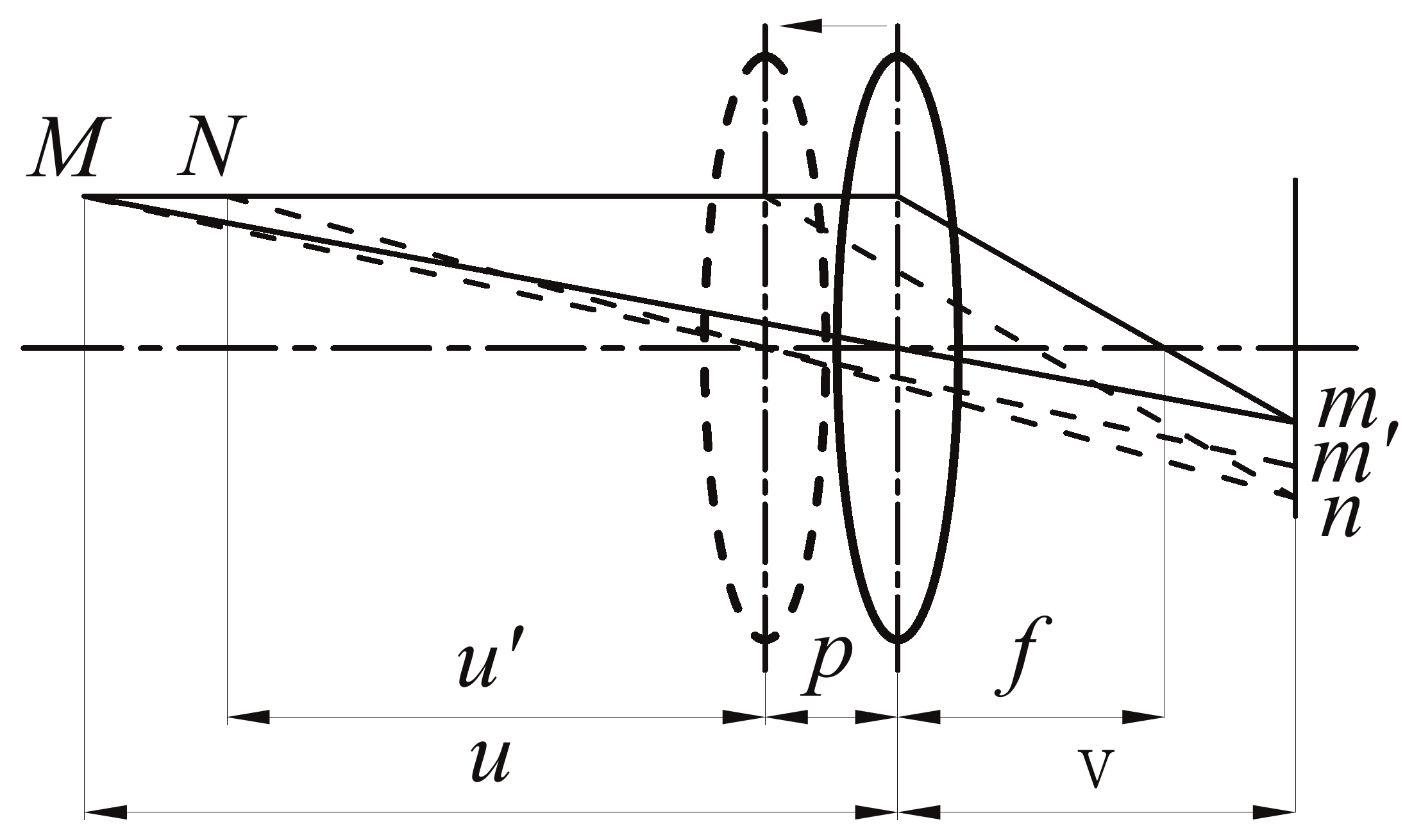}
	\caption{A scene point $M$ at a distance of $u$ from the lens is imaged in focus by an image detector at a distance of v from the lens with a focal length $f$. If the lens moves forward with a distance of $p$, $M$ is imaged as a blurred circle around $m^\prime$, while the near scene point $N$ at a distance of $u^\prime$ from the lens is imaged in focus at $n$.}
	\label{Fig1}
\end{figure}

In Fig. \ref{Fig1}, the image detector is at a distance of $v$ from a lens with focal length $f$. A scene point $M$, at a distance of $u$ from the lens, is imaged in focus at $m$. If the lens moves forward with a distance of $p$ from the lens, then $M$ is imaged as a blurred circle centered around $m^{\prime}$, while the scene point $N$ at a distance of $u^{\prime}~(u^{\prime} < u)$ from the lens is imaged in focus at $n$. In optics, if the distance between $m$ and $m^{\prime}$ is less than the radius of the circle of confusion (CoC) in the image plane, all the scene points between $M$ and $N$ appear acceptably sharp in the image. This indicates that by changing the distance between the lens and image detector while capturing images, objects at different distance ranges appear focused in order in the captured multi-focus source images. DoF can be divided into back DoF (denoted by $b\_DoF$ in this work) and front DoF (denoted by $f\_DoF$ in this work), which indicate the depth range of objects after and before the precisely in-focus scene point that can appear acceptably sharp in an image.

\begin{figure*}[!t]
	\centering
	\subfigure[]{\label{Fig2.1} \includegraphics[width=0.26\textwidth]{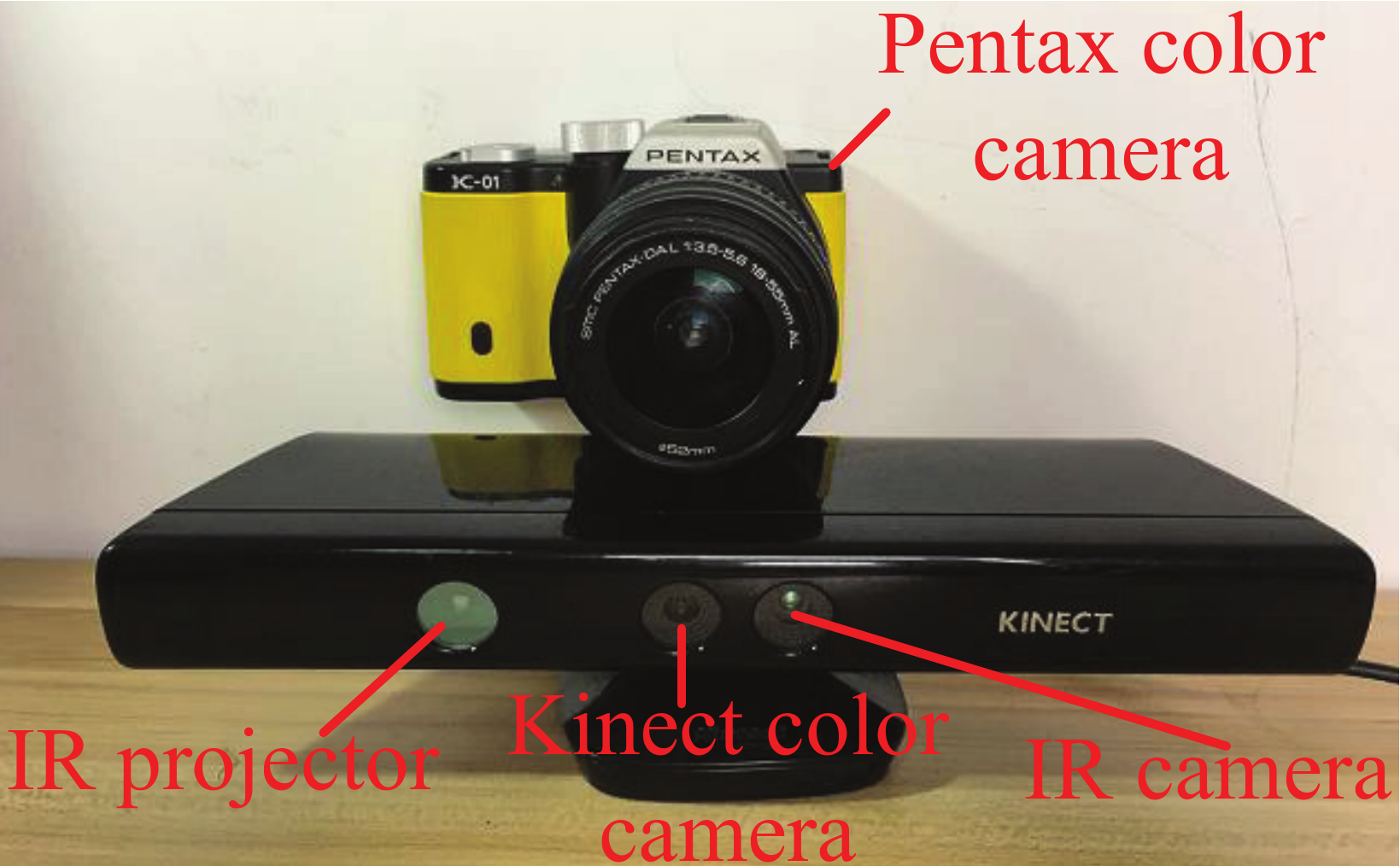}}
	\subfigure[]{\label{Fig2.2} \includegraphics[width=0.72\textwidth]{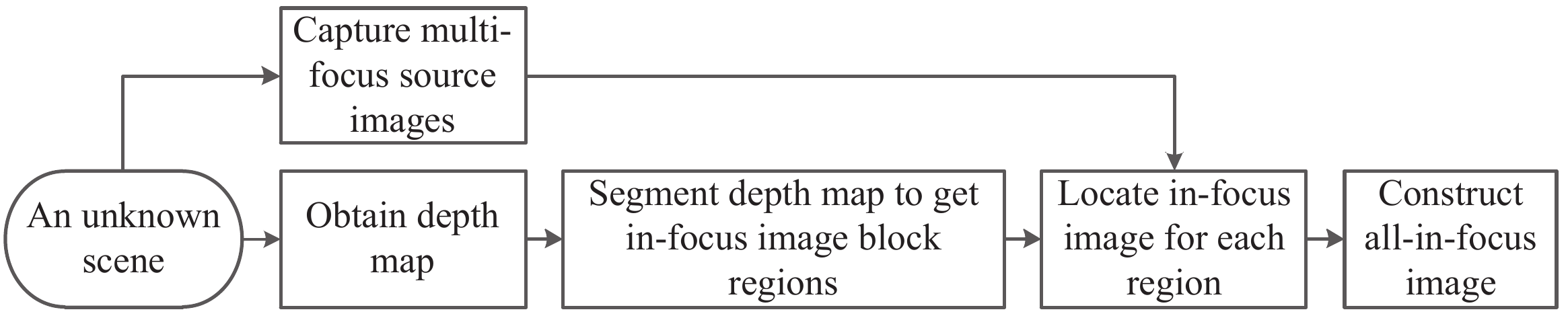}}
	\caption{(a) Setup used in this work for evaluating the proposed multi-focus image fusion method. (b) Flow chart of the method.}
	\label{Fig2}
\end{figure*}

Fig. \ref{Fig2.1} shows our multi-focus image fusion system, which consists of a focus-tunable Pentax K01 color camera with an $18-55~mm$ lens and a Kinect depth sensor. The diameter of the CoC of the color camera $\delta$ is $0.019~mm$, the aperture value $F$ was set to $4.0$, and the focal length $f$ was set to $24~mm$. The flow chart of the proposed multi-focus image fusion method is shown in Fig. \ref{Fig2.2}. In this method, the depth map and multi-focus source images of an unknown scene are captured using the Kinect depth sensor and Pentax color camera, respectively. Then, the depth map is segmented into multiple in-focus image block regions, and the objects in each region are within a DoF and all appear focused. These segmented in-focus image block regions are used to guide the focus algorithm to locate an in-focus image from among multi-focus source images for each region. Finally, the all-in-focus image is constructed by combining the in-focus images of all segmented regions.

\section{Detailed Methods}
\label{section3}
\begin{figure*}[!t]
	\centering
	\includegraphics[width=1.0\textwidth]{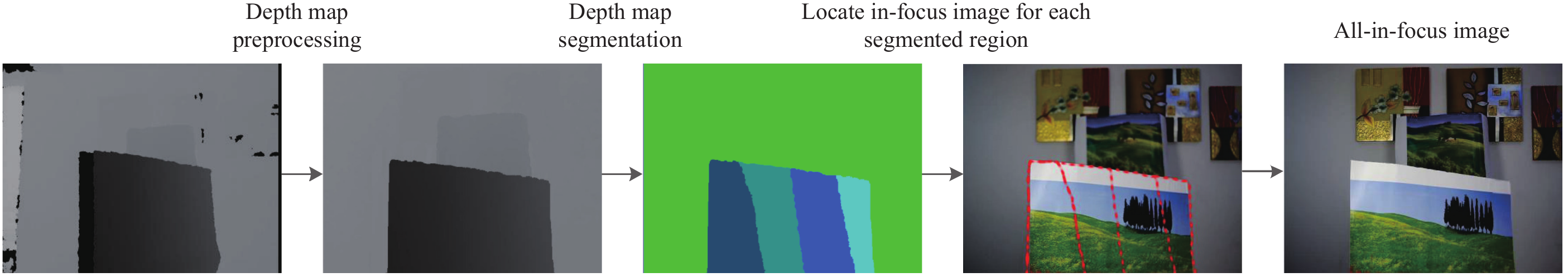}
	\caption{Main steps and intermediate results when applying the proposed multi-focus image fusion method to construct an all-in-focus image of a real scene.}
	\label{Fig3}
\end{figure*}

Fig. \ref{Fig3} uses an example to illustrate the main steps and intermediate results when the proposed multi-focus image fusion method is applied to construct an all-in-focus image of a scene. Firstly, the depth map from the Kinect depth sensor is preprocessed to align with the color image captured with the color camera, based on a stereo calibration method, and to recover the missing depth values. A graph-based image segmentation algorithm is then used to segment the preprocessed depth map into regions. A focus algorithm is used to locate an in-focus image for each region from among multi-focus source images to construct an all-in-focus image.

\subsection{Depth map preprocessing}
\subsubsection{Align depth map with color image}

Microsoft Kinect contains a depth sensor and an RGB camera that provides both depth and color streams with a resolution of 640 $\times$ 480 at 30 Hz. The depth sensor consists of an infrared (IR) projector combined with an IR camera. The IR projector projects a set of IR dots, the IR camera observes each dot and matches it with a dot in the known projector pattern to obtain a depth map. The operating range of the present Kinect depth sensor is between 0.5 m to 5.0 m \cite{khoshelham2012accuracy}.

Due to the different spatial positions and intrinsic parameters of the IR camera of the Kinect depth sensor and of the Pentax color camera, the depth map is not aligned with the color image. To align the depth map with the color image, the depth map is first mapped to 3D points in the IR camera's coordinate system using the intrinsic parameters of the IR camera. Then, these 3D points are transformed to the Pentax color camera's coordinate system using extrinsic parameters that relate the IR camera's coordinate system and the color camera's coordinate system. Finally, the transformed 3D points are mapped to the color image coordinate system using the intrinsic parameters of the color camera.

Let $(u_{0}, v_{0})$ denote the coordinates of the principal point of the IR camera, $f_x$ and $f_y$ denote the scale factors in image $u$ and $v$ axes of the IR camera, and $u_{0}$, $v_{0}$, $f_x$ and $f_y$ be the intrinsic parameters of the IR camera. Let $[u,v,Z]$ represent a pixel in the depth map, $Z$ represent the depth value in $[u,v]$, and $[X,Y,Z]^T$ represent the mapped 3D point of $[u,v]$ in the IR camera coordinate system. According to the pinhole camera model, the values of $X$ and $Y$ can be calculated according to
\begin{equation}
\begin{array}{l}
X{\rm{ = }}(u - {u_0})Z/{f_x},\\
Y = (v - {v_0})Z/{f_y}.
\end{array}
\end{equation}
Let $R$ and $T$ represent the rotation and translation that relate the coordinate system of the IR camera of the Kinect depth sensor and the color camera's coordinate system. $R$ and $T$ are the extrinsic parameters. $R$ is a $3 \times 3$ matrix, and $T$ is a $3 \times 1$ matrix. The relationship between the transformed 3D point ${{\left[ {X}',{Y}',{Z}' \right]}^{T}}$ in the color camera's coordinate system and $[X,Y,Z]^T$ can be expressed as
\begin{equation}
{\left[ {X',Y',Z'} \right]^T} = R{\left[ {X,Y,Z} \right]^T} + T.
\end{equation}
Let $(u_{0}', v_{0}')$ denote the coordinates of the principal point of the color camera and ${f_x}'$ and ${f_y}'$ denote the scale factors in image $u'$ and $v'$ axes of the color camera. After mapping ${{\left[ {X}',{Y}',{Z}' \right]}^{T}}$ to the color image coordinate system, the aligned depth point $[u',v',Z']$ can be obtained, where $u'$ and $v'$ are calculated according to
\begin{equation}
\begin{array}{l}
u' = \frac{{X'}}{{Z'}}{f_x}^\prime  + {u_0}^\prime \\
v' = \frac{{Y'}}{{Z'}}{f_y}^\prime  + {v_0}^\prime.
\end{array}
\end{equation}
The intrinsic parameters of the IR camera of the Kinect depth sensor and the color camera and their extrinsic parameters are determined using a stereo camera calibration method. In the example shown in Fig. \ref{Fig4.2}, there are many pixels in regions 1 and 2 that have a value of zero because the aligned depth regions 1 and 2 are larger than their corresponding regions 1 and 2 in Fig. \ref{Fig4.1}, and these pixels do not obtain depth values from Fig. \ref{Fig4.1}. A dilation operation is used to recover the depth value of these pixels. A 3$\times$3 rectangular structuring element is used to dilate the source depth map to determine the shape of a pixel`s neighborhood over which maximum is taken, according to
\begin{equation}
dst(x,y) = \mathop {\max }\limits_{(x',y'):element(x',y') \ne 0} src(x + x',y + y').
\end{equation}

\subsubsection{Depth map hole filling}
\begin{figure}[!t]
	\centering
	\subfigure[]{\label{Fig4.1} \includegraphics[width=0.23\textwidth]{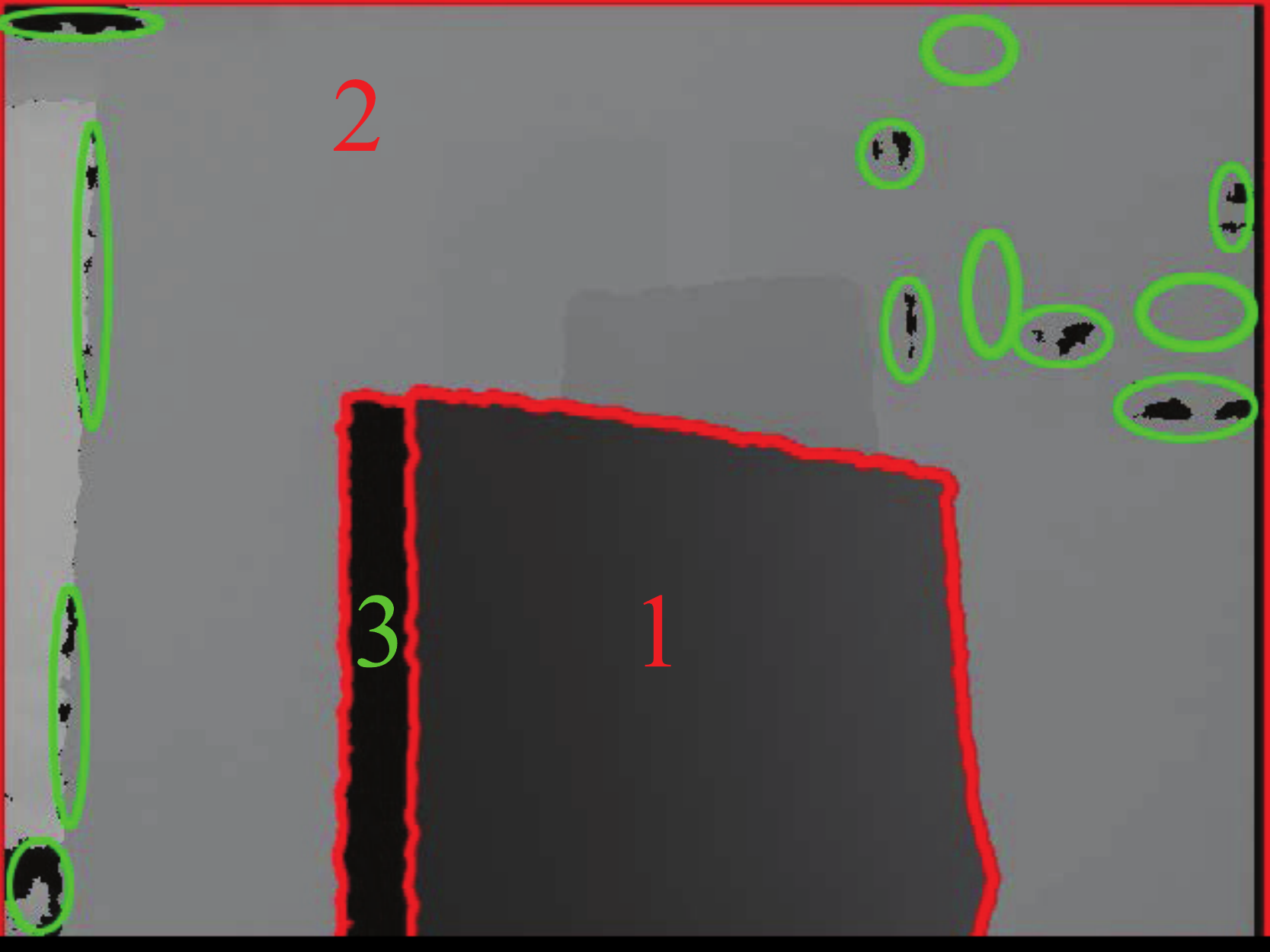}}
	\subfigure[]{\label{Fig4.2} \includegraphics[width=0.23\textwidth]{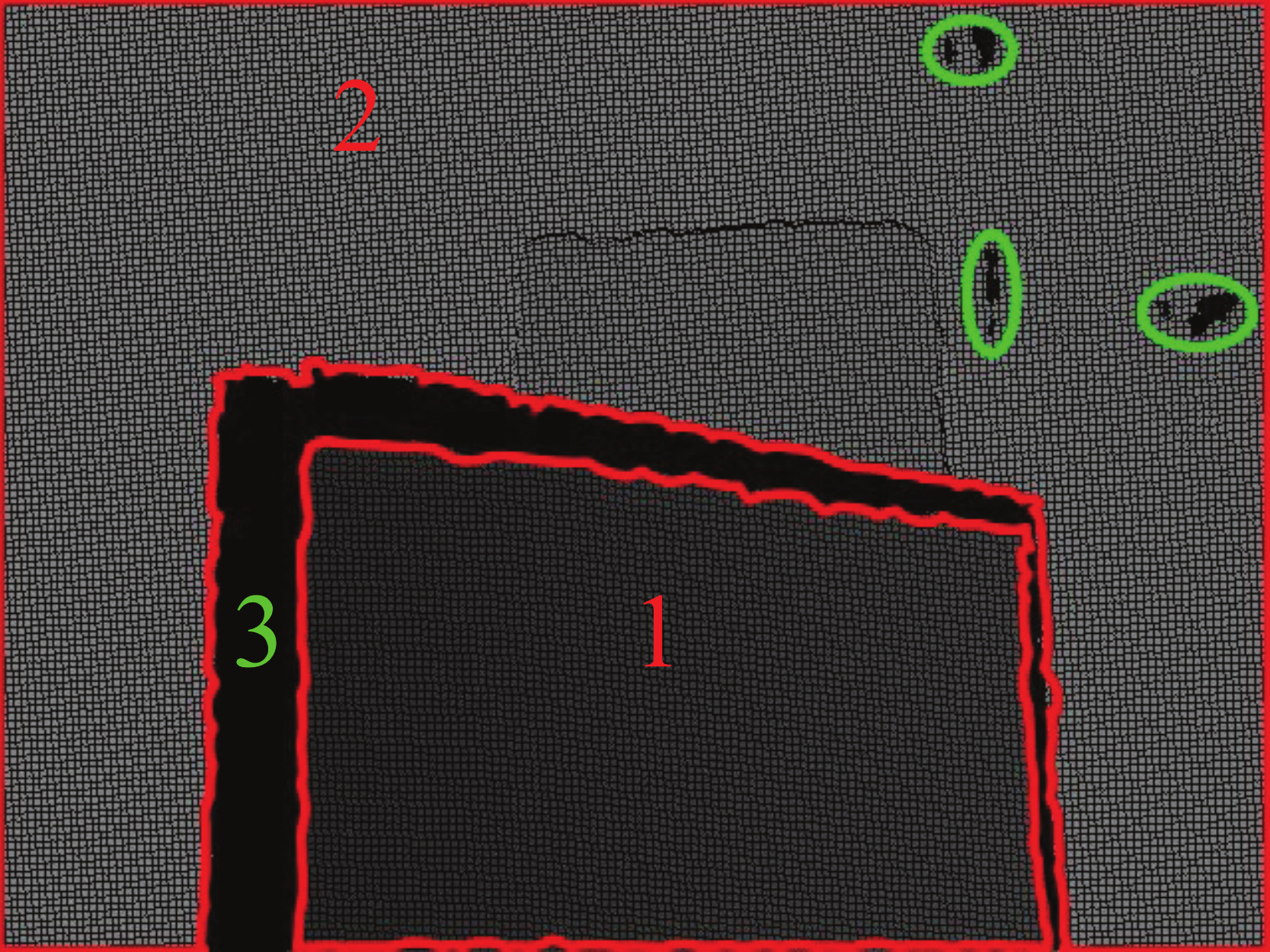}}
	\subfigure[]{\label{Fig4.3} \includegraphics[width=0.23\textwidth]{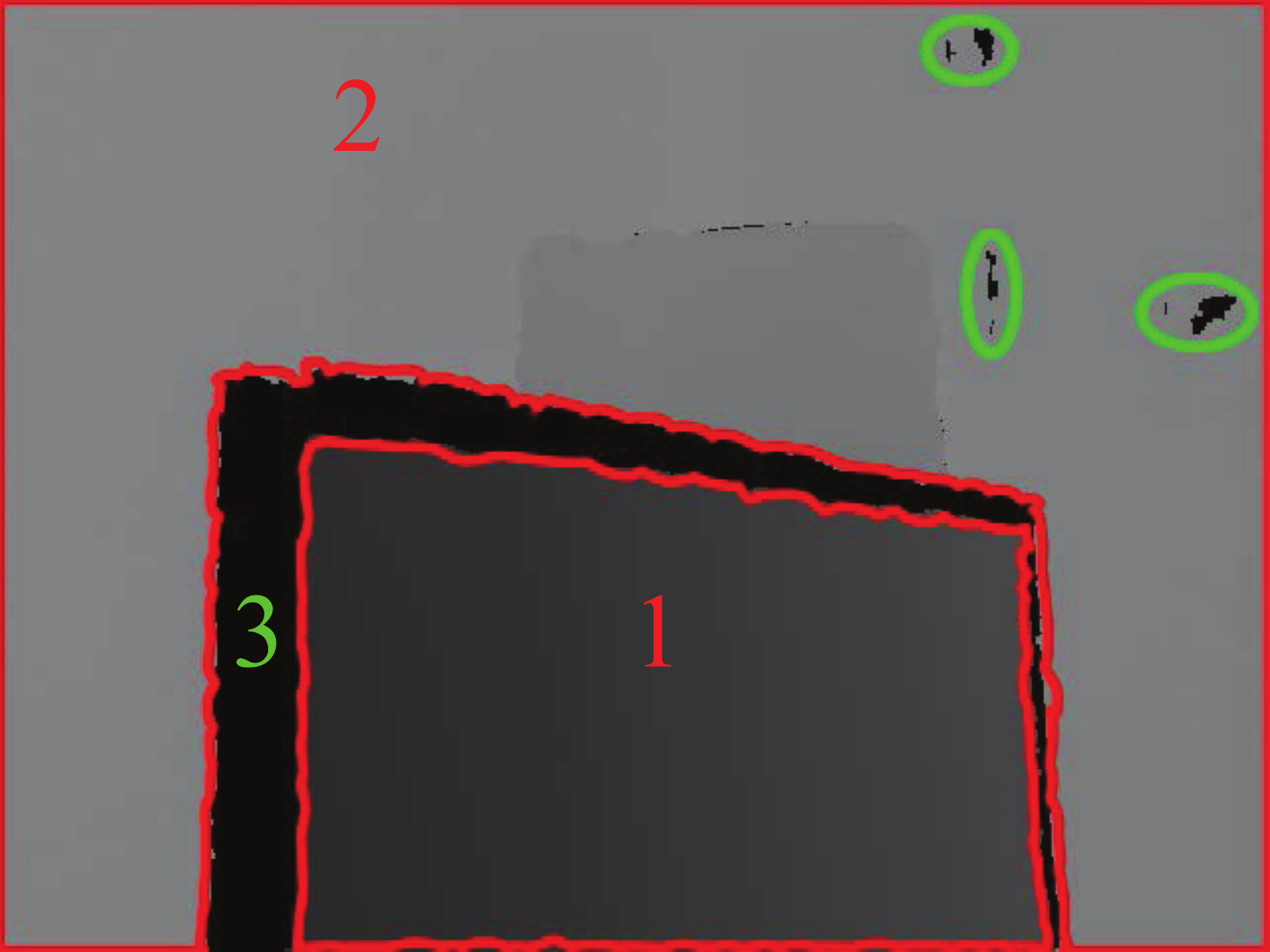}}
	\subfigure[]{\label{Fig4.4} \includegraphics[width=0.23\textwidth]{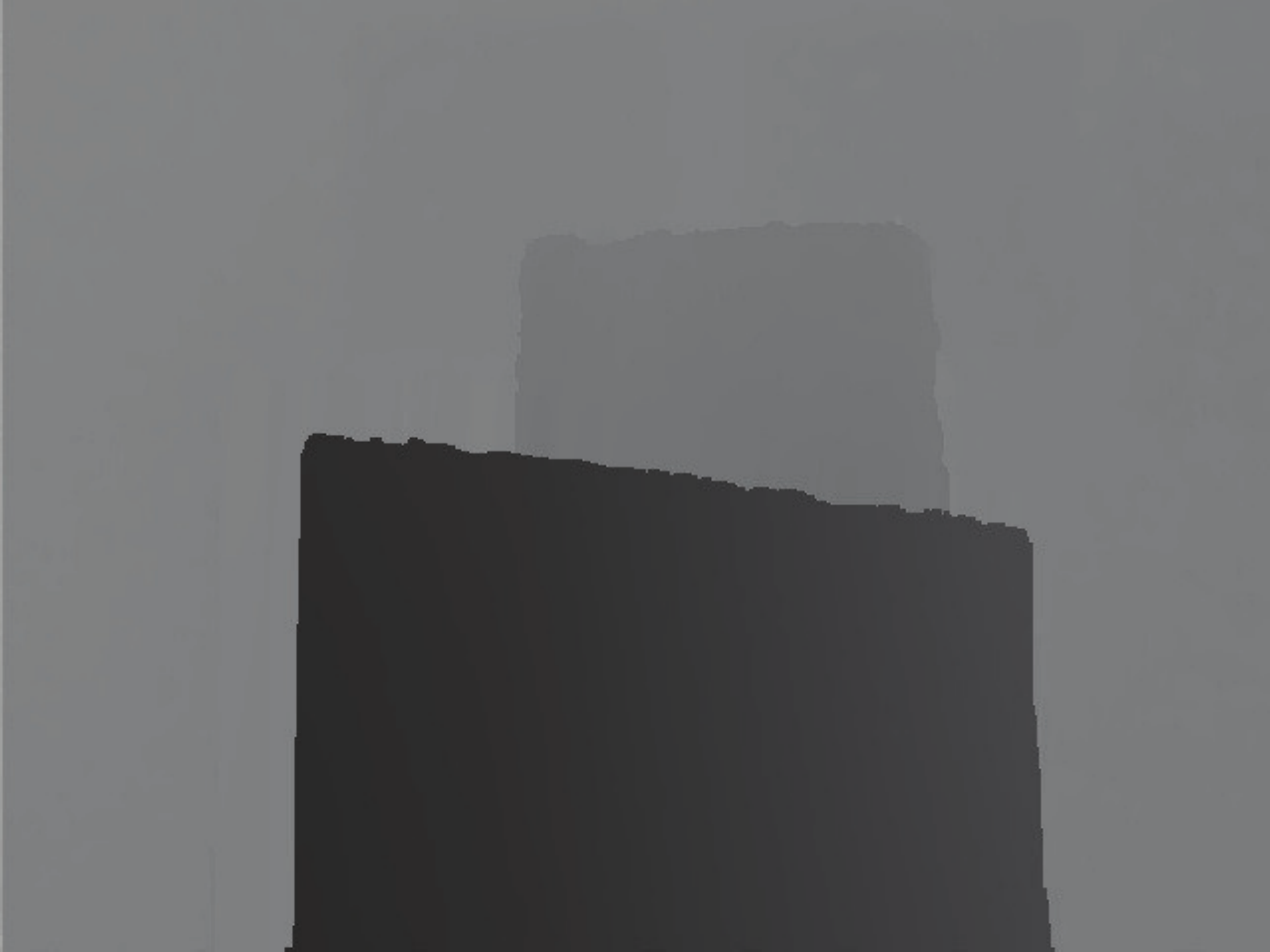}}
	\caption{Depth map preprocessing: (a) raw depth map from Kinect depth sensor, (b) aligned depth map, (c) aligned depth map after dilation, and (d) aligned depth map after hole filling. Black holes are labeled with green-colored ellipses, the largest hole is labeled with "3" in green color, the front object region and background region are labeled with "1" and "2" respectively.}
	\label{Fig4}
\end{figure}

From the aligned depth map (Fig. \ref{Fig4.3}), there still exist a number of black holes that are labelled with green-colored ellipses, and the largest hole labeled with ``3" in green color. These holes are caused by the structured light that the IR projector of the Kinect depth sensor emits, which was reflected in multiple directions, encountered with transparent objects, and scattered from object surfaces \cite{han2013enhanced}. To avoid incorrect segmentation, these depth holes must be filled.

The task is to use valid depth values around depth holes to fill the depth holes. Vijayanagar et al. \cite{vij2014real} proposed a multi-resolution anisotropic diffusion (AD) method, which uses the color image to diffuse the depth map and requires this process to be iterated many times in the multi-resolutions of the color image for each resolution. Differently, as discussed in the next sub-section on depth map segmentation, the depth value of a filled hole only needs to be within the DoF at its neighboring valid depth value. Therefore, the AD method is applied more efficiently in our work. (1) The AD filter is only applied to the depth map of the original size. (2) The conduction coefficients are only computed from the depth map. (3) Only one iteration of AD is applied because after one iteration, the differences between the depth value of the recovered pixel and its neighbors become less than the DoF at the recovered depth value, and thus, incorrect segmentation is avoided.

For an image $I$, the discrete form of the anisotropic diffusion equation, according to \cite{Perona1990Scale}, is
\begin{equation}
\begin{array}{l}\label{eq5}
I_{(i,j)}^{t + 1} = I_{(i,j)}^t + \lambda ({C_N} \cdot d_N + {C_S} \cdot d_S\\
\;\;\;\;\;\;\;\; + {C_W} \cdot d_W + {C_E} \cdot d_E)_{(i,j)}^t,
\end{array}
\end{equation}
where ${\rm{0}} \le \lambda  \le {\rm{0.25}}$ for the equation to be stable, $t$ indicates the current iteration, $d$ represents the depth value difference between the pixel $I_{(i,j)}$ and one of its four neighbors, and the subscripts $N$, $S$, $E$, $W$ denote the neighboring pixels to the north, south, east and west. The conduction coefficient $C$ is
\begin{equation}
C = g(d) = {e^{( - {{( d /K)}^2})}},
\end{equation}
where $K$ is the standard deviation.

To recover the depth value of $I_{(i,j)}$, since the IR projector is located on the right side of Kinect and the IR camera is on the left side, the main depth holes (region 3 in Fig. \ref{Fig4.3}) is always to the left of an object, we replace $I_{(i,j)}$ with $I_{(i - 2,j)}$ to fill depth holes. Thus, (\ref{eq5}) is rewritten as
\begin{equation}
\begin{array}{l}
{I_{(i,j)}} = {I_{(i - 2,j)}} + \lambda ({C_N} \cdot {d_N} + {C_S} \cdot {d_S}\\
\;\;\;\;\;\;\;\; + {C_W} \cdot {d_W} + {C_E} \cdot {d_E})_{(i-2,j)},
\end{array}
\end{equation}
where
\begin{equation}
\begin{array}{l}
{d_N} = {I_{(i - 3,j)}} - {I_{(i - 2,j)}},\\{d_S} = {I_{(i - 1,j)}} - {I_{(i - 2,j)}},\;\\
{d_W} = {I_{(i - 2,j - 1)}} - {I_{(i - 2,j)}},\\{d_E} = {I_{(i - 2,j + 1)}} - {I_{(i - 2,j)}}.\;
\end{array}
\end{equation}
The aligned depth map after hole filling is shown in Fig. \ref{Fig4.4}.

\subsection{Graph-based depth map segmentation}
\begin{figure}[!t]
	\centering
	\includegraphics[width=0.35\textwidth]{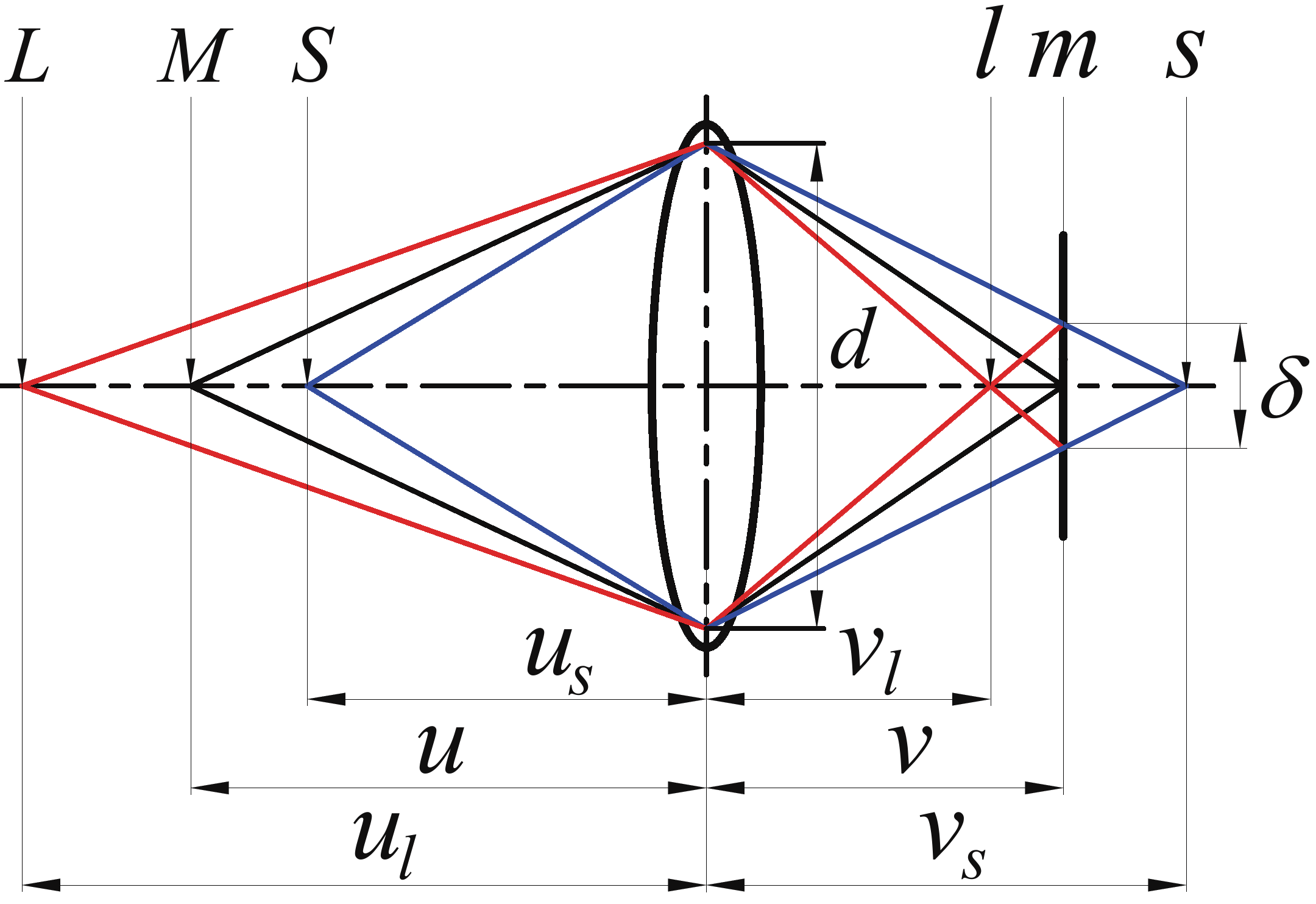}
	\caption{Diagram for calculating depth of field (DoF).}
	\label{Fig5}
\end{figure}

After preprocessing the depth map, the depth map is segmented into distinct image block regions. Each segmented region must satisfy the DoF rule, as described below, to ensure all objects in this region appear in focus. In Fig.\ref{Fig5}, scene point $L$ is at a distance of $u_l$ from the lens, $M$ is at a distance of $u$, and $S$ is at a distance of $u_s$. The three points are imaged as $l$ at a distance of v$_l$, $m$ at a distance of v, and $s$ at a distance of v$_s$. Among the three scene points, only $M$ is imaged in perfect focus at the image detector; $L$ and $S$ are imaged as a blurred circle with diameter $\delta$ centered around $m$. The DoF consists of two parts, the back DoF ($b\_DoF$) and front DoF ($f\_DoF$), and their values at a distance of $u$ can be derived as
\begin{equation}\label{eq9}
b\_DoF(u) = {u_l} - u = F\delta {u^2}/({f^2} - F\delta u)
\end{equation}
\begin{equation}\label{eq10}
f\_DoF(u) = u - {u_s} = F\delta {u^2}/({f^2} + F\delta u)
\end{equation}
where $F = f/d$ is the aperture value. Let $Min$ and $Max$ represent the minimum and maximum depth values in a segmented region, respectively, and let $Diff$ represent the difference between $Min$ and $Max$ (i.e., $Diff$ = $Max$ - $Min$). Let $b\_DoF(Min)$ represent the back DoF when the camera is in focus at $Min$, $f\_DoF(Max)$ represent the front DoF when the camera is in focus at $Max$, and $MaxDoF$ represent the larger value between $b\_DoF(Min)$ and $f\_DoF(Max)$. To ensure all objects in a segmented region all appear focused, the DoF rule requires that $Diff$ must be smaller than $MaxDoF$ (i.e., $Diff$ $<$ $MaxDoF$).

In graph theory-based segmentation algorithms, a graph with vertices, image pixels, and edges corresponding to pairs of neighboring vertices is established. Each edge has a weight initialized by the difference between the values of pixels on each side of the edge. In existing graph theory-based segmentation algorithms, blocks of pixels with low variability tend to be segmented into a single region. For an object with a wide depth range, the entire object crosses multiple DoFs and cannot appear focused in one focus setting. In this case, standard graph-based segmentation algorithms would incorrectly segment the entire object into a single region. For objects within a specific DoF of the camera but with different depth values, the standard graph-based segmentation algorithms may unnecessarily segment these objects into different regions.

\begin{figure}[!t]
	\centering
	\subfigure[]{\label{Fig6.1} \includegraphics[width=0.23\textwidth]{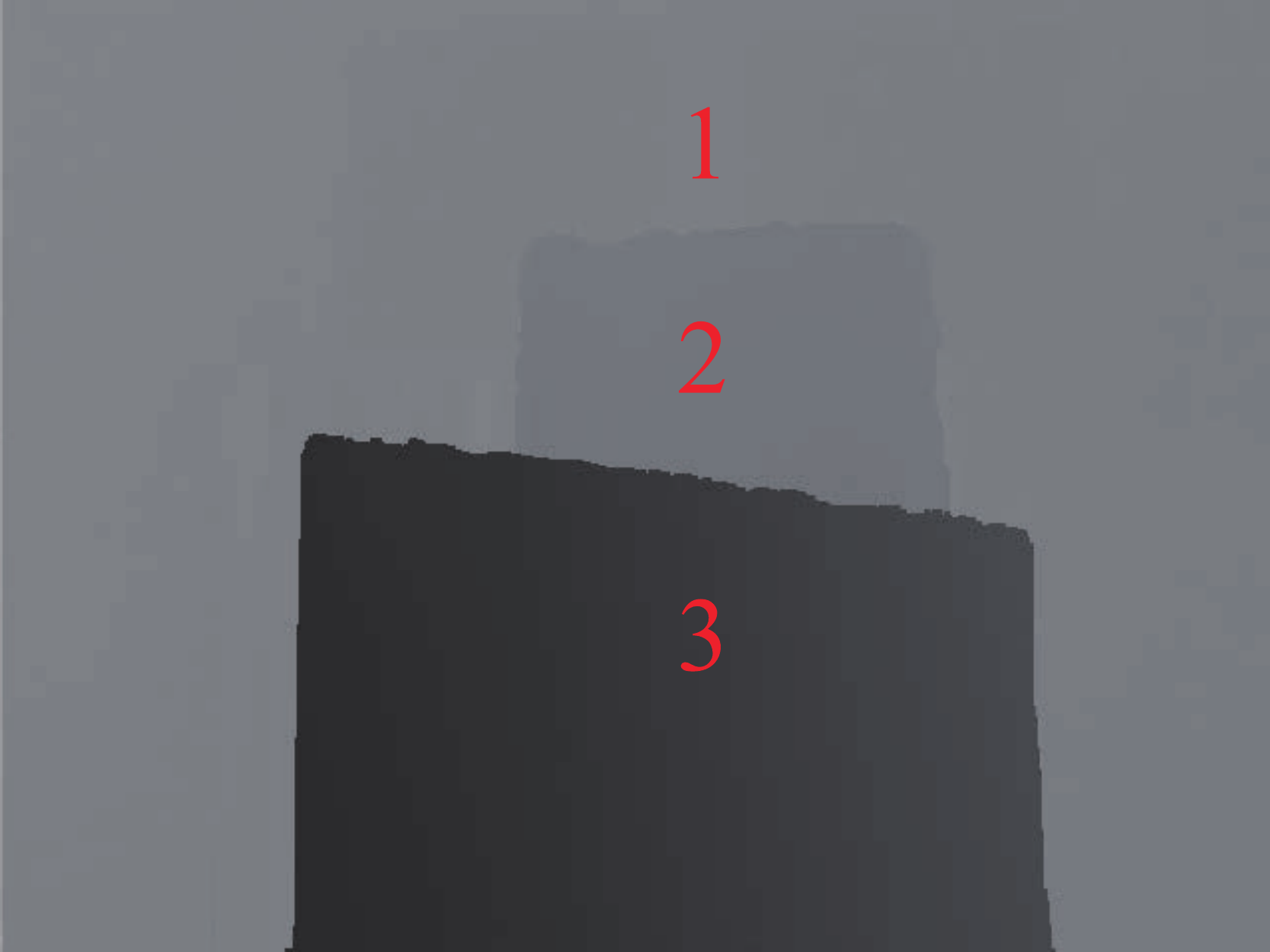}}
	\subfigure[]{\label{Fig6.2} \includegraphics[width=0.23\textwidth]{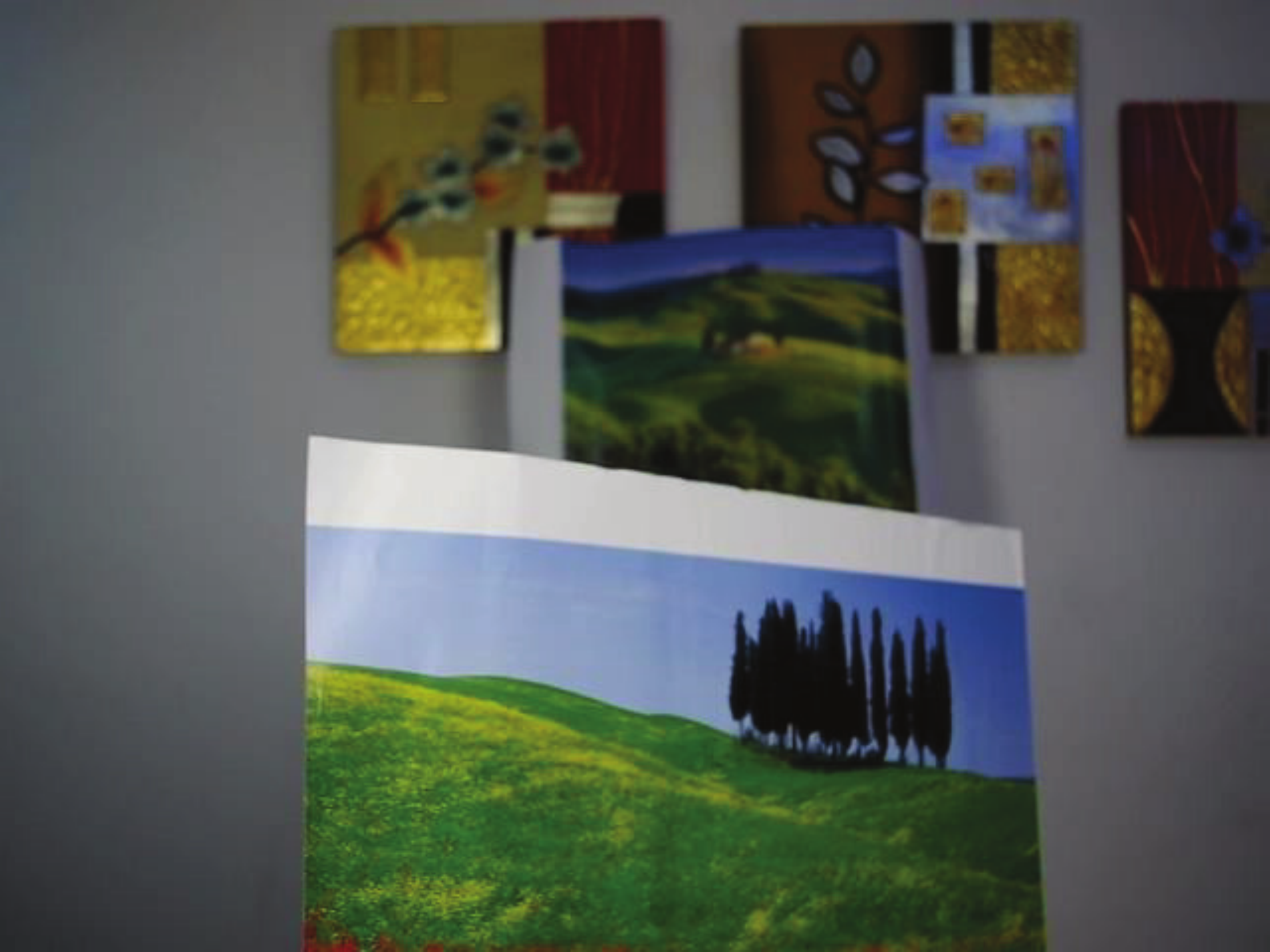}}
	\subfigure[]{\label{Fig6.3} \includegraphics[width=0.23\textwidth]{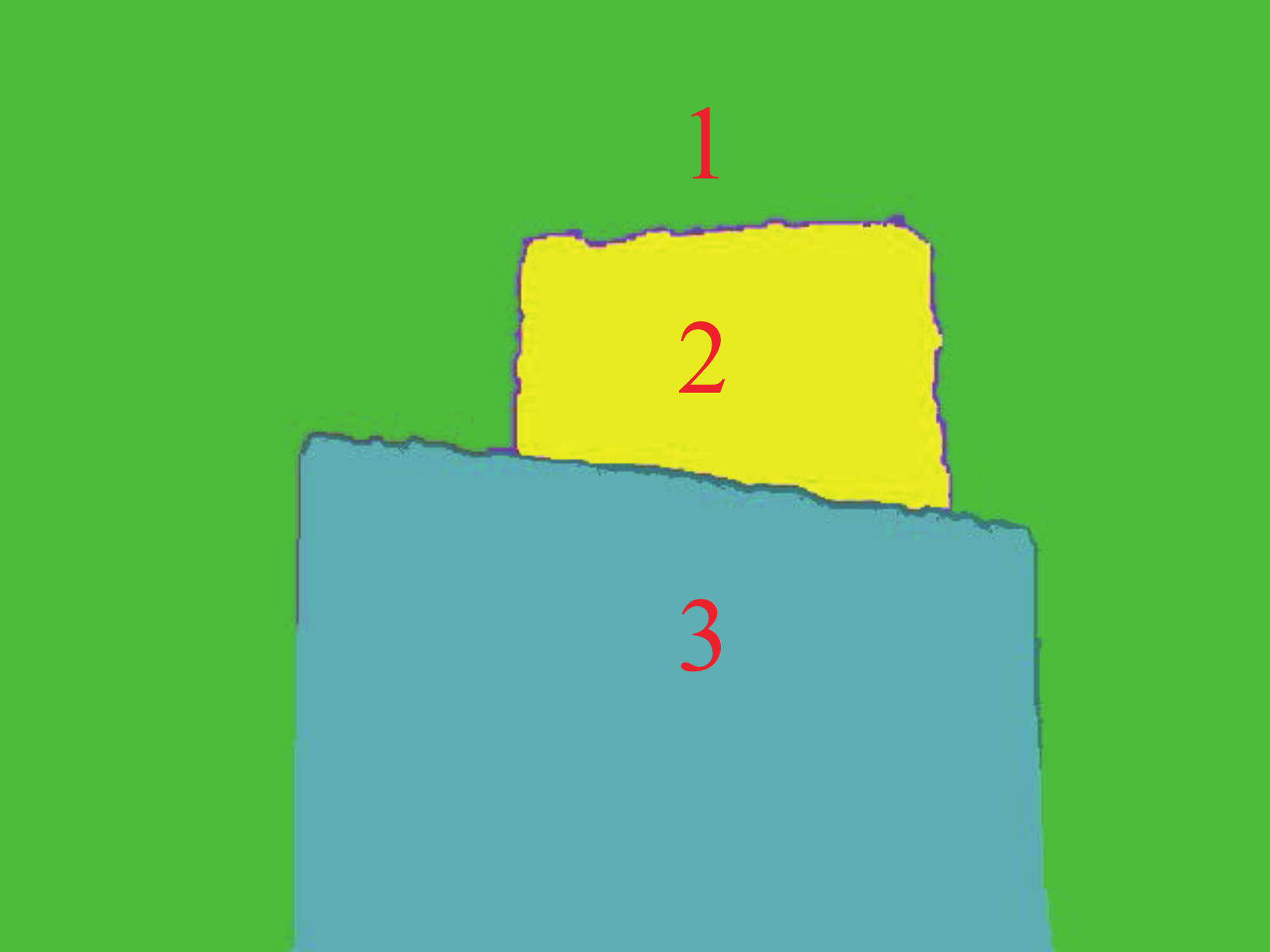}}
	\subfigure[]{\label{Fig6.4} \includegraphics[width=0.23\textwidth]{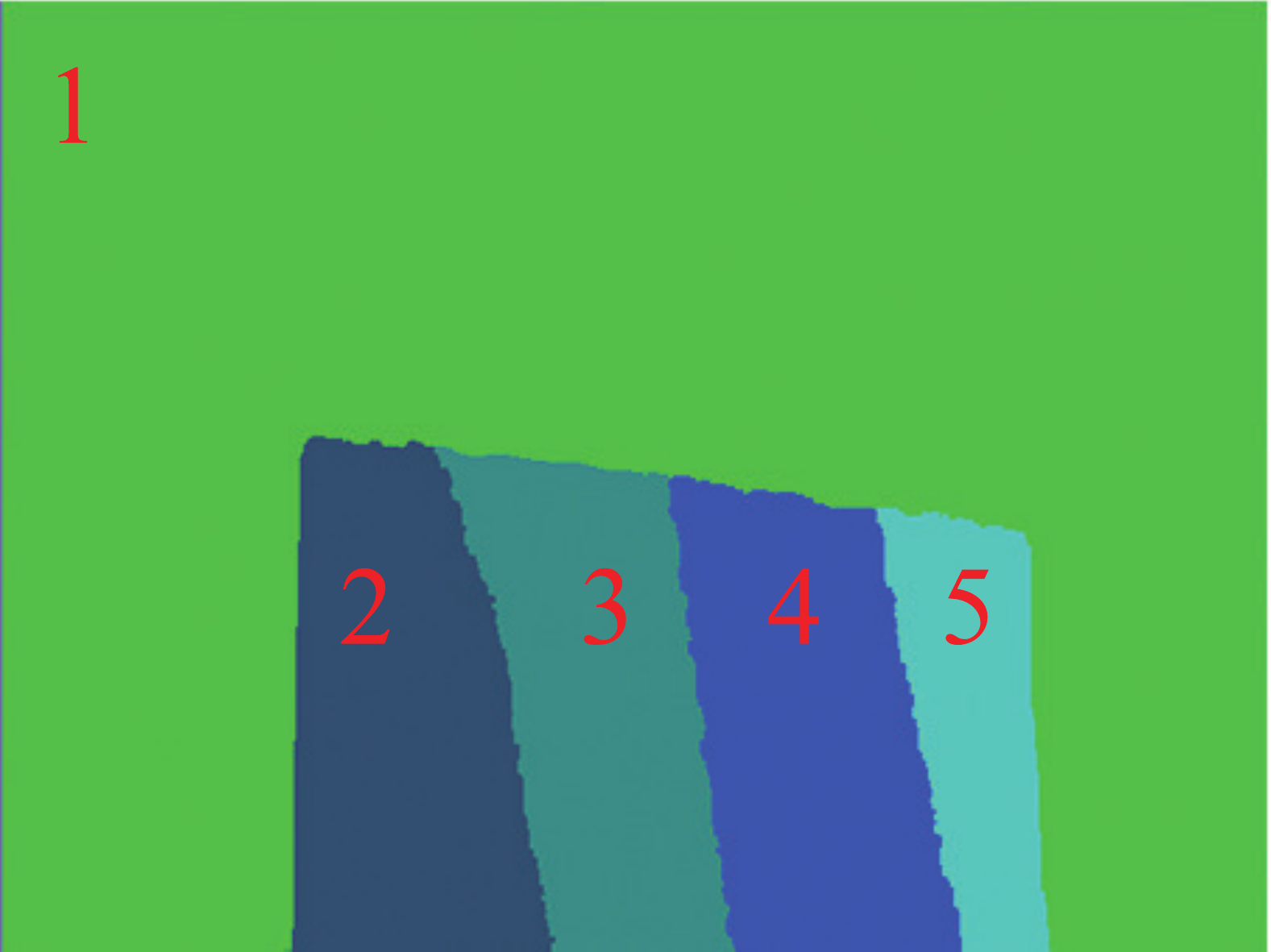}}
	\caption{Depth map segmentation using the standard graph-based segmentation algorithm and our modified graph-based segmentation algorithm. (a) Raw depth map of a real scene. (b) Color image of the scene.  (c) Segmentation result by using the standard graph-based Felz algorithm. (d) Segmentation result using our modified algorithm.}
	\label{Fig6}
\end{figure}

Fig. \ref{Fig6.1} is the depth map of a real scene with its corresponding color image shown in Fig. \ref{Fig6.2}). We first applied the classic graph-based segmentation algorithm (Felz algorithm) \cite{Felzenszwalb2004Efficient} which segmented the depth map into three regions (Fig. \ref{Fig6.3}). TABLE \ref{T1} summarizes the values of $Min$, $Max$, $Diff$ and $MaxDoF$ of each segmented region. For region 3, $Diff$ is larger than $MaxDoF$, indicating that all the objects in region 3 cannot appear focused in one focus setting. Since the depth values in region 3 change gradually from 832 $mm$ to 1360 $mm$, they were incorrectly segmented into a single region. For regions 1 and 2, when the camera was set to focus at the minimum depth value in region 2 (2417 $mm$), $b\_DoF$ was 1132 $mm$, which is larger than the difference (698 $mm$) between the minimum depth value in region 2 (2417 $mm$) and the maximum depth value in region 1 (3115 $mm$), indicating that the objects in regions 1 and 2 can appear focused in one focus setting. In summary, with the Felz algorithm, regions 1 and 2 in Fig. \ref{Fig6.3} were unnecessarily segmented into two regions, and region 3 in Fig. \ref{Fig6.3} was incorrectly regarded as a single region.

In our depth map segmentation, a graph-based representation of the depth map is first established, in which pixels are nodes, and edge weights measure the dissimilarity between nodes (e.g., depth differences). Given two components, $C_1$ and $C_2$, let $min$ and $max$ represent the minimum and maximum depth values among all the depth pixels within $C_1$ and $C_2$, $diff$ equal $max$ minus $min$, and $b\_DoF(min)$ and $f\_DoF(max)$ represent the back DoF and front DoF when the camera is set to focus at $min$ and $max$, respectively. To ensure that the final segmented regions can all appear focused in one focus setting of the camera, we then impose the rule of DoF, i.e., only if $diff$ is less than the larger value of $b\_DoF(min)$ and $f\_DoF(max)$ can the two components be merged.

The segmentation result using the proposed graph-based depth map segmentation algorithm is shown in Fig. \ref{Fig6.4}. The $Min$, $Max$, $Diff$ and $MaxDoF$ values of each segmented region are shown in TABLE \ref{T2}. It can be seen that in every region, $Diff$ is less than $MaxDoF$, indicating that all objects within each region can appear focused in one focus setting.

\begin{table}[h]
	\small\centering
	\caption{Depth values (in $mm$) of segmented regions in Fig. \ref{Fig6.3}}
	\setlength{\tabcolsep}{3pt}
	\begin{tabular}{cccccc}
		\toprule
		\toprule
		Region&$Min$&$Max$&$Diff$&$MaxDoF$&$Diff<MaxDoF~?$\\
		\hline			
		1&2722&3115&393&1526&Yes \\
		2&2417&2639&222&1132&Yes \\
		3&832&1360&528&207&No	\\	
		\bottomrule
		\bottomrule
	\end{tabular}
	\label{T1}
\end{table}

\begin{table}[h]
	\small\centering
	\caption{Depth values (in $mm$) of segmented regions in Fig. \ref{Fig6.4}}
	\setlength{\tabcolsep}{3pt}
	\begin{tabular}{cccccc}
		\toprule
		\toprule
		Region&$Min$&$Max$&$Diff$&$MaxDoF$&$Diff<MaxDoF~?$\\
		\hline			
		1&2463&3140&677&1186&Yes \\
		2&855&962&107&109&Yes \\
		3&950&1085&135&136&Yes	\\	
		4&1088&1269&181&182&Yes	\\	
		5&1273&1412&139&257&Yes	\\	
		\bottomrule
		\bottomrule
	\end{tabular}\\
	\label{T2}
\end{table}

\section{Experiments}\label{Section4}
\subsection{Evaluation metrics}
Seven representative fusion methods were selected for comprehensive comparisons with our proposed method. These five methods are discrete wavelet transform (DWT) \cite{Li2002Multisensor}, nonsubsampled contourlet transform (NSCT) \cite{Zhang2009Multifocus}, image matting (IM) \cite{2013IM}, guided filtering (GF) \cite{Li2013ImageGuided}, spatial frequency-motivated pulse coupled neural networks in nonsubsampled contourlet transform domain (NSCT-PCNN) \cite{Xiao2008ImageNSCTPCNN}, dense SIFT (DSIFT) \cite{Liu2015Multi}, and deep convolutional neural network (DCNN) \cite{Liu2017Multi}. DWT and NSCT are multi-scale transform methods; IM and GF are spatial methods; NSCT-PCNN is a PCNN-based and multi-scale transform method; DSIFT is a feature space method; and DCNN is a deep learning method. The source codes of these algorithms were obtained on line (see Supplementary Material).

In image fusion applications, there lacks a reference image or a fused image as ground truth for comparing different algorithms. As reported in \cite{Liu2012Objective}, fusion metrics are categorized into four groups: (1) information theory-based metrics, (2) image feature-based metrics, (3) image structural similarity-based metrics, and (4) human perception-inspired fusion metrics. In the experiments, six fusion metrics covering all the four categories were chosen, including normalized mutual information  ${Q_{MI}}$ \cite{Hossny2008Comments}, nonlinear correlation information entropy ${Q_{NCIE}}$ \cite{Wang2005A}, gradient-based fusion metric ${Q_{G}}$ \cite{Xydeas2000Objective}, phase congruency-based fusion metric ${Q_{P}}$ \cite{Liu2012Objective}, Yang's fusion metric ${Q_{Y}}$ \cite{Yang2008A}, and the Chen-Blum metric ${Q_{CB}}$ \cite{Chen2009A}. ${Q_{MI}}$ and ${Q_{NCIE}}$ are information theory-based metrics; ${Q_{G}}$ and ${Q_{P}}$ are image feature-based metrics; ${Q_{Y}}$ is an image structural similarity-based metric; and ${Q_{CB}}$ is a human perception-based metric. Detailed definitions of these metrics are provided in Supplementary Material. These six fusion metrics were implemented using the image fusion evaluation toolbox at \url{https://github.com/zhengliu6699}. For all the six metrics, a larger value indicates a better fusion result.

\subsection{Source images}
Multi-focus source images from five different scenes were captured and used in this study (Fig. S2 in Supplementary Material). Fig. \ref{Fig7} shows the source images of one of the scenes. Fig. \ref{Fig7.1} is the depth map of the scene. Depth map segmentation resulted in only two regions: the front region and the background region, as shown in Fig. \ref{Fig7.2}. Thus, the focus algorithm was guided to locate the two multi-focus source images (Fig. \ref{Fig7.3} and Fig. \ref{Fig7.4}), which were then used to construct an all-in-focus image. The scenes tested in this work were intentionally set to have only two regions, and there were only two multi-focus source images because the on-line image fusion evaluation toolbox (\url{https://github.com/zhengliu6699}) was designed to evaluate the fusion performance of two source images, and all the source codes of different multi-focus image fusion methods were also designed to fuse two images. In Supplementary Material, Fig. S1 shows the use of the ZED stereo camera for obtaining the depth map of more complex nature scenes and the segmentation results of the proposed depth map segmentation algorithm. The source images of other scenes are provided in Supplementary Material and can be downloaded from the author's GitHub website (https://github.com/robotVisionHang).

\subsection{Comparison results}

The assessment metric values of the all-in-focus images constructed using our proposed method and other multi-focus image fusion algorithms for different scenes are summarized in TABLE \ref{T3}. For each metric, the numbers in parentheses denote the score of each of the seven methods. The highest score is 7, and the lowest score is 1. The higher the score, the better the method.

TABLE \ref{T4} shows the number of times of each method receiving a score, the total score of each method, and the overall ranking of the eight methods. Among the eight methods, our proposed method received a score of 8 for the highest number of times and had the highest overall ranking. The results also reveal that our proposed method, DCNN, DSIFT and IM outperformed GF, NSCT-PCNN, DWT, and NSCT, and GF performed better than other multi-scale transform methods (NSCT-PCNN, DWT, and NSCT).

The core process of state-of-the-art RGB-based multi-focus image fusion methods (e.g., DCNN, DSIFT, GF and IM) is to compute a weight map by comparing the relative clearness level of multi-focus source images based on deep convolutional neural network, dense SIFT feature, guided filter, and image matting, respectively. In our proposed method, the weight map is generated through segmenting the depth map. Take $A$ and $B$ as two multi-focus source images, and $W$ is the weight map. A fused image, $F$ is constructed according to
\begin{equation}
F{\rm{ = }}(1.0 - W)*A + W*B,
\label{t11}
\end{equation}
where $*$ is an operation of pixel-wise multiplication. The range of values for $W$ is $0.0$ to $1.0$. In a position $(i,j)$ within $W$, a value of $0.0$ means the fusion method judges that $A$ is definitely clearer than $B$, and a value of $1.0$ means $B$ is definitely clearer than $A$ in $(i,j)$. If the fusion method is uncertain about whether $A$ is definitely clearer than $B$, it assigns a value between $0.0$ and $1.0$ to represent the clearness level of $A$ compared with $B$. A value less than $0.5$ indicates that $A$ is considered to be probably clearer than $B$; a value of $0.5$ indicates that $A$ and $B$ are considered to be equally clear; and a value higher than $0.5$ indicates that $B$ is considered to be probably clearer than $A$.

The better performance of our proposed method than other multi-focus image fusion methods can be understood by examining the weight maps they generated. For GF, the weight map for the detail layer was used to reconstruct the base layer and the detail layer of the fused image due to its more detailed reflection of the level of sharpness compared with the weight map for the base layer. Interestingly, the fused image reconstructed only with a detail layer (vs. with both base layer and detail layer \cite{Li2013ImageGuided}) generally obtained a higher score (see TABLE S1 in Supplementary Material).

The values in the weight map of DSIFT can take on $0.0$, $0.5$, or $1.0$, and for DCNN, IM and GF, the values range from $0.0$ to $1.0$. In our proposed method, the weight map is generated through the segmented regions. For a scene with only two segmented regions, the values in the weight map within a segmented region are all zeros since the pixels of one multi-focus source image within this region is considered in best focus. Similarly, the values in the weight map within the other segmented region are all ones.

\begin{figure*}[]
	\centering	
	\subfigure[]{\label{Fig7.1} \includegraphics[width=0.25\textwidth]{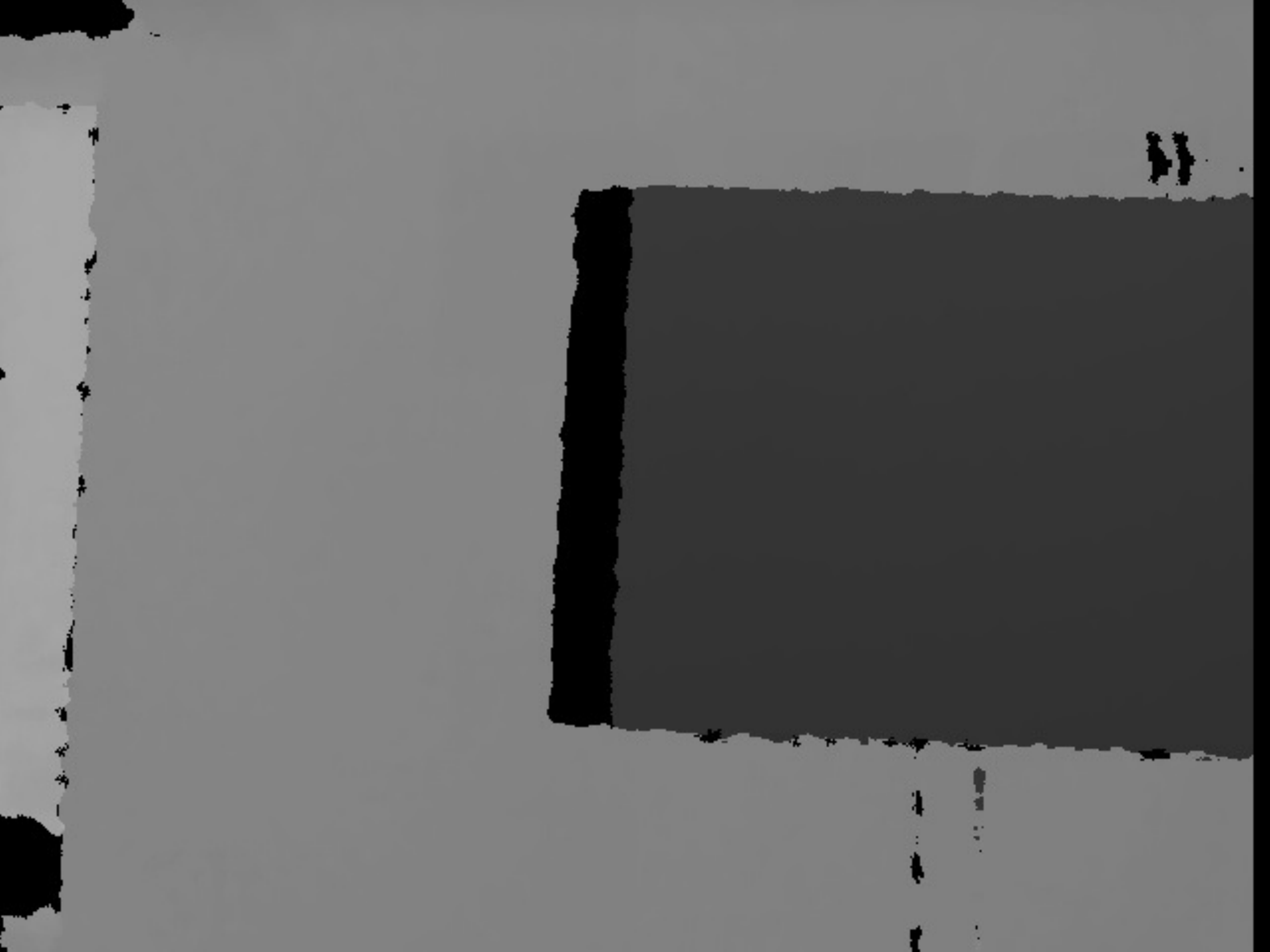}}
	\subfigure[]{\label{Fig7.2} \includegraphics[width=0.25\textwidth]{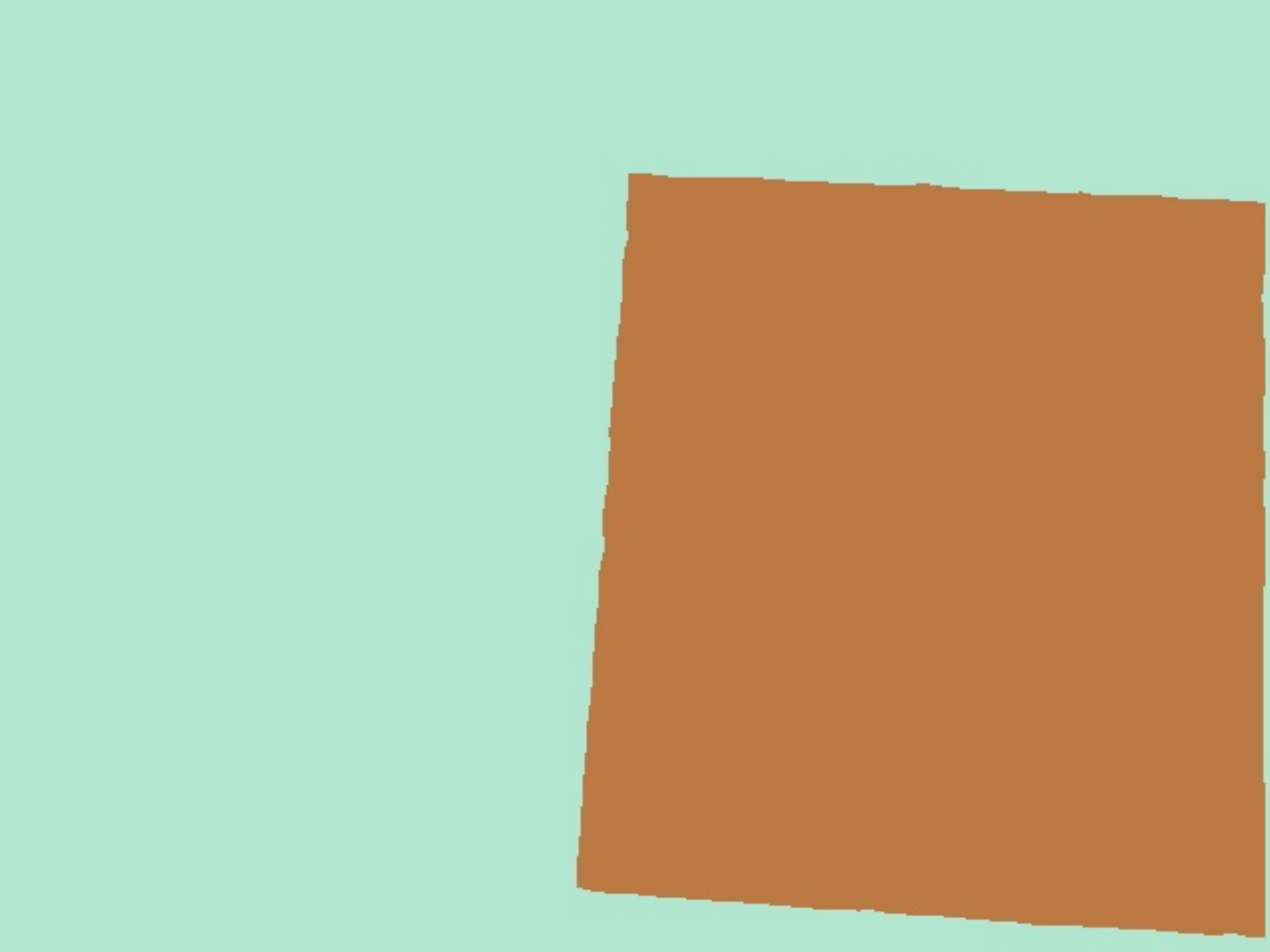}}
	\subfigure[]{\label{Fig7.3} \includegraphics[width=0.25\textwidth]{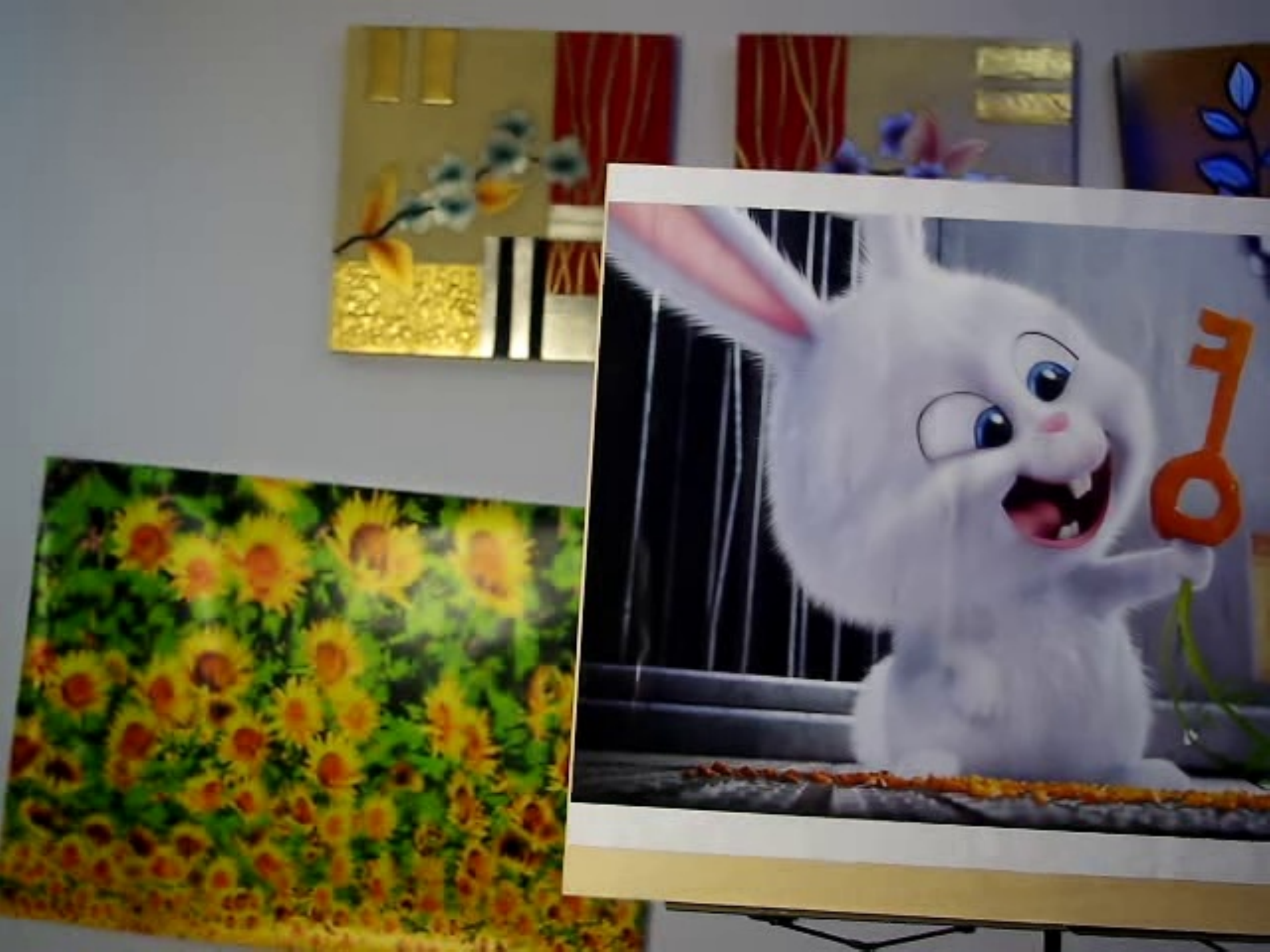}}
	\subfigure[]{\label{Fig7.4} \includegraphics[width=0.25\textwidth]{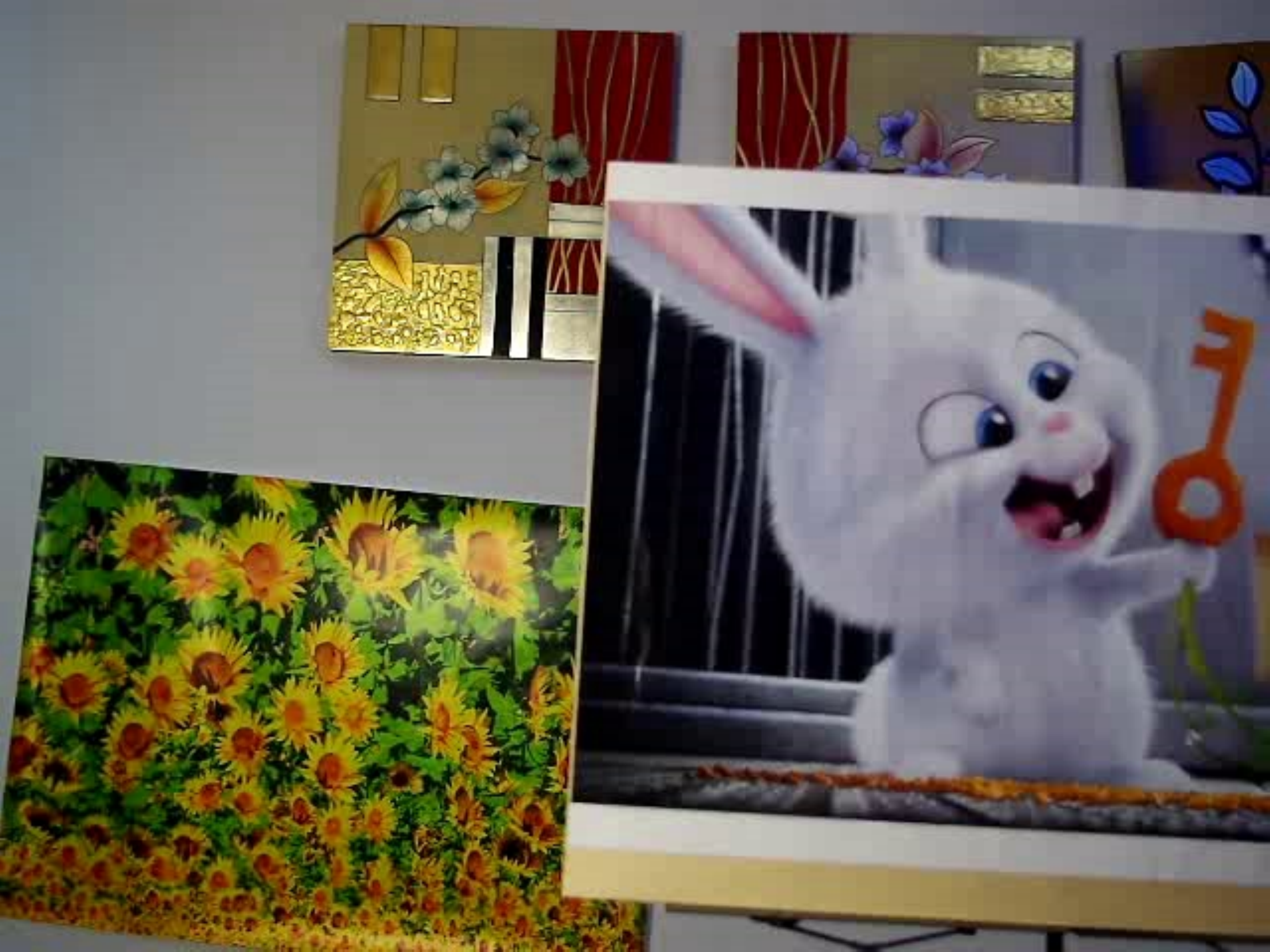}}
	\subfigure[]{\label{Fig8.1} \includegraphics[width=0.25\textwidth]{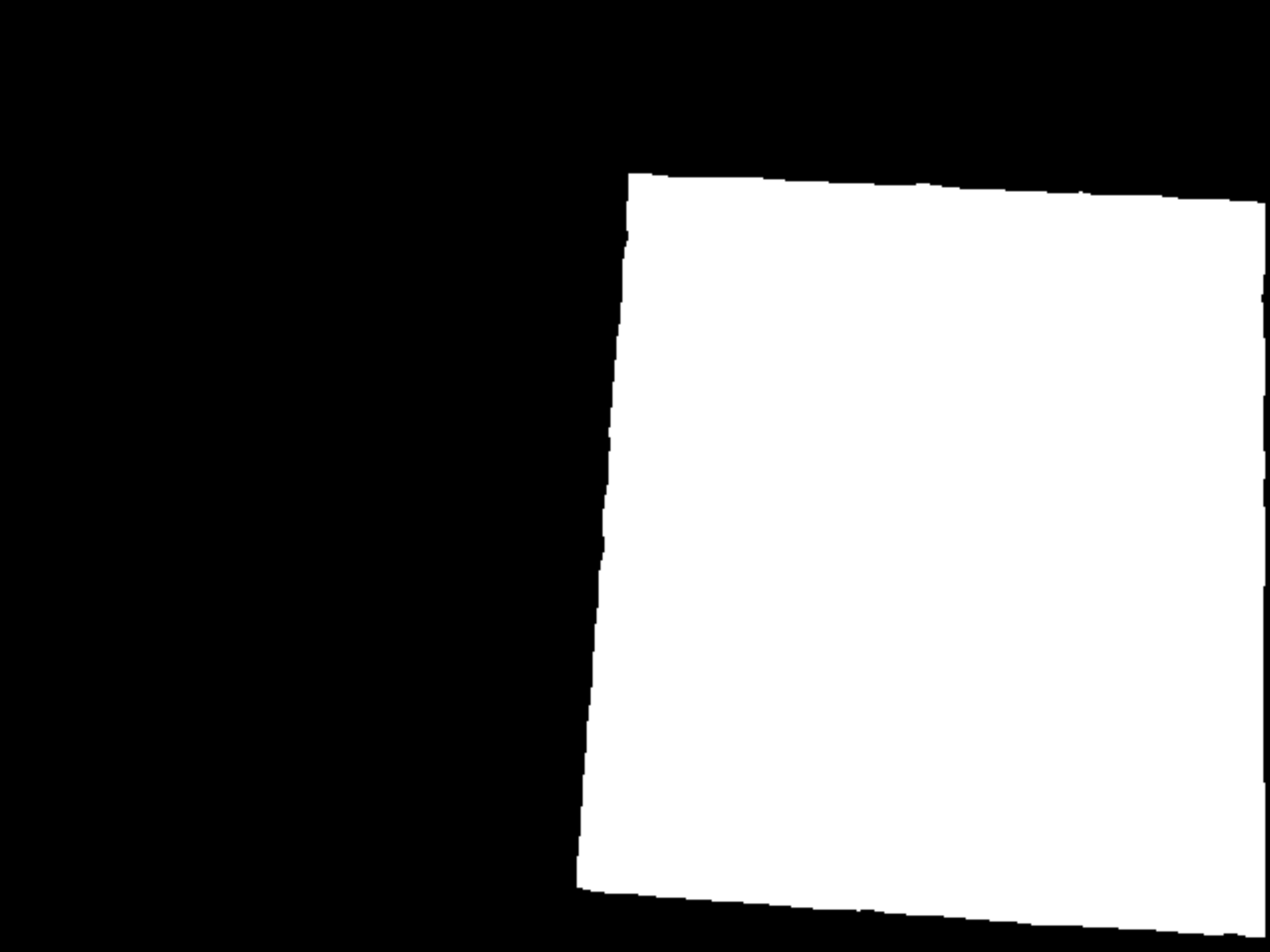}}
	\subfigure[]{\label{Fig8.5} \includegraphics[width=0.25\textwidth]{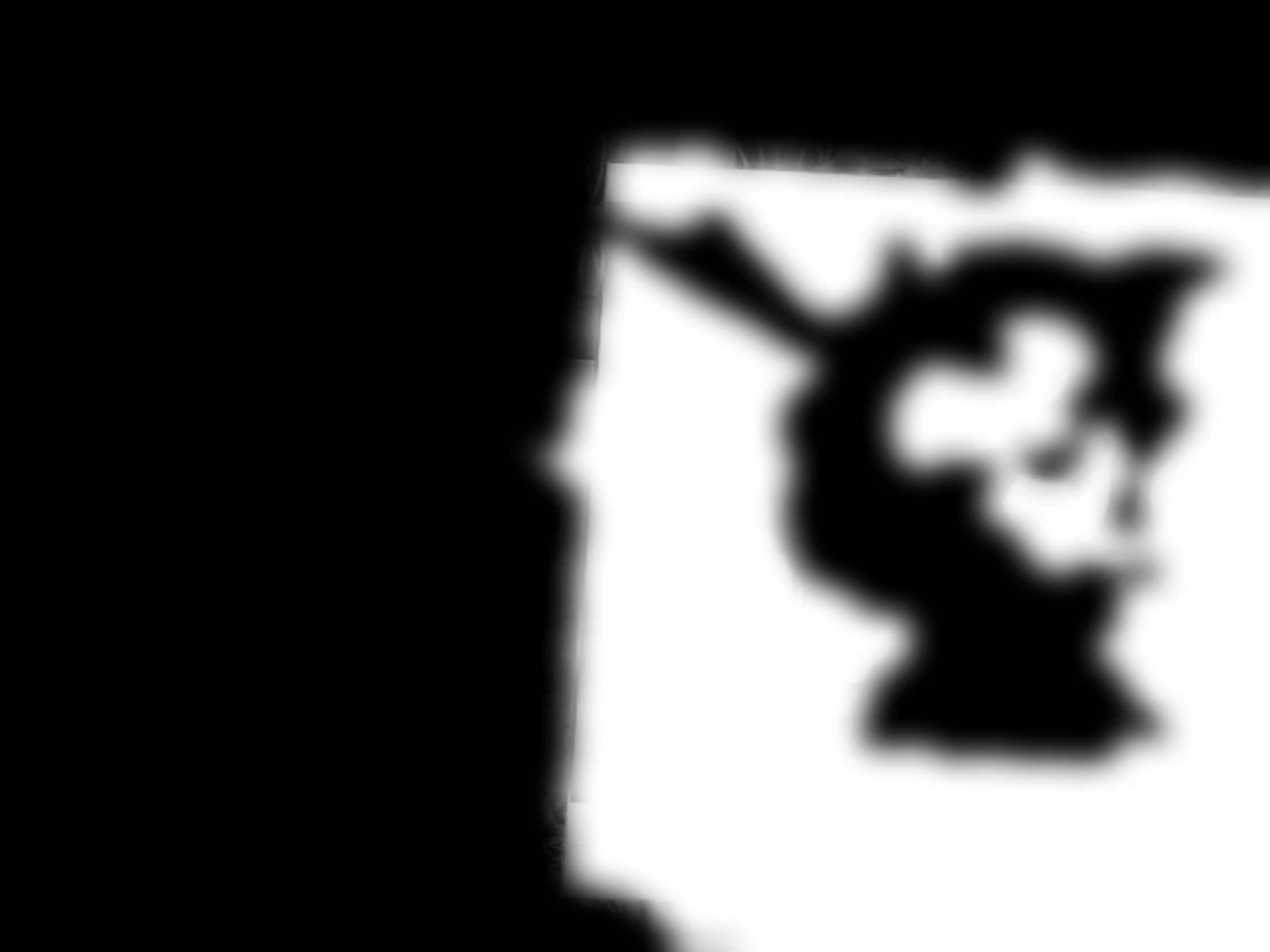}}	
	\subfigure[]{\label{Fig8.2} \includegraphics[width=0.25\textwidth]{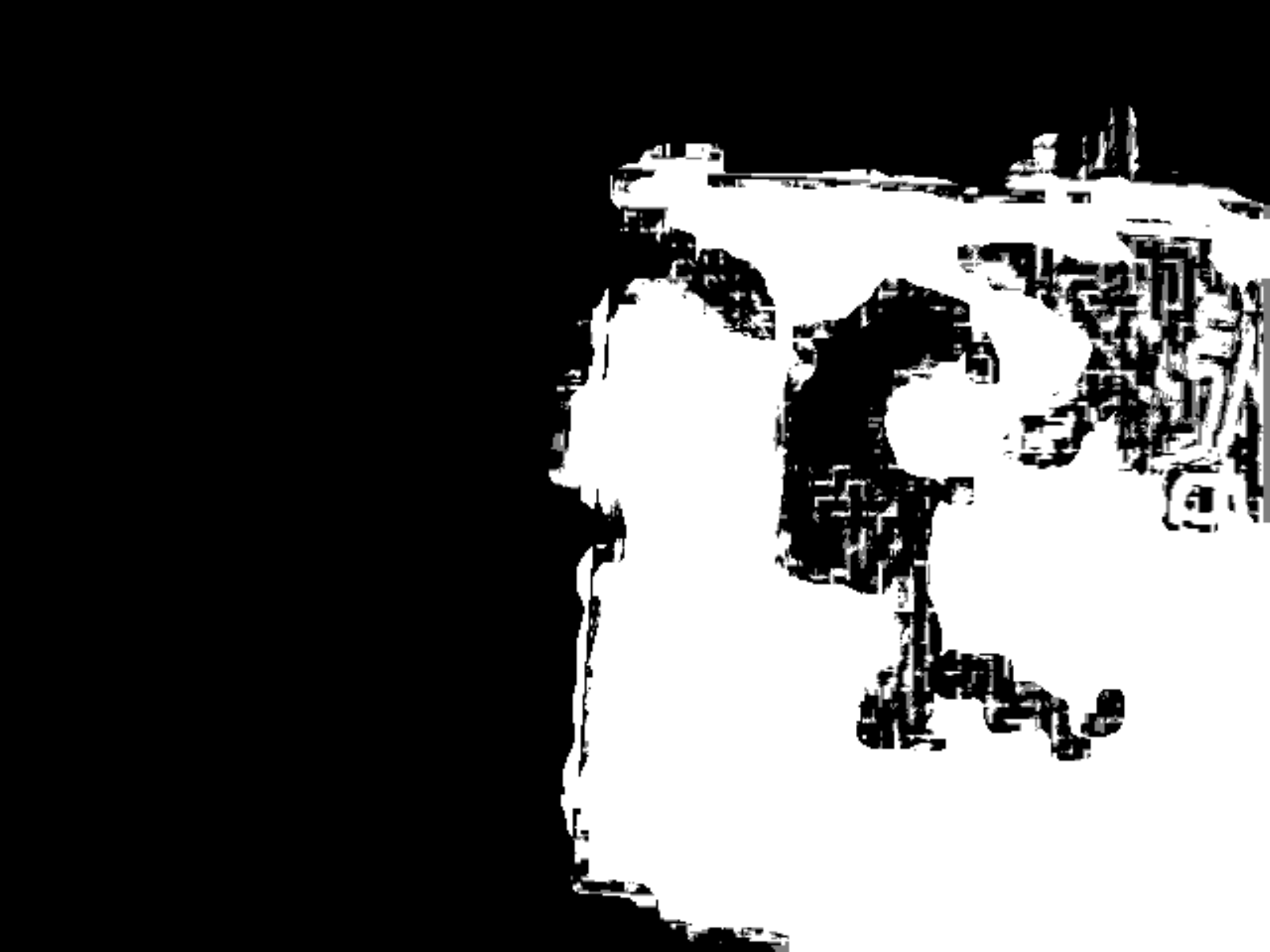}}
	\subfigure[]{\label{Fig8.3} \includegraphics[width=0.25\textwidth]{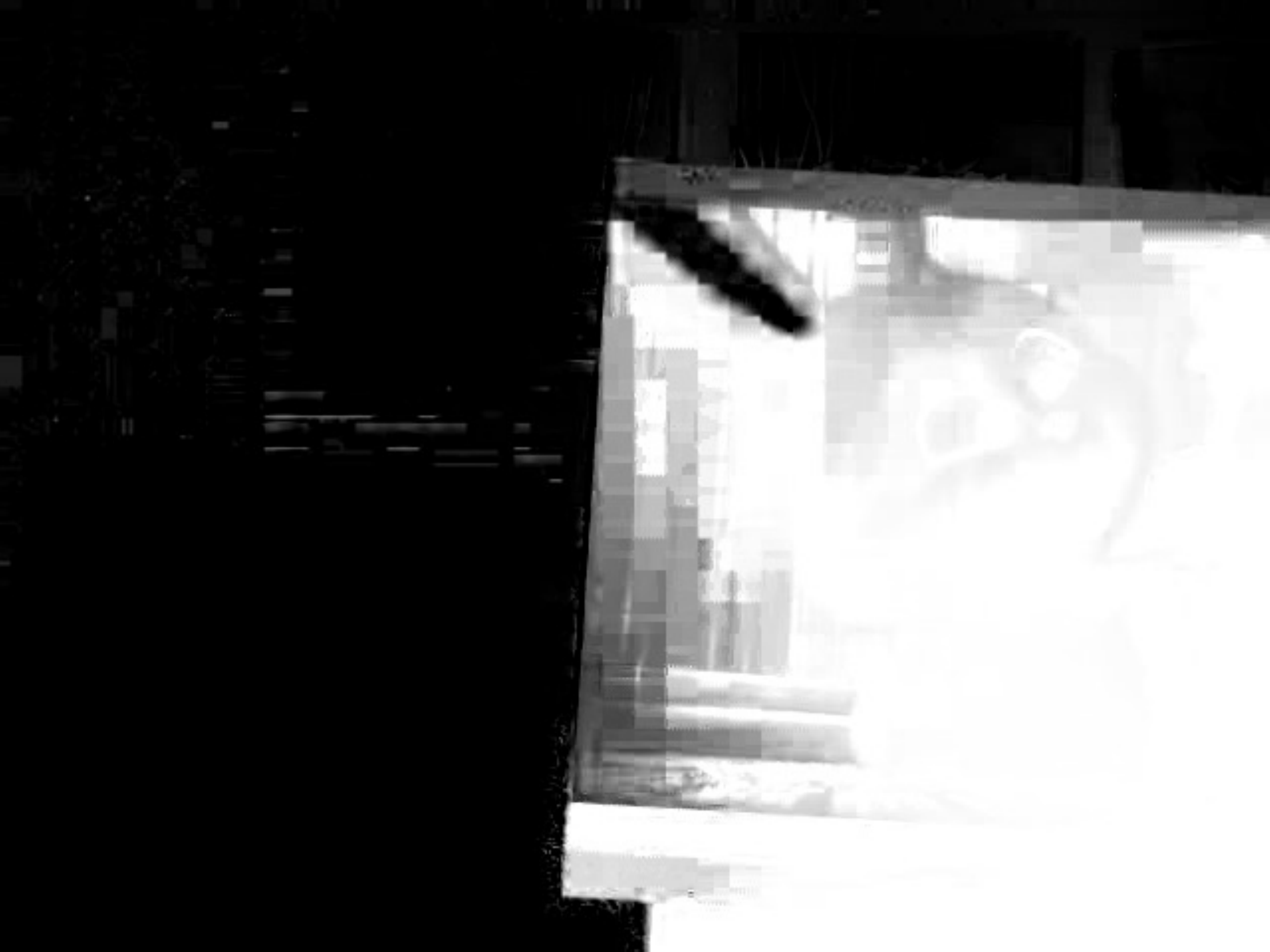}}
	\subfigure[]{\label{Fig8.4} \includegraphics[width=0.25\textwidth]{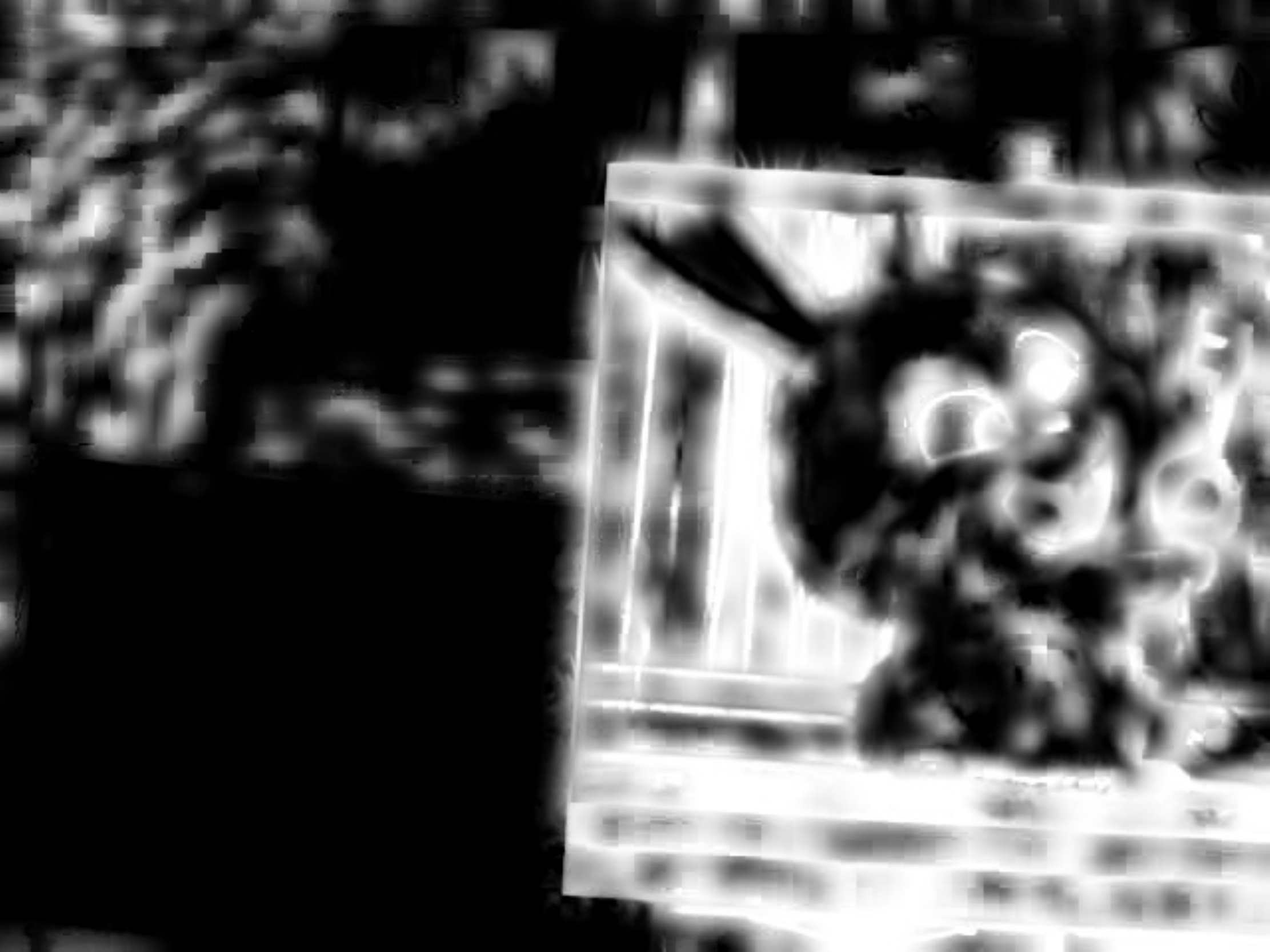}}
	\caption{Images of one test scene (a-d) and weight maps of different multi-focus image fusion methods (e-i). This is the fifth scene in Fig. S2 (Supplementary Material). (a) Depth map of the scene. (b) Segmentation result of the depth map. (c) Multi-focus source image with the front object in best focus. (d) Multi-focus source image with the background objects in best focus. Weight maps generated by (e) our proposed method, (f) DCNN, (g) DSIFT, (h) IM, and (i) GF. These weight maps are also shown in the fifth group of weight maps in Fig. S3 (Supplementary Material).}
	\label{Fig7}
\end{figure*}

To fuse multi-focus source images shown in Fig. \ref{Fig7.3} and Fig. \ref{Fig7.4}, the weight maps generated by our proposed method, DCNN, DSIFT, IM, and GF are shown in Fig. \ref{Fig7}. The weight maps of other test scenes can be found in Supplementary Material. This scene only contains two regions, the front region and the background region. During image capturing, the distance from the front region and the background region was set to be sufficiently large to ensure that when one region is in focus, the other region is defocused. Fig. \ref{Fig8.1} shows that the white front region and black background region are completely separated; the weight values in the front region are all ones and the weight values in background region are all zeros, accurately reflecting the sharpness level of this scene. However, in Fig. \ref{Fig8.5} to Fig. \ref{Fig8.4}, it can be seen that none of the DCNN, DSIFT, IM, and GF methods was able to generate a weight map as clean as the weight map generated by our proposed method because they rely on the color information of the multi-focus source images for computing weight maps, which is susceptible to lighting, noise and the texture of objects. Differently, our proposed method circumvent these limitations by making use of the depth map to directly determine weight maps.

The time consumption for constructing an all-in-focus image using our proposed method and other multi-focus image fusion algorithms was also quantified and compared. The sizes of the multi-focus source images and depth maps were $640 \times 480$. Tests were conducted on a computer with a 4 GHz CPU and 32 GB of RAM. The time consumption of our proposed method reported in TABLE \ref{T5} includes preprocessing the depth map, segmenting the depth map, and selecting in-focus images from multi-focus source images to construct an all-in-focus image. Our method took 33 $ms$ on average to construct an all-in-focus image, among which preprocessing the depth holes costed 5 $ms$, segmenting the depth map costed 27.5 $ms$, and selecting in-focus images to construct the all-in-focus image costed 0.5 $ms$. The significantly lower time consumption of our method, compared to the RGB-based methods (see TABLE \ref{T5}) is due to the low computational complexity stemming from the assistance of the depth map. Note that in practice, there are usually more than two multi-focus source images to be used to construct an all-in-focus image of a scene, and in accordance, the time consumption of other multi-focus image fusion methods increases linearly. Differently, for the proposed all-in-focus imaging method, since the time cost of preprocessing and segmenting the depth map is linear to the size of the depth map \cite{Felzenszwalb2004Efficient}, as long as the size of the depth map from the depth sensor is fixed, the time cost of preprocessing and segmenting the depth map stays constant. Although the time cost of selecting in-focus images is linear to the number of multi-focus source images, due to its low computational complexity, the time consumption of our proposed method does not increase significantly when the number of multiple multi-focus source images becomes higher.

The proposed multi-focus image fusion method is highly dependent on the depth map from the depth sensor. Presently, the range of the Kinect depth sensor is limited to 0.5 $m$-5 $m$. However, the proposed method is not dependent on a specific depth sensor. For instance, the ZED stereo camera has a significantly larger operating range (0.5 $m$ – 20 $m$) and can obtain depth maps with a size up to 4416 $times$ 1242 at 15 fps. In Supplementary Material, Fig. S1 shows the use of the ZED stereo camera for obtaining the depth map of more complex nature scenes.

\begin{table*}[]
	\centering
	\caption{Quantitative assessments of the proposed all-in-focus imaging method and other existing multi-focus image fusion methods. Parentheses denote the scores of a method when compared with other six methods. The higher the score, the better the method. Eight is the highest score, and one is the lowest score.}
	\begin{tabular}{cccccccccc}
		\toprule
		\toprule
		Scenes&Metrics&Methods& & & & & &  \\
		\cline{3-10}
		& &DWT&NSCT&IM&GF&NSCT-PCNN&DSIFT&DCNN&Ours \\
		\hline		
		1&${Q_{MI}}$	&1.1478(2)	&1.0451(1)	&1.3869(5)	&1.3402(4)	&1.3372(3)	&1.4235(8) &1.3903(6)	&1.4201(7) \\
		&${Q_{NCIE}}$	&0.8463(2)	&0.8408(1)	&0.8629(4)	&0.8597(3)	&0.8646(6)	&0.8681(8) &0.8635(5)	&0.8653(7) \\
		&${Q_{G}}$		&0.6694(3)	&0.4408(1)	&0.6998(5)	&0.6946(4)	&0.6421(2)	&0.7079(6) &0.7094(7)	&0.7153(8) \\
		&${Q_{P}}$		&0.8344(2)	&0.7255(1)	&0.9129(8)	&0.9023(4)	&0.8516(3)	&0.9049(5) &0.9112(7)	&0.9099(6) \\
		&${Q_{Y}}$		&0.8992(2)	&0.7262(1)	&0.9548(5)	&0.9412(4)	&0.9275(3)	&0.9710(6) &0.9721(7)	&0.9766(8) \\
		&${Q_{CB}}$		&0.7372(2)	&0.6935(1)	&0.7688(5)	&0.7634(4)	&0.7977(8)	&0.7575(3) &0.7708(6)	&0.7742(7) \\
		\bottomrule
		2&${Q_{MI}}$	&0.9504(2)	&0.8125(1)	&1.2323(7)	&1.1674(4)	&1.0457(3)	&1.2308(6) &1.2250(5)	&1.2504(8) \\
		&${Q_{NCIE}}$	&0.8308(2)	&0.8250(1)	&0.8480(7)	&0.8426(4)	&0.8374(3)	&0.8468(6) &0.8465(5)	&0.8489(8) \\
		&${Q_{G}}$		&0.6387(3)	&0.3889(1)	&0.6855(6)	&0.6747(4)	&0.5777(2)	&0.6834(5) &0.6879(7)	&0.6954(8) \\
		&${Q_{P}}$		&0.8273(3)	&0.6922(1)	&0.9159(5)	&0.9175(6)	&0.8269(2)	&0.9141(4) &0.9206(8)	&0.9191(7) \\
		&${Q_{Y}}$		&0.9012(3)	&0.6908(1)	&0.9655(6)	&0.9431(4)	&0.8976(2)	&0.9627(5) &0.9716(7)	&0.9832(8) \\
		&${Q_{CB}}$		&0.7231(2)	&0.6681(1)	&0.7856(6)	&0.7627(3)	&0.7744(4)	&0.7832(5) &0.7887(7)	&0.7977(8) \\
		\hline
		3&${Q_{MI}}$	&0.9101(2)	&0.8422(1)	&1.1820(5)	&1.1500(4)	&1.0052(3)	&1.2015(7) &1.1927(6)	&1.2089(8) \\
		&${Q_{NCIE}}$	&0.8284(2)	&0.8255(1)	&0.8437(5)	&0.8414(4)	&0.8344(3)	&0.8448(7) &0.8442(6)	&0.8454(8) \\
		&${Q_{G}}$		&0.6608(3)	&0.4649(1)	&0.7039(5)	&0.6998(4)	&0.5672(2)	&0.7079(6) &0.7099(7)	&0.7143(8) \\
		&${Q_{P}}$		&0.8266(3)	&0.7660(1)	&0.9070(5)	&0.9115(7)	&0.8053(2)	&0.9112(6) &0.9127(8)	&0.9033(4) \\
		&${Q_{Y}}$		&0.9151(3)	&0.7796(1)	&0.9742(5)	&0.9602(4)	&0.8834(2)	&0.9759(6) &0.9799{7}	&0.9825(8) \\
		&${Q_{CB}}$		&0.7059(2)	&0.6699(1)	&0.7816(5)	&0.7681(4)	&0.7169(3)	&0.7903(6) &0.7949(7)	&0.7954(8) \\
		\hline
		4&${Q_{MI}}$	&0.8384(2)	&0.7653(1)	&1.1384(5)	&1.0978(4)	&0.9426(3)	&1.1727(7) &1.1520(6)	&1.1828(8) \\
		&${Q_{NCIE}}$	&0.8249(2)	&0.8220(1)	&0.8408(5)	&0.8382(4)	&0.8310(3)	&0.8430(7) &0.8415(6)	&0.8439(8)  \\
		&${Q_{G}}$		&0.6269(3)	&0.4355(1)	&0.6738(5)	&0.6642(4)	&0.5434(2)	&0.6786(6) &0.6822(7)	&0.6886(8)  \\
		&${Q_{P}}$		&0.7967(3)	&0.7586(2)	&0.8972(4)	&0.9039(6)	&0.7443(1)	&0.9020(5) &0.9048(7)	&0.9067(8) \\
		&${Q_{Y}}$		&0.9047(3)	&0.7491(1)	&0.9692(5)	&0.9500(4)	&0.8729(2)	&0.9777(6) &0.9837(7)	&0.9890(8) \\
		&${Q_{CB}}$		&0.6908(2)	&0.6486(1)	&0.7713(5)	&0.7527(4)	&0.7075(3)	&0.7828(6) &0.7852(8)	&0.7834(7) \\
		\hline
		5(Fig. \ref{Fig7}) &${Q_{MI}}$	&0.9352(2)	&0.8659(1)	&1.1746(5)	&1.1420(4)	&0.9868(3)	&1.2248(7) &1.1968(6)	&1.2311(8) \\
		&${Q_{NCIE}}$	&0.8305(2)	&0.8276(1)	&0.8444(5)	&0.8435(4)	&0.8335(3)	&0.8481(7) &0.8465(6)	&0.8482(8) \\
		&${Q_{G}}$		&0.6432(3)	&0.4472(1)	&0.6720(5)	&0.6594(4)	&0.5506(2)	&0.6751(6) &0.6753(7)	&0.6885(8) \\
		&${Q_{P}}$		&0.8381(3)	&0.7649(1)	&0.9011(7)	&0.8953(4)	&0.7858(2)	&0.8973(5) &0.8984(6)	&0.9214(8) \\
		&${Q_{Y}}$		&0.9016(3)	&0.7483(1)	&0.9628(5)	&0.9419(4)	&0.8702(2)	&0.9698(6) &0.9769(7)	&0.9802(8) \\
		&${Q_{CB}}$ 	&0.7117(2)	&0.6785(1)	&0.7860(5)	&0.7607(4)	&0.7186(3)	&0.7966(7) &0.7964(6)	&0.8014(8) \\
		\bottomrule
		\bottomrule
	\end{tabular}
	\label{T3}
\end{table*}

\begin{table*}[]
	\centering
	\caption{Scores and ranking of the methods. }
	\begin{tabular}{c|cccccccc|cc}
		\toprule
		\toprule
		\diagbox{Methods}{Number of Times}{Scores}&8&7&6&5&4&3&2&1&Total Scores&Ranking \\
		\hline
		Ours&23&5&1&0&1&0&0&0&229&1	\\
		\hline
		DCNN&3&14&10&3&0&0&0&0&197&2 \\
		\hline
		DSIFT&2&7&13&6&1&1&0&0&180&3 \\
		\hline
		IM&1&3&3&21&2&0&0&0&160&4 \\
		\hline
		GF&0&1&2&0&25&2&0&0&125&5 \\
		\hline
		NSCT-PCNN&1&0&1&0&1&14&12&1&85&6 \\
		\hline
		DWT&0&0&0&0&0&0&13&17&73&7 \\
		\hline
		NSCT&0&0&0&0&0&0&1&29&31&8	\\
		\bottomrule
		\bottomrule
	\end{tabular}
	\label{T4}
\end{table*}

\begin{table*}[]
	\centering
	\caption{Running time (seconds) of the proposed method and existing multi-focus image
		fusion algorithms for the five test scenes.}
	\begin{tabular}{cccccccccc}
		\toprule
		\toprule
		Scenes&Methods& & & & & & & & \\
		\cline{2-10}
		&DWT&NSCT&IM&GF&NSCT-PCNN&DSIFT&DCNN&Our \\
		\hline	
		1&0.2054&35.7285&3.2084&0.3351&243.2443&8.8385&132.9873 &0.030 \\
		2&0.2031&35.5960&3.1097&0.3491&243.8029&11.4488&131.7024  &0.035 \\
		3&0.2061&35.7128&2.9816&0.3473&243.4221&7.6047&131.6626 &0.033 \\
		4&0.2039&35.7426&2.9719&0.3457&243.8831&7.3378&127.3014 &0.032 \\
		5 (Fig. \ref{Fig7})&0.2050&35.7939&2.9131&0.3452&243.1754&9.4629&132.2269 &0.035 \\
		\hline
		Average&0.2047&35.7148&3.0369&0.3445&243.5056&8.9385&131.1761 &0.033 \\
		\bottomrule
		\bottomrule
	\end{tabular}
	\label{T5}
\end{table*}

\section{Conclusion}\label{Section5}
This paper reported an efficient multi-focus image fusion method assisted by depth sensing. The depth map from a depth sensor was segmented with a modified graph-based segmentation algorithm. The segmented regions were used to guide a focus algorithm to locate an in-focus image for each region from among multi-focus images. The all-in-focus image was constructed by combining the in-focus images selected in each segmented region. The experimental results demonstrated the advantages of the proposed method by comparing the method with other algorithms in terms of six fusion metrics and time consumption. The proposed method enables the construction of an all-in-focus image within 33 $ms$ and provides a practical approach for constructing high-quality all-in-focus images that can potentially be used as reference images.

\section{Acknowledgment}
We would like to thank the authors of paper \cite{Liu2012Objective} having shared their image fusion evaluation toolbox.
This work was supported by the National Natural Science Foundation of China (grant numbers 61525305 and 61625304), the Shanghai Natural Science Foundation (grant numbers 17ZR1409700 and 18ZR1415300), and the basic research project of Shanghai Municipal Science and Technology Commission (grant number 16JC1400900).

\bibliographystyle{abbrv}
\bibliography{mybibfile}
\end{document}


\title{Supplementary Material}

\author{Hang Liu\footnotemark[1], Hengyu Li\footnotemark[1], Jun Luo\footnotemark[1], Shaorong Xie\footnotemark[1], Yu Sun\footnotemark[2]}

\maketitle
\thispagestyle{empty}

\footnotetext[1]{Hang Liu, Hengyu Li, Jun Luo, Shaorong Xie are with the School of Mechatronic Engineering and Automation, Shanghai University, China. Hengyu Li is the corresponding author. \href{scholar.hang@gmail.com}{scholar.hang@gmail.com} \href{lihengyu@shu.edu.cn}{lihengyu@shu.edu.cn} }
\footnotetext[2]{Yu Sun is with the Department of Mechanical and Industrial Engineering, University of
	Toronto, Canada, and the Shanghai University. \href{sun@mie.utoronto.ca}{sun@mie.utoronto.ca}}

\begin{abstract}
This document provides supplementary information to ``Construction of all-in-focus images assisted by depth sensing``.
\end{abstract}

\section{Detailed Definition of Six Multi-focus Image Fusion Metrics}
\cite{Liu2012Objective} proposed a comprehensive survey of 12 evaluation metrics for image fusion. The 12 popular fusion metrics are categorized into four classes, namely information theory-based metrics, image feature-based metrics, image structural similarity-based metrics, and human perception-based metrics. The MATLAB source code of 12 fusion metrics can be found at Liu's GitHub website \url{https://github.com/zhengliu6699}. We choose six of them covering all four categories to compare the proposed multi-focus image fusion method with other multi-focus image fusion algorithms. For convenience, the detailed definition of  six metrics from \cite{Liu2012Objective} are introduced as follows. Uniformly, Let $A(i,j)$ and $B(i,j)$ denote two source images while $F(i,j)$ the fused image. 
\subsection{Normalized Mutual Information (${Q_{MI}}$)}
$Q_{MI}$ is an information theory-based metric that measures the amount of information in the fused image inherited from the source images. $Q_{MI}$ is defined as
\begin{equation}
{Q_{MI}} = 2\left[ {\frac{{MI(A,F)}}{{H(A) + H(F)}} + \frac{{MI(B,F)}}{{H(B) + H(F)}}} \right], \tag{S1}
\end{equation}
where $H(X)$ is the entropy of image $X$ and $MI(X,Y)$ is the mutual information between image $X$ and $Y$,
\begin{equation}
\begin{array}{l}
MI(X,Y) = H(X) + H(Y) - H(X,Y),\\
H(X) =  - \sum\limits_x {p(x){{\log }_2}p(x)} ,\\
H(Y) =  - \sum\limits_y {p(y){{\log }_2}p(y)} ,\\
H(X,Y) =  - \sum\limits_{x,y} {p(x,y){{\log }_2}p(x,y)} , \tag{S2}
\end{array}
\end{equation}
where $p(x,y)$ is the joint probability distribution function of $X$ and $Y$, and $p(x)$ and $p(y)$ are the marginal probability distribution function of $X$ and $Y$.
\subsection{Normalized Mutual Information (${Q_{NCIE}}$)}
For two discrete variables $U = {\{ {u_i}\} _{1 \le i \le N}}$ and $V = {\{ {v_i}\} _{1 \le i \le N}}$, the nonlinear correlation coefficient (NCC) is defined as
\begin{equation}
NCC(U,V) = H'(U) + H'(V) - H'(U,V). \tag{S3}
\end{equation}
Considering $NCC(A,B)$ for images $A$ and $B$, the entropies are defined as
\begin{equation}
\begin{array}{l}
H'(A,B) =  - \sum\limits_{i = 1}^b {\sum\limits_{j = 1}^b {{h_{AB}}(i,j){{\log }_b}{h_{AB}}(i,j)} } ,\\
H'(A) =  - \sum\limits_{i = 1}^b {{h_A}(i){{\log }_b}{h_A}(i)} ,\\
H'(B) =  - \sum\limits_{i = 1}^b {{h_B}(i){{\log }_b}{h_B}(i)} , \tag{S4}
\end{array}
\end{equation}
where $b$ is determined by the intensity level, i.e., b = 256. A nonlinear correlation matrix of the input image $A(i,j)$, $B(i,j)$, and fused image $F(i,j)$ is defined as
\begin{equation}
\begin{array}{l}
R = \left( {\begin{array}{*{20}{c}}
	{NC{C_{AA}}}&{NC{C_{AB}}}&{NC{C_{AF}}}\\
	{NC{C_{BA}}}&{NC{C_{BB}}}&{NC{C_{BF}}}\\
	{NC{C_{FA}}}&{NC{C_{FB}}}&{NC{C_{FF}}}
	\end{array}} \right)\\
\;\;\; = \left( {\begin{array}{*{20}{c}}
	1&{NC{C_{AB}}}&{NC{C_{AF}}}\\
	{NC{C_{BA}}}&1&{NC{C_{BF}}}\\
	{NC{C_{FA}}}&{NC{C_{FB}}}&1
	\end{array}} \right). \tag{S5}
\end{array}
\end{equation}  
The eigenvalue of the nonlinear correlation matrix $R$ is ${\lambda _i}~(i=1,2,3)$. Therefore, the nonlinear correlation information entropy ${Q_{NCIE}}$ can be obtained :
\begin{equation}
{Q_{NCIE}} = 1 + \sum\limits_{i = 1}^3 {\frac{{{\lambda _i}}}{3}} {\log _b}\frac{{{\lambda _i}}}{3}. \tag{S6}
\end{equation}  
\subsection{Gradient-Based Fusion Performance (${Q_{G}}$)}
Xydeas and Petrovic proposed a metric to evaluate the amount of edge information, which is transferred from input images to the fused image \cite{xydeas2000objective}. A Sobel edge operator is applied to get the edge strength of input image $A(i,j)$, ${g_A}(i,j)$, and orientation ${\alpha _A}(i,j)$:
\begin{equation}
\begin{array}{l}
{g_A}(i,j) = \sqrt {S_A^x{{(i,j)}^2} + S_A^y{{(i,j)}^2}} ,\\
{\alpha _A}(i,j) = {\tan ^{ - 1}}(\frac{{S_A^x(i,j)}}{{S_A^y(i,j)}}),
\end{array} \tag{S7}
\end{equation}
where $S_A^x{{(i,j)}}$ and $S_A^y{{(i,j)}}$ are the convolved results with the horizontal and vertical Sobel templates. The relative strength ?($G^{AF}$) and orientation values ($\Delta ^{AF}$) between input image $A$ and fused image $F$ are
\begin{equation}
\begin{array}{l}
{G^{AF}}(i,j) = \left\{ {\begin{array}{*{20}{c}}
	{\frac{{{g_F}(i,j)}}{{{g_A}(i,j)}},\;\;\;{g_A}(i,j) > {g_F}(i,j),}\\
	{\frac{{{g_A}(i,j)}}{{{g_F}(i,j)}},\;\;\;Otherwise,\;\;\;\;\;\;\;\;\;\;}
	\end{array}} \right.\\
{\Delta ^{AF}}(i,j) = 1 - \frac{{\left| {{\alpha _A}(i,j) - {\alpha _F}(i,j)} \right|}}{{\pi /2}}. \tag{S8}
\end{array}
\end{equation}
The edge strength and orientation preservation values can be derived:
\begin{equation}
\begin{array}{l}
Q_g^{AF}(i,j) = \frac{{{\Gamma _g}}}{{1 + {e^{{K_g}({G^{AF}}(i,j) - {\sigma _g})}}}},\\
Q_\alpha ^{AF}(i,j) = \frac{{{\Gamma _\alpha }}}{{1 + {e^{{K_\alpha }({G^{AF}}(i,j) - {\sigma _\alpha })}}}}.
\end{array} \tag{S9}
\end{equation}
The constants ${\Gamma _g}$, ${K_g}$, $\sigma _g$ and  ${\Gamma _\alpha}$, ${K_\alpha}$, $\sigma _\alpha$ determine the shape of the sigmoid functions used to form the edge strength and orientation preservation value. Edge information preservation value is then defined as
\begin{equation}
{Q^{AF}}(i,j) = Q_g^{AF}(i,j)Q_\alpha ^{AF}(i,j). \tag{S10}
\end{equation}
The final assessment is obtained from the weighted average of the edge information preservation values.
\begin{equation}
{Q_G} = \frac{{\sum\nolimits_{n = 1}^N {\sum\nolimits_{m = 1}^M {\left[ {{Q^{AF}}(i,j){\omega ^A}(i,j) + {Q^{BF}}(i,j){\omega ^B}(i,j)} \right]} } }}{{\sum\nolimits_{n = 1}^N {\sum\nolimits_{m = 1}^M {({\omega ^A}(i,j) + {\omega ^B}(i,j))} } }} \tag{S11}
\end{equation}
\subsection{Image Fusion Metric-Based on Phase Congruency (${Q_{P}}$)}
Zhao et al. and Liu et al. used the phase congruency, which provides an absolute measure of image feature, to define an evaluation metric \cite{zhao2007performance, liu2008feature}. In \cite{zhao2007performance}, the principal (maximum and minimum) moments of the image phase congruency were employed to define the metric because the moments contain the information for corners and edges. The metric is defined as a product of three correlation coefficients,  
\begin{equation}
{Q_P} = {({P_p})^\alpha }{({P_M})^\beta }{({P_m})^\gamma } \tag{S12}
\end{equation} 
where $p$, $M$, $m$ refers to phase congruency ($p$), maximum, and maximum moments, respectively, and there are 
\begin{equation}
\begin{array}{l}
{P_P} = \max (C_{AF}^p,C_{BF}^p,C_{SF}^p),\\
{P_M} = \max (C_{AF}^M,C_{BF}^M,C_{SF}^M),\\
{P_m} = \max (C_{AF}^m,C_{BF}^m,C_{SF}^m). \tag{S13}
\end{array}
\end{equation}
Herein, $C_{xy}^k$, ${k|p,M,m}$ stands for the correlation coefficients between two sets $x$ and $y$:
\begin{equation}
\begin{array}{l}
C_{xy}^k = \frac{{\sigma _{xy}^k + C}}{{\sigma _x^k\sigma _y^k + C}},\\
{\sigma _{xy}} = \frac{1}{{N - 1}}\sum\limits_{i = 1}^N {({x_i} - \mathop {\bar x}\limits^{} )} ({y_i} - \mathop {\bar x}\limits^{} ). \tag{S14}
\end{array}
\end{equation}
The suffixes $A$, $B$, $F$, and $S$ correspond to the two inputs, fused image, and maximum-select map. The exponential parameters $\alpha$, $\beta$, and $\gamma$ can be adjusted based on the importance of the three components \cite{zhao2007performance}.
\subsection{Yang's Metric (${Q_{Y}}$)}
Yang et al. proposed another way to use SSIM for fusion assessment\cite{yang2008novel}:
\begin{equation}
{Q_Y} = \left\{ \begin{array}{l}
\lambda (w)SSIM(A,F|w) + (1 - \lambda (w))SSIM(B,F|w),\\
\;\;\;\;\;\;\;\;\;\;\;\;\;\;\;\;\;\;\;\;\;\;\;\;\;\;\;\;\;\;\;\;\;\;\;\;\;SSIM(A,B|w) \ge 0.75,\\
\max \{ SSIM(A,F|w),SSIM(B,F|w)\} ,\\
\;\;\;\;\;\;\;\;\;\;\;\;\;\;\;\;\;\;\;\;\;\;\;\;\;\;\;\;\;\;\;\;\;\;\;\;\;SSIM(A,B|w) < 0.75.\;\;\;\;\;\;\;\;\;\;\;\;\;\;\;\;\;\;\;\;\;\;\;\;\;\;\;\;\;\;\;\;\;\;\;\;\;\;\;
\end{array} \right. \tag{S15}
\end{equation}
The definition of local weight $\lambda (w)$ is defined as
\begin{equation}
\lambda (\omega ) = \frac{{s(A|\omega )}}{{s(A|\omega ) + s(B|\omega )}}, \tag{S16}
\end{equation}
herein, $s(A|\omega )$ is a local measure of image salience.
\subsection{Chen-Blum Metric (${Q_{CB}}$)}
There are five steps involved in Chen-Blum metric \cite{chen2009new}:
\begin{enumerate}
	\item Contrast sensitivity filtering: Filtering is implemented in the frequency domain. Image ${I_A}(i,j)$ is transformed into the frequency domain  and get ${I_A}(m,n)$. The filtered image is obtained: $\widetilde {{I_A}}(m,n) = {I_A}(m,n)S(r)$, where $S(r)$ is the CSF filter in polar form with 
	$r = \sqrt {{m^2} + {n^2}}$. In \cite{chen2009new}, there are three choices suggested for CSF, which include Mannos-Sakrison, Barton, and DoG filter.
	\item Local contrast computing: Peli's contrast is defined as
	\begin{equation}
	C(i,j) = \frac{{{\phi _k}(i,j) * I(i,j)}}{{{\phi _{k + 1}}(i,j)*I(i,j)}} - 1. \tag{S17}
	\end{equation}
	A common choice for ${\phi _k}$ would be 
	\begin{equation}
	{G_k}(x,y) = \frac{1}{{\sqrt {2\pi } {\sigma _k}}}{e^{\frac{{{x^2} + {y^2}}}{{2\sigma _k^2}}}}, \tag{S18}
	\end{equation}
	with a standard deviation ${\sigma _k}=2$. 
	\item Contrast preservation calculation: The masked contrast map for input image ${I_A}(i,j)$ is calculated as 
	\begin{equation}
	{{C'}_A} = \frac{{t{{({C_A})}^p}}}{{h{{({C_A})}^q} + Z}}. \tag{S19}
	\end{equation}
	Here, $t$, $h$, $p$, $q$ and $Z$ are real scalar parameters that determine the shape of the nonlinearity of the masking function \cite{chen2009new}.
	\item Saliency map generation: The saliency map for ${I_A}(i,j)$ is defined as
	\begin{equation}
	{\lambda _A}(i,j) = \frac{{{{C'}_A}^2(i,j)}}{{{{C'}_A}^2(i,j) + {{C'}_B}^2(i,j)}}. \tag{S20}
	\end{equation}
	The information preservation value is computed as
	\begin{equation}
	{Q_{AF}}(i,j) = \left\{ {\begin{array}{*{20}{c}}
		{\frac{{{{C'}_A}(i,j)}}{{{{C'}_F}(i,j)}},\;\;\;if\;{{C'}_A}(i,j) < {{C'}_A}(i,j),}\\
		{\frac{{{{C'}_A}(i,j)}}{{{{C'}_F}(i,j)}},\;\;\;otherwise\;\;\;\;\;\;\;\;\;\;\;\;\;\;\;}
		\end{array}} \right. \tag{S21}
	\end{equation}
	\item Global quality map:
	\begin{equation}
	{Q_{GQM}}({\rm{i}},j) = {\lambda _A}(i,j){Q_{AF}}(i,j) + {\lambda _B}(i,j){Q_{BF}}(i,j). \tag{S22}
	\end{equation}	
\end{enumerate}
The metric value is obtained by average the global quality map, i.e., ${Q_{CB}} = \overline {{Q_{GQM}}({\rm{i}},j)}$.

\section{Website of Compared Multi-focus Image Fusion Algorithms}
In the experiment, the DWT method is implemented based on O. Rockinger's image fusion toolbox \url{http://www.metapix.de/toolbox.htm}. The NSCT method is implemented based on the Nonsubsampled Contourlet Toolbox in MATLAB Central (\url{http://cn.mathworks.com/matlabcentral/fileexchange/10049-nonsubsampled-contourlet-toolbox?s_tid=srchtitle}). The NSCT-PCNN method is implemented using the code downloaded from Xiaobo Qu's homepage (\url{http://www.quxiaobo.org/index.html}) and the codes of IM and GF methods are available on Xudong Kang's homepage (\url{http://xudongkang.weebly.com/index.html}), the codes of DSIFT and DCNN methods are available on Yu Liu's homepage (http://www.escience.cn/people/liuyu1/Codes.html).
\section{Source Images}
In our experiments, 5 pairs of multi-focus images that belong to different scenes shown in Fig.\ref{FigS2} are utilized to compare the proposed multi-focus image fusion method and other multi-focus image fusion algorithms. The source images can be downloaded from the author's GitHub website (https://github.com/robotVisionHang).
\begin{table*}[h]
	\small\centering
	\renewcommand{\thetable}{S\arabic{table}}
	\caption{The quantitative assessments of the GF method reconstructs a fused image using two different base and detail layers $(GF\_DIFF)$ and the same detail layer $(GF\_SAME)$.}
	\begin{tabular}{c|c|cccccc}
		\toprule
		\toprule
		\multirow{2}*{Scenes} & \multirow{2}*{Methods} & \multicolumn{6}{c}{Metrics} \\
		\cline{3-8}
		&&MI&NCIE&G&P&Y&CB \\
		\hline	
		\multirow{2}*{1}&GF\_DIFF&1.3402&0.8597&0.6946&0.9023&0.9412&0.7634 \\
		&GF\_SAME&1.3625&0.8615&0.7031&0.9024&0.9499&0.7657 \\
		\hline
		\multirow{2}*{2}&GF\_DIFF&1.1674&0.8426&0.6747&0.9175&0.9431&0.7627 \\
		&GF\_SAME&1.1794&0.8434&0.6828&0.9177&0.9488&0.7647 \\
		\hline
		\multirow{2}*{3}&GF\_DIFF&1.1500&0.8414&0.6998&0.9115&0.9602&0.7681 \\
		&GF\_SAME&1.1645&0.8423&0.7044&0.9116&0.9650&0.7709 \\
		\hline
		\multirow{2}*{4}&GF\_DIFF&1.0978&0.8382&0.6642&0.9039&0.9500&0.7527 \\
		&GF\_SAME&1.1260&0.8400&0.6741&0.9039&0.9575&0.7662 \\
		\hline
		\multirow{2}*{5}&GF\_DIFF&1.1420&0.8435&0.6594&0.8953&0.9419&0.7607 \\
		&GF\_SAME&1.1602&0.8447&0.6659&0.8955&0.9462&0.7638 \\
		\bottomrule
		\bottomrule
	\end{tabular}\\
	\label{T1}
\end{table*}

\begin{figure*}[h]
	\centering
	\subfigure[]{\label{Figs1.1} \includegraphics[width=0.325\textwidth]{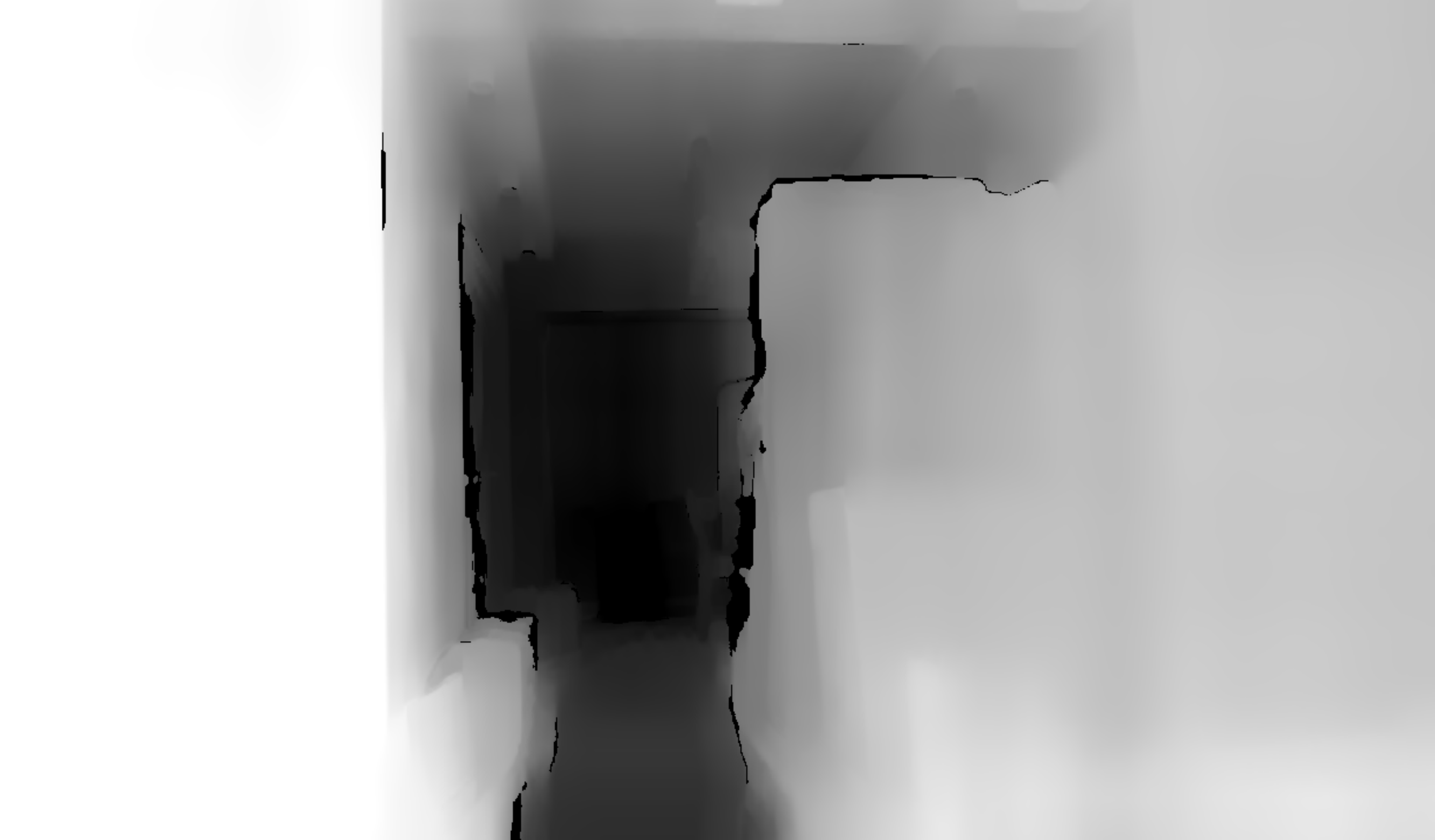}} 
	\subfigure[]{\label{Figs1.1} \includegraphics[width=0.325\textwidth]{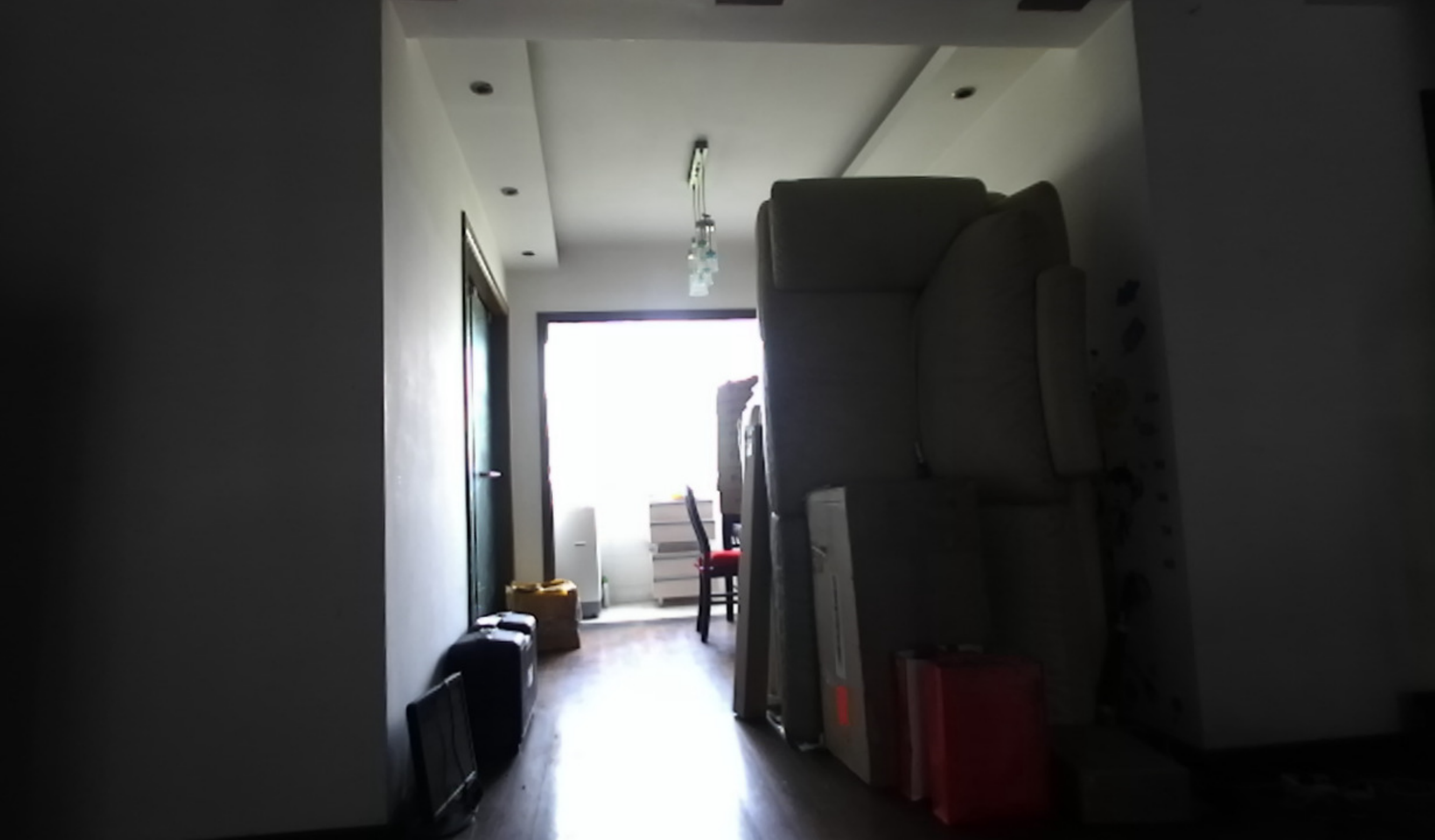}}
	\subfigure[]{\label{Figs1.1} \includegraphics[width=0.325\textwidth]{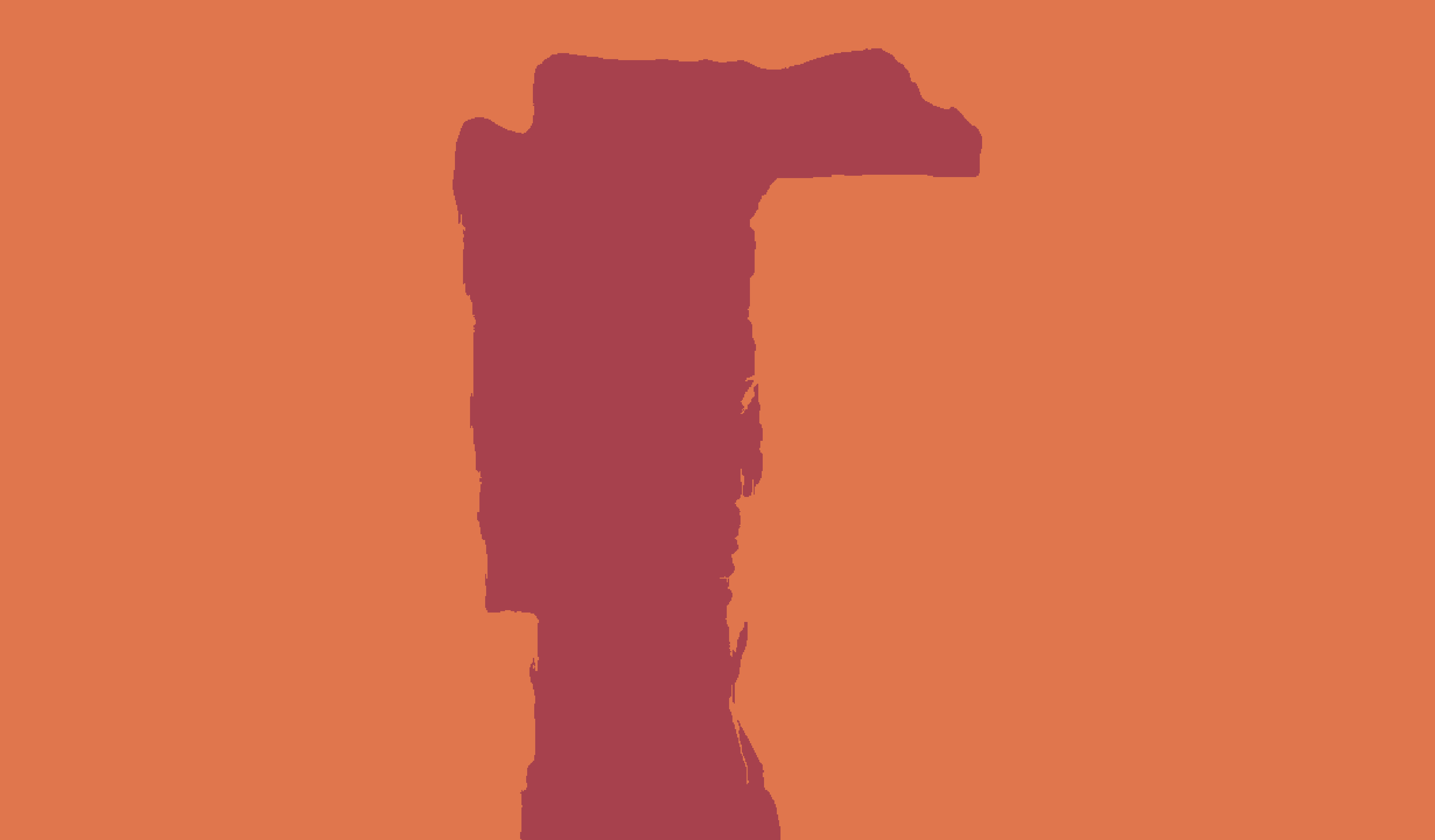}}
	\subfigure[]{\label{Figs1.1} \includegraphics[width=0.325\textwidth]{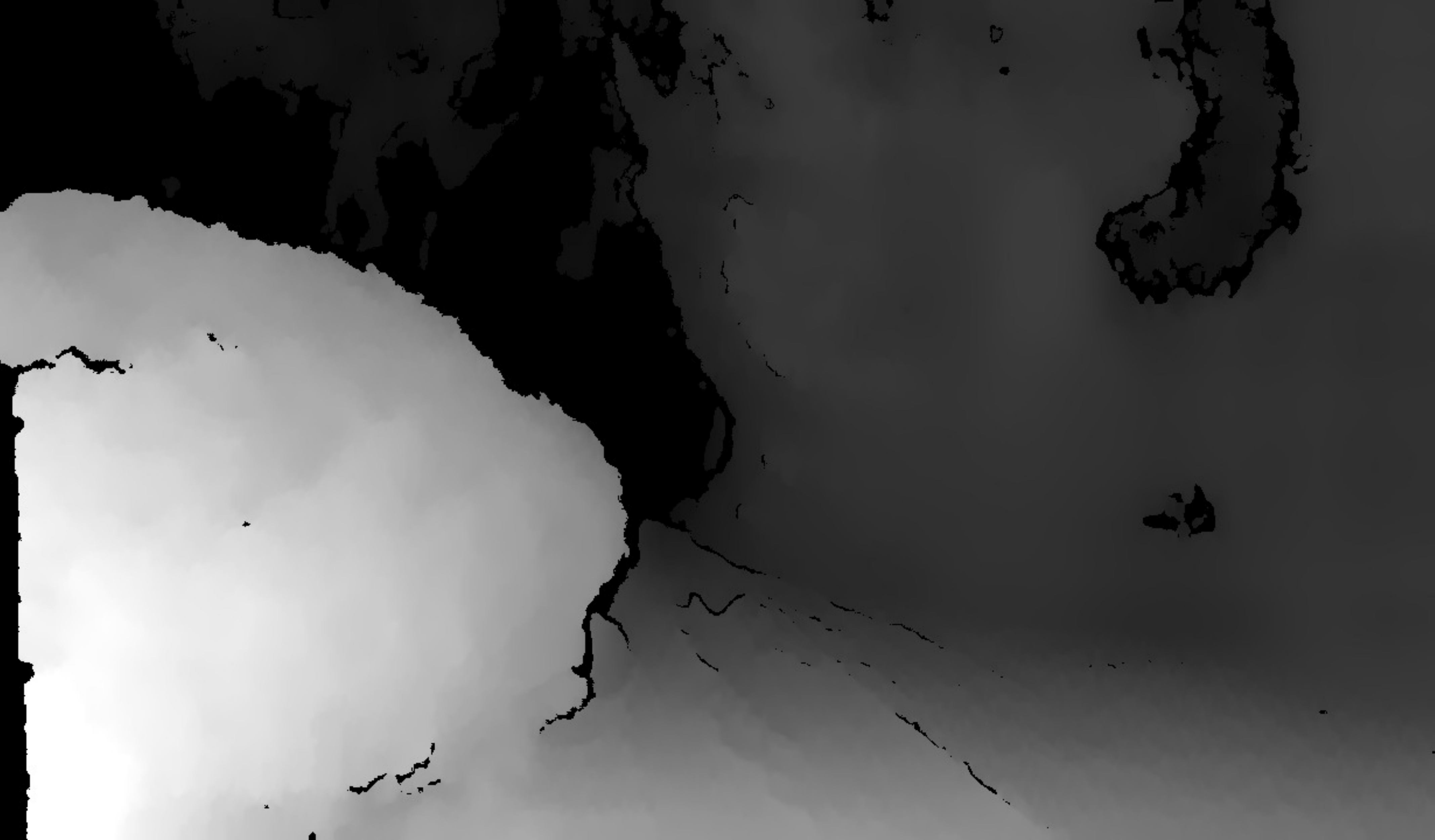}} 
	\subfigure[]{\label{Figs1.1} \includegraphics[width=0.325\textwidth]{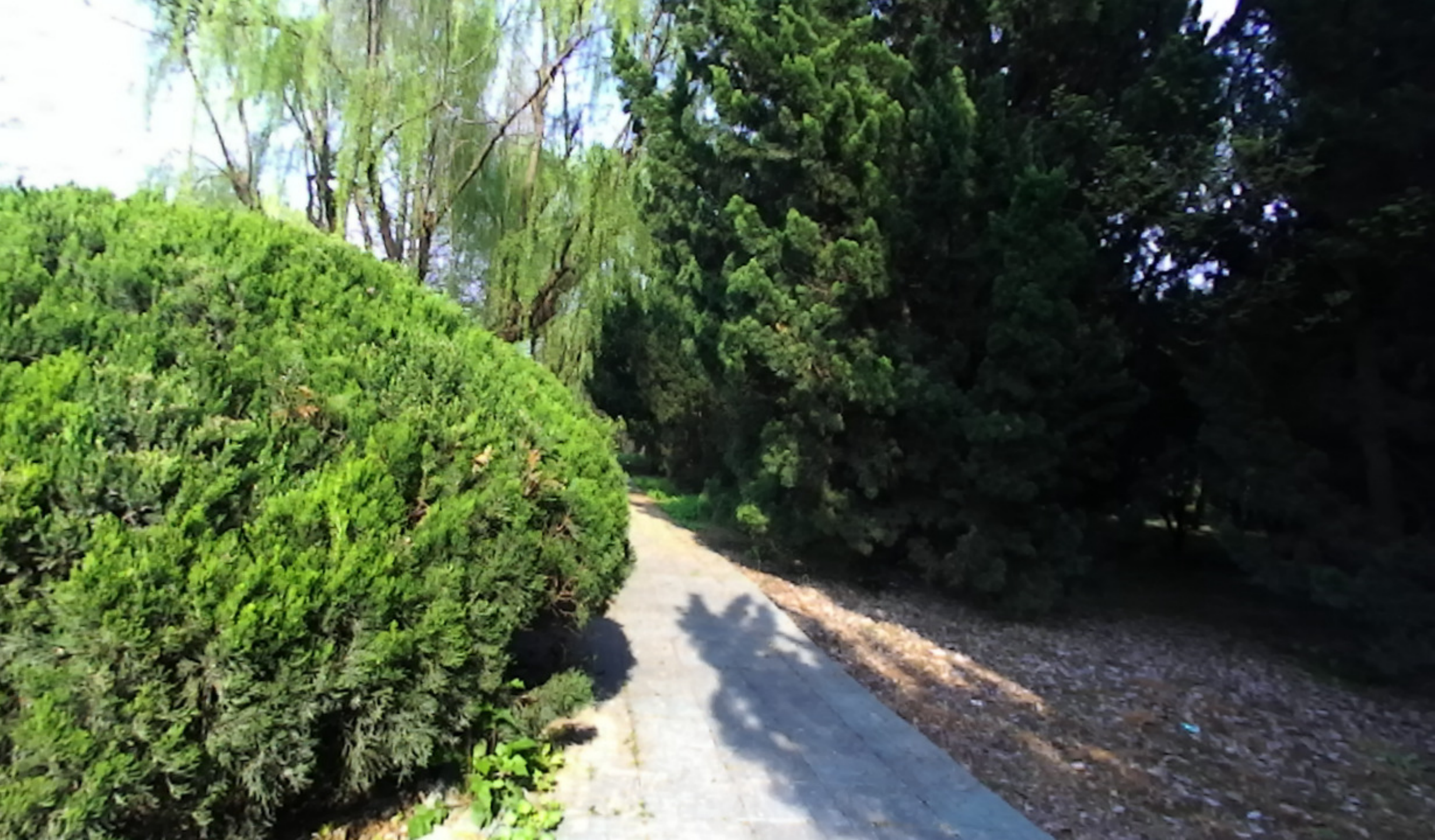}}
	\subfigure[]{\label{Figs1.1} \includegraphics[width=0.325\textwidth]{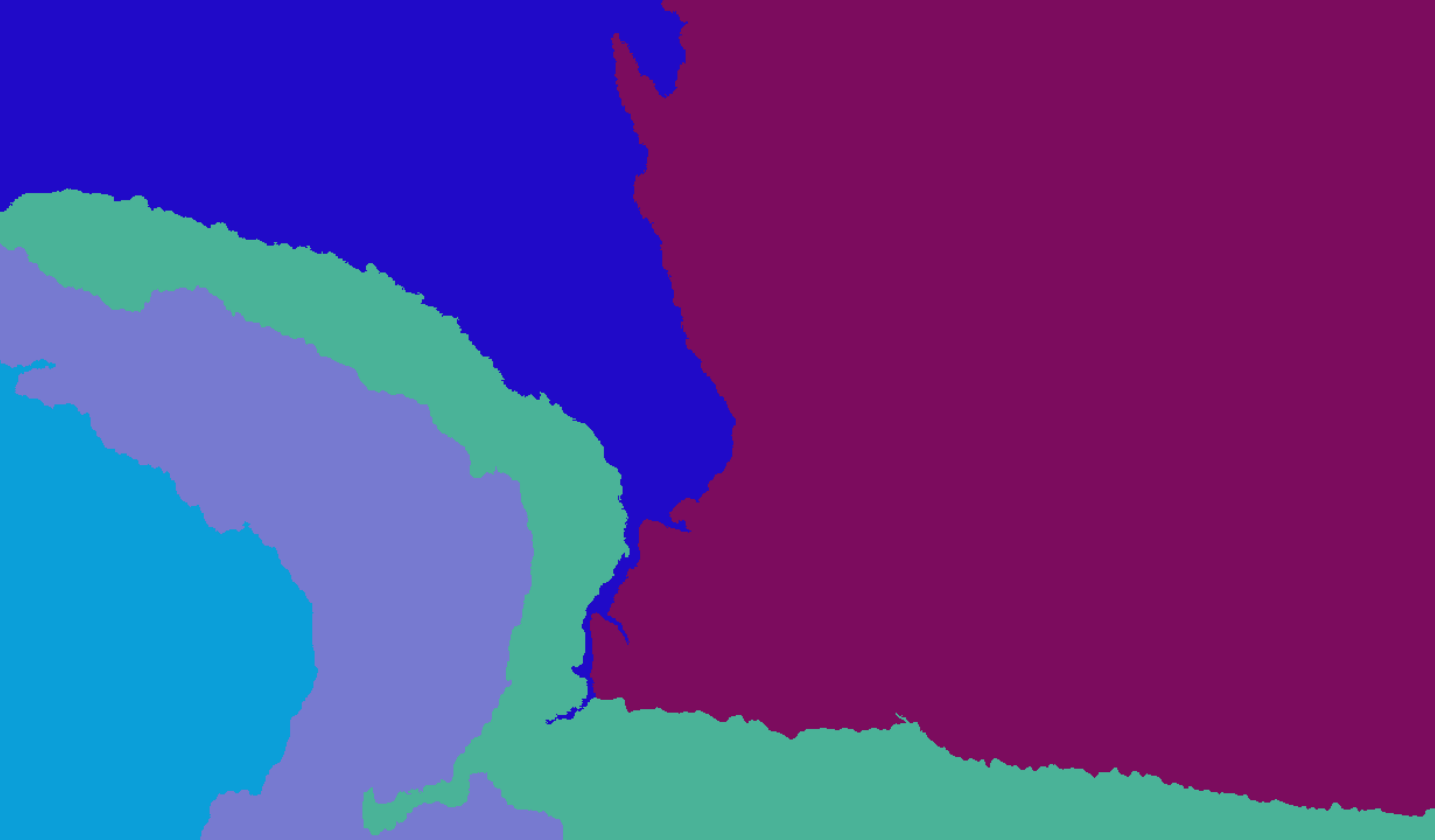}}
	\subfigure[]{\label{Figs1.1} \includegraphics[width=0.325\textwidth]{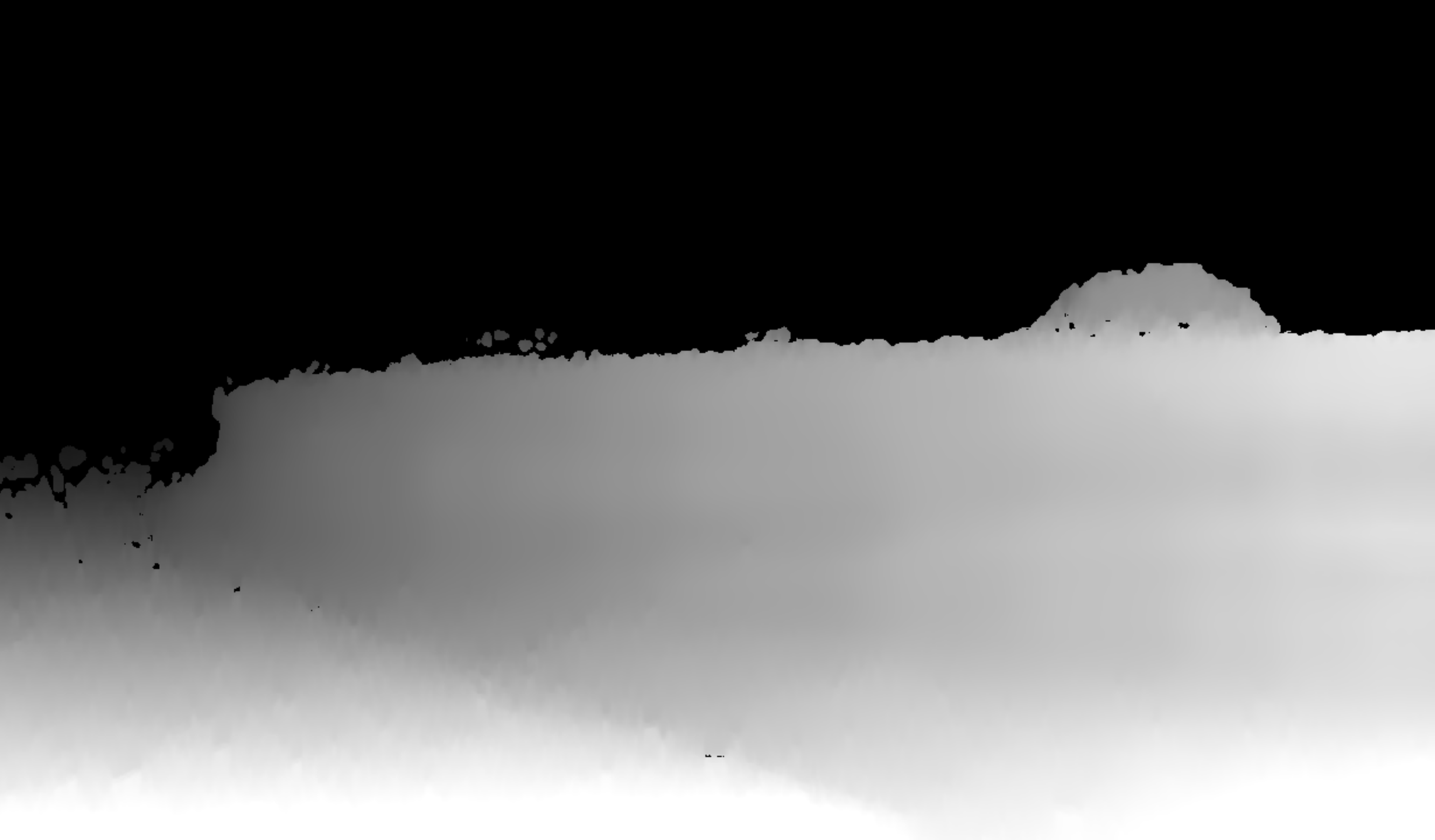}} 
	\subfigure[]{\label{Figs1.1} \includegraphics[width=0.325\textwidth]{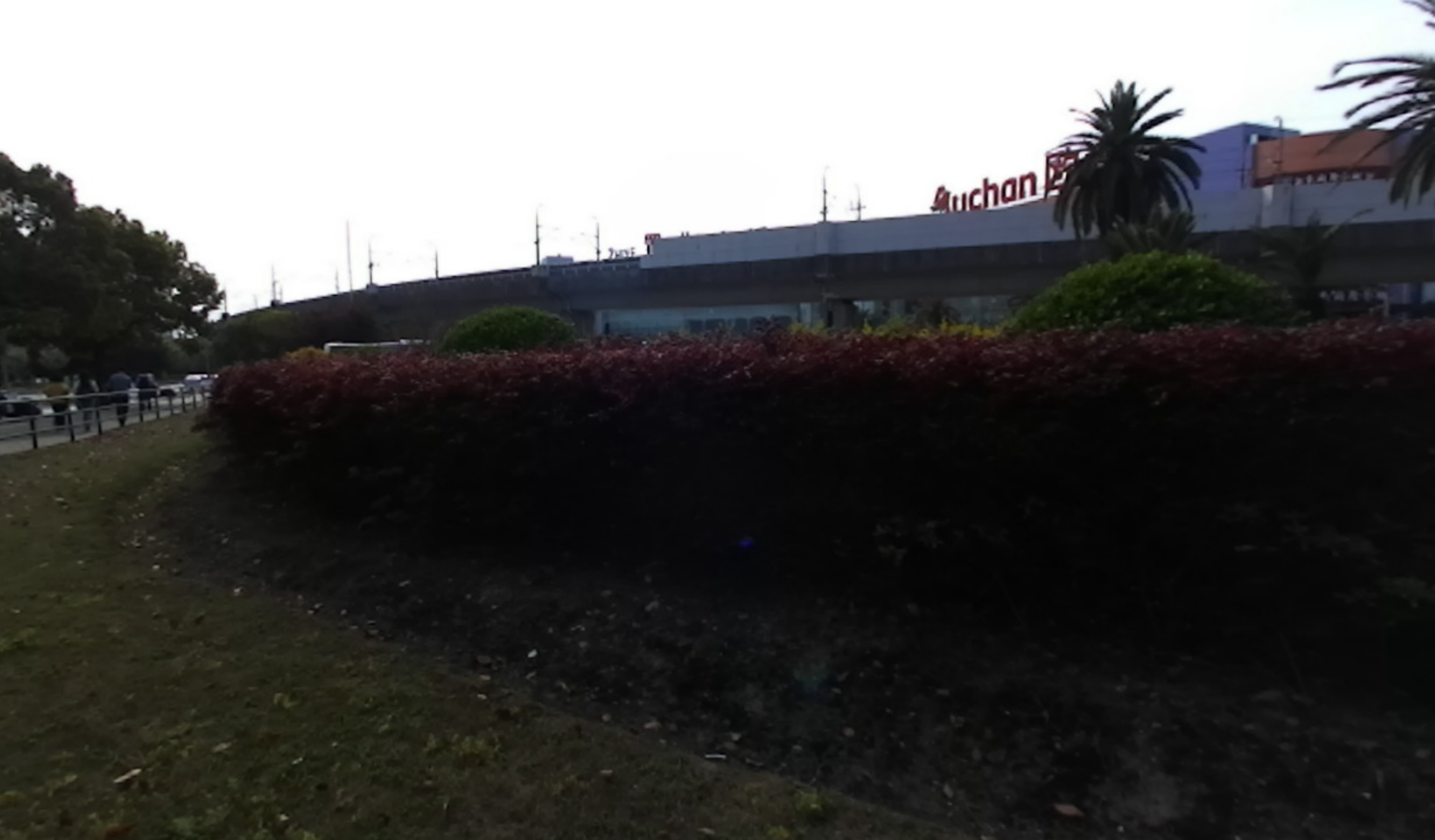}}
	\subfigure[]{\label{Figs1.1} \includegraphics[width=0.325\textwidth]{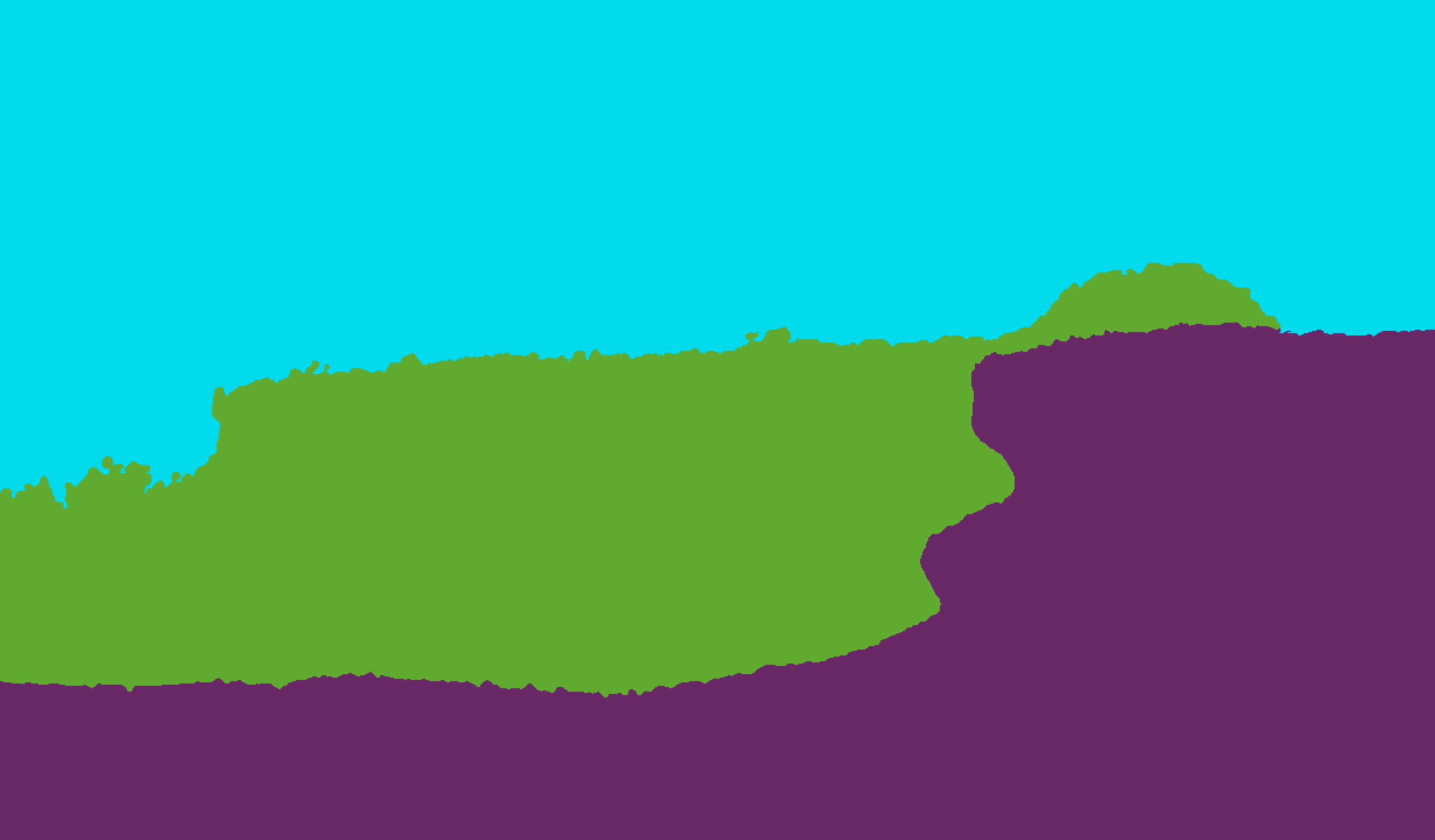}}
	\makeatletter 
	\renewcommand{\thefigure}{S\@arabic\c@figure}
	\makeatother
	\caption{(a), (d), (g) Depth maps obtained via the use of a ZED stereo camera. (b), (e), (h) Corresponding color images. (c),(f),(i) In-focus image block regions determined by segmenting depth maps.}
	\label{FigS1}
\end{figure*}

\begin{figure*}[h]
	\centering
	\subfigure[]{\label{Fig10.1} \includegraphics[width=0.23\textwidth]{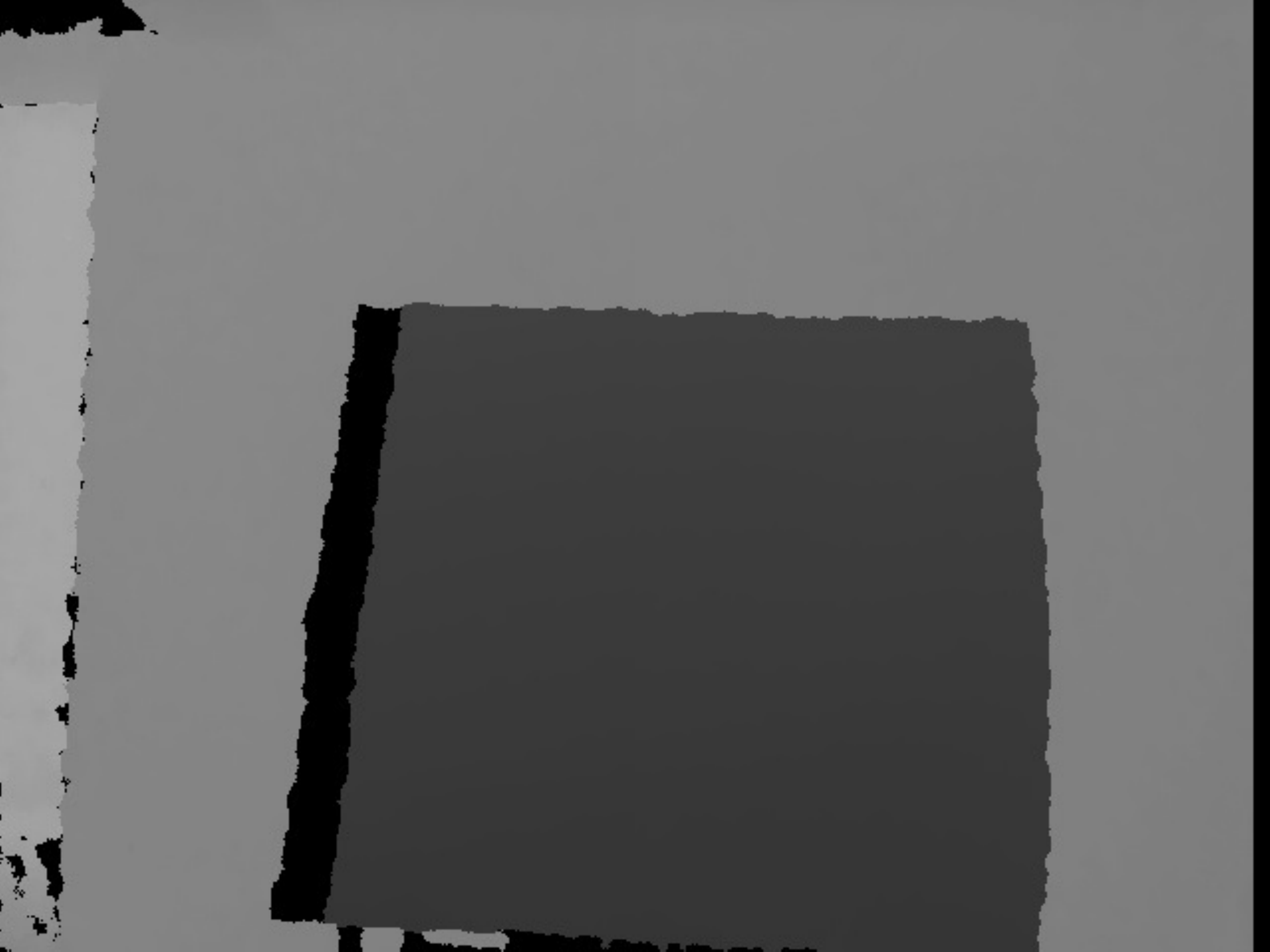}} 
	\subfigure[]{\label{Fig10.2} \includegraphics[width=0.23\textwidth]{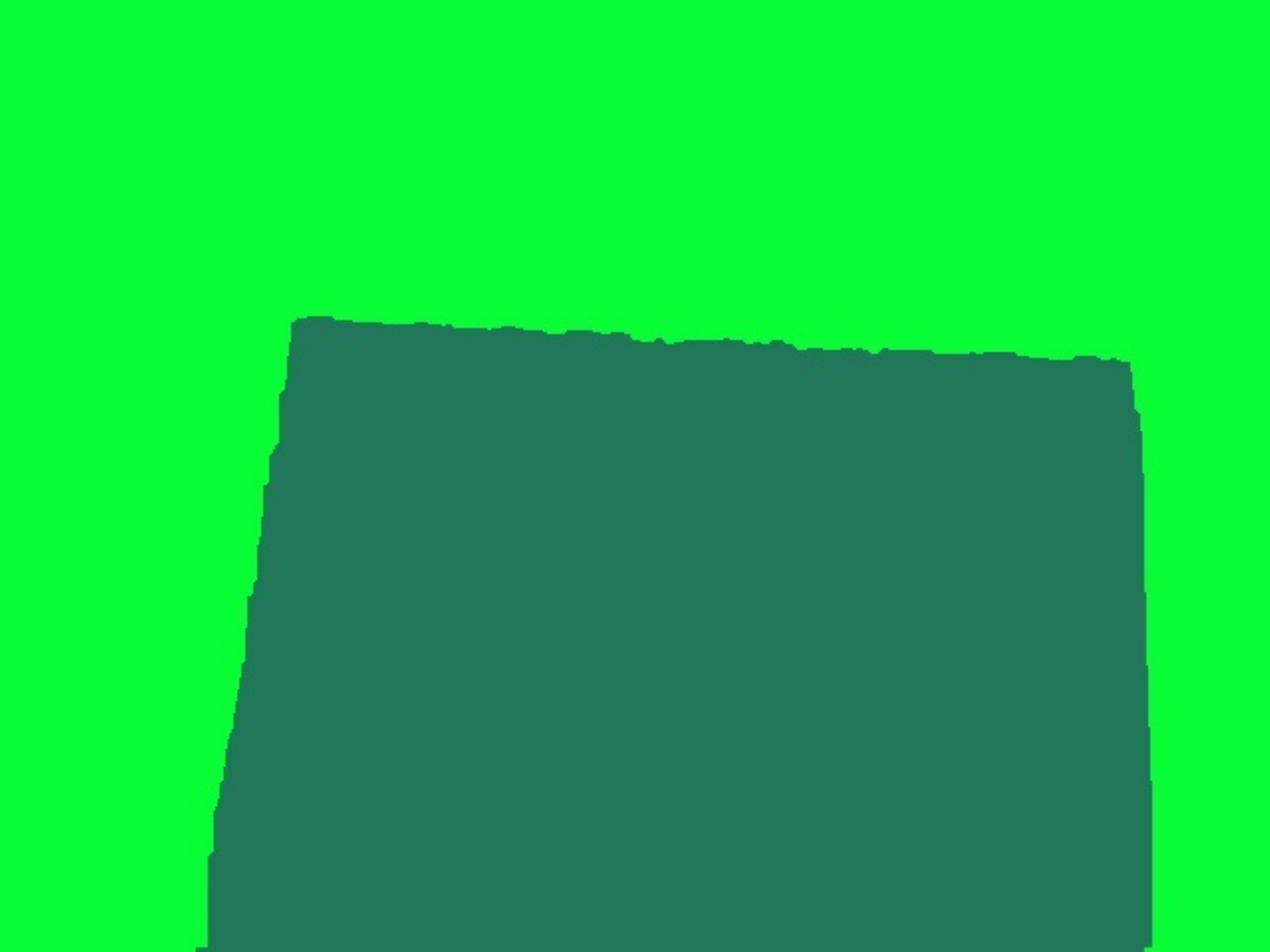}} 
	\subfigure[]{\label{Fig10.3} \includegraphics[width=0.23\textwidth]{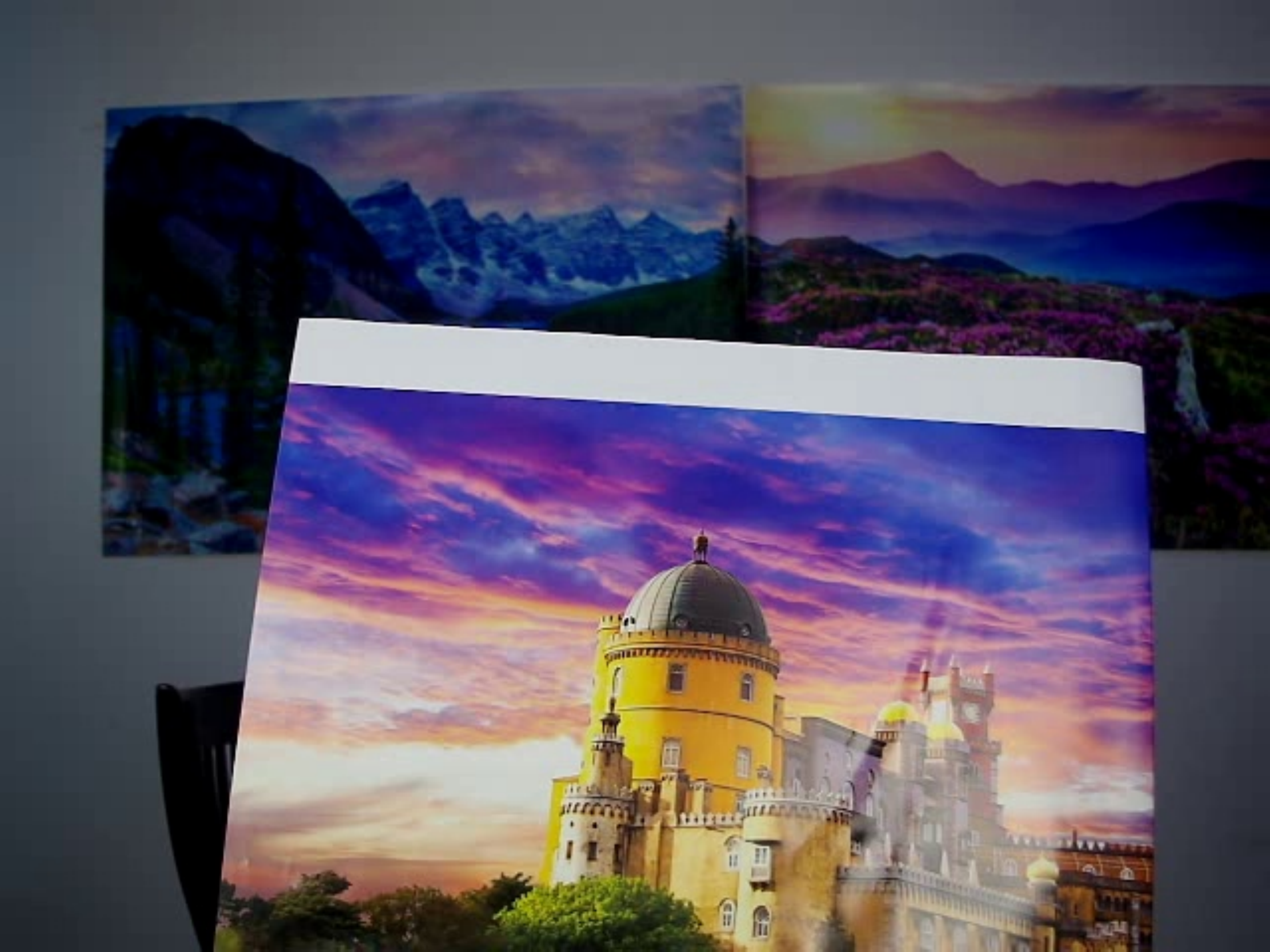}} 
	\subfigure[]{\label{Fig10.4} \includegraphics[width=0.23\textwidth]{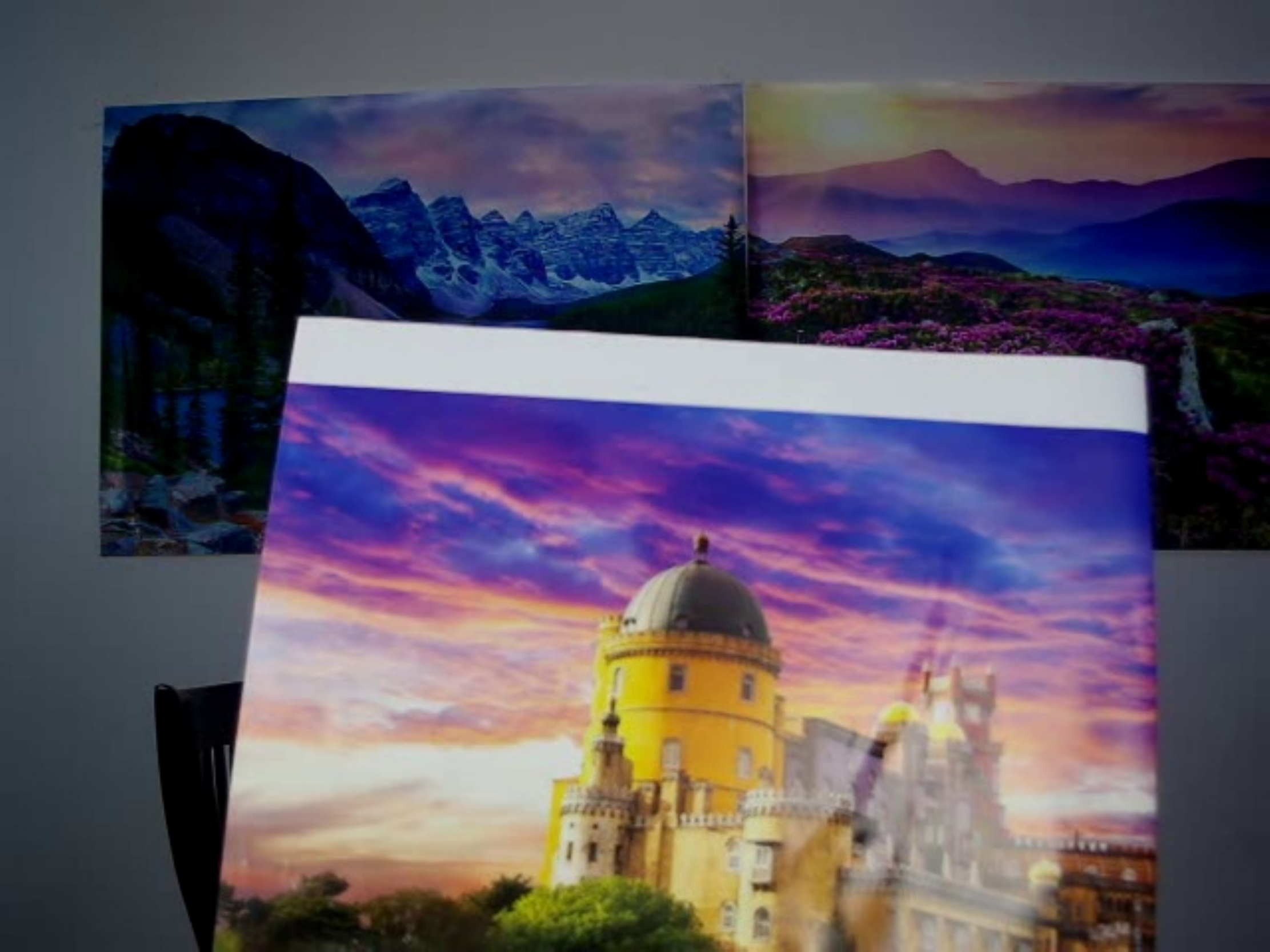}} 
	\subfigure[]{\label{Fig10.5} \includegraphics[width=0.23\textwidth]{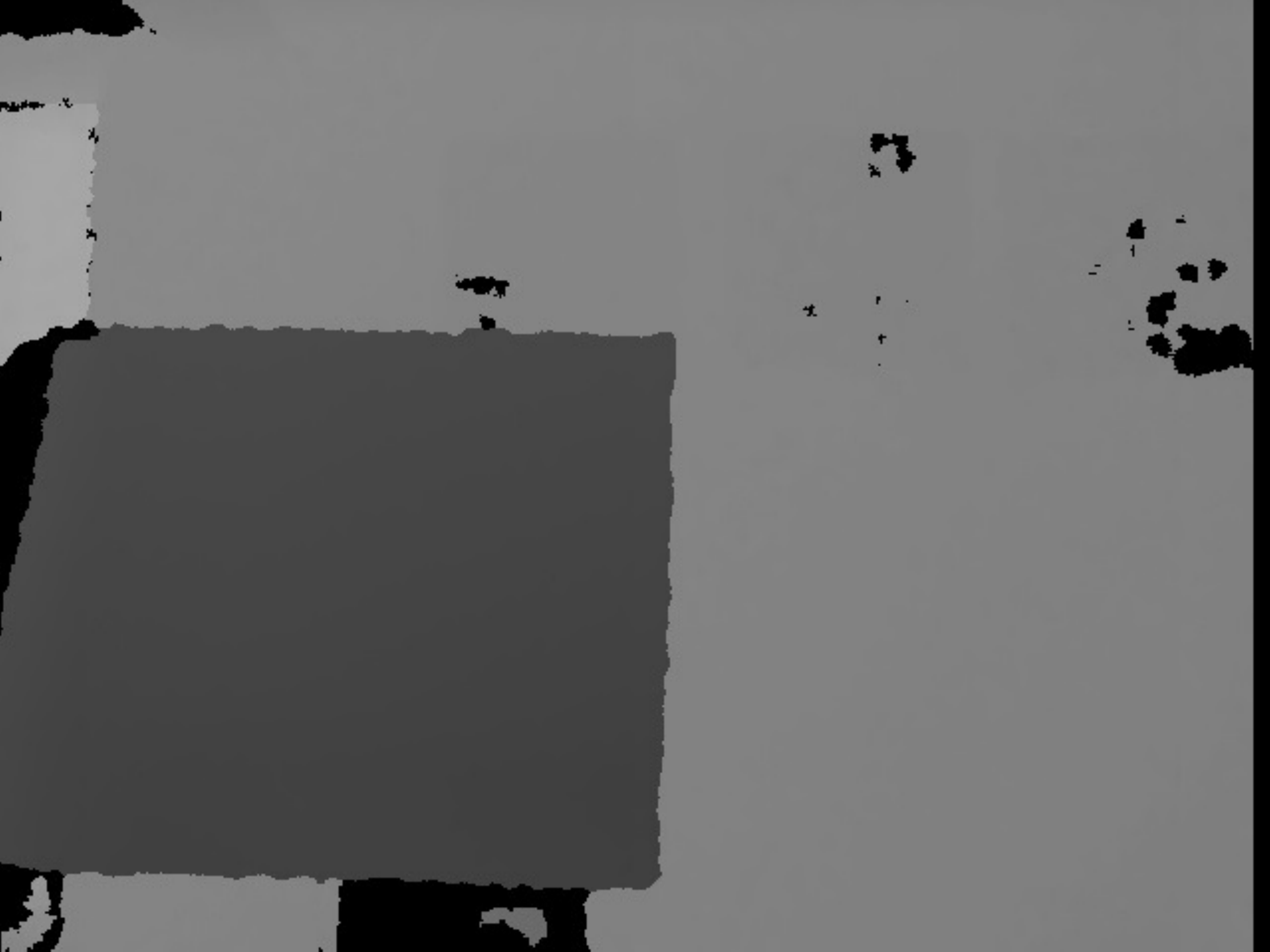}} 
	\subfigure[]{\label{Fig10.6} \includegraphics[width=0.23\textwidth]{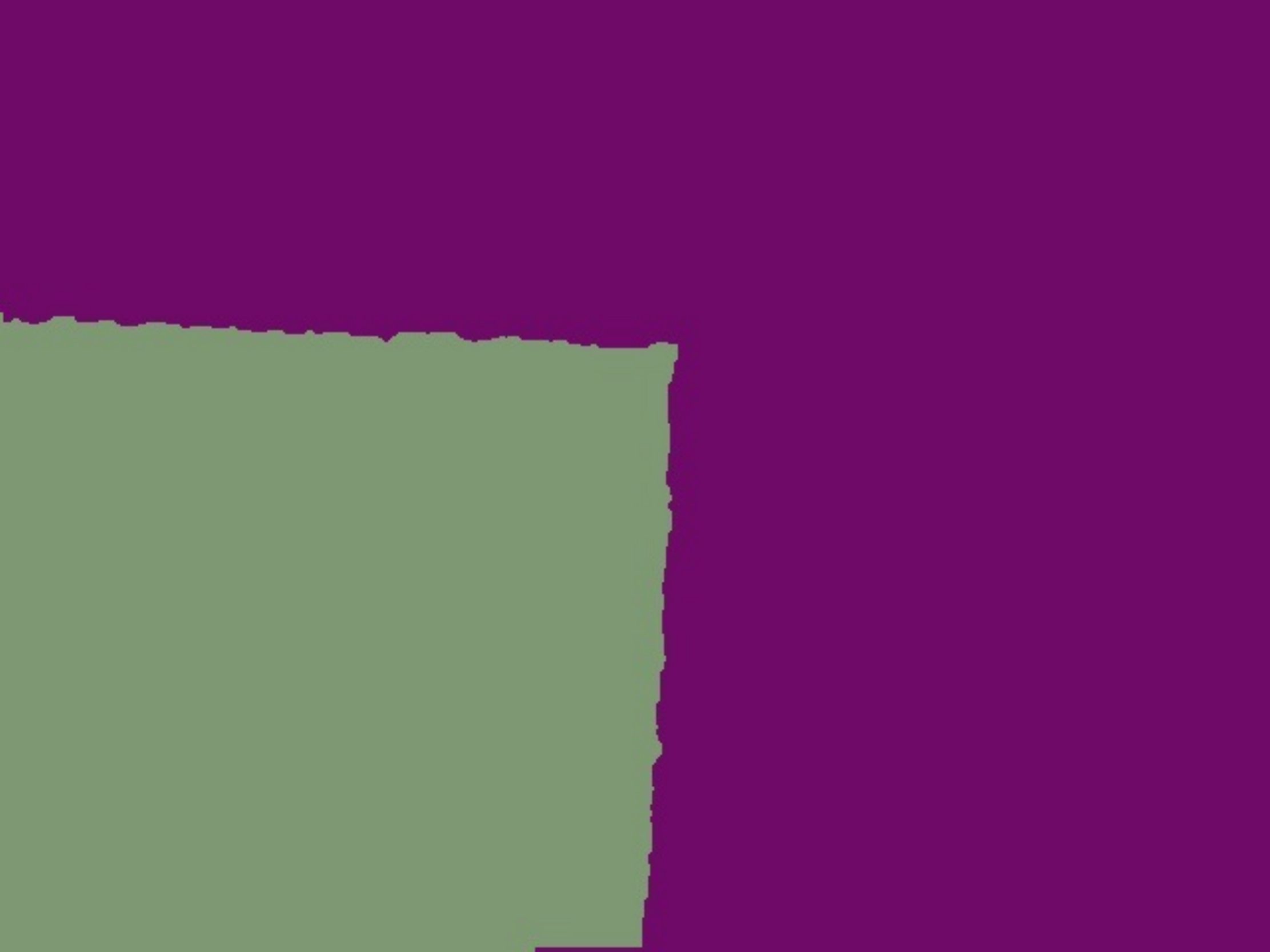}} 
	\subfigure[]{\label{Fig10.7} \includegraphics[width=0.23\textwidth]{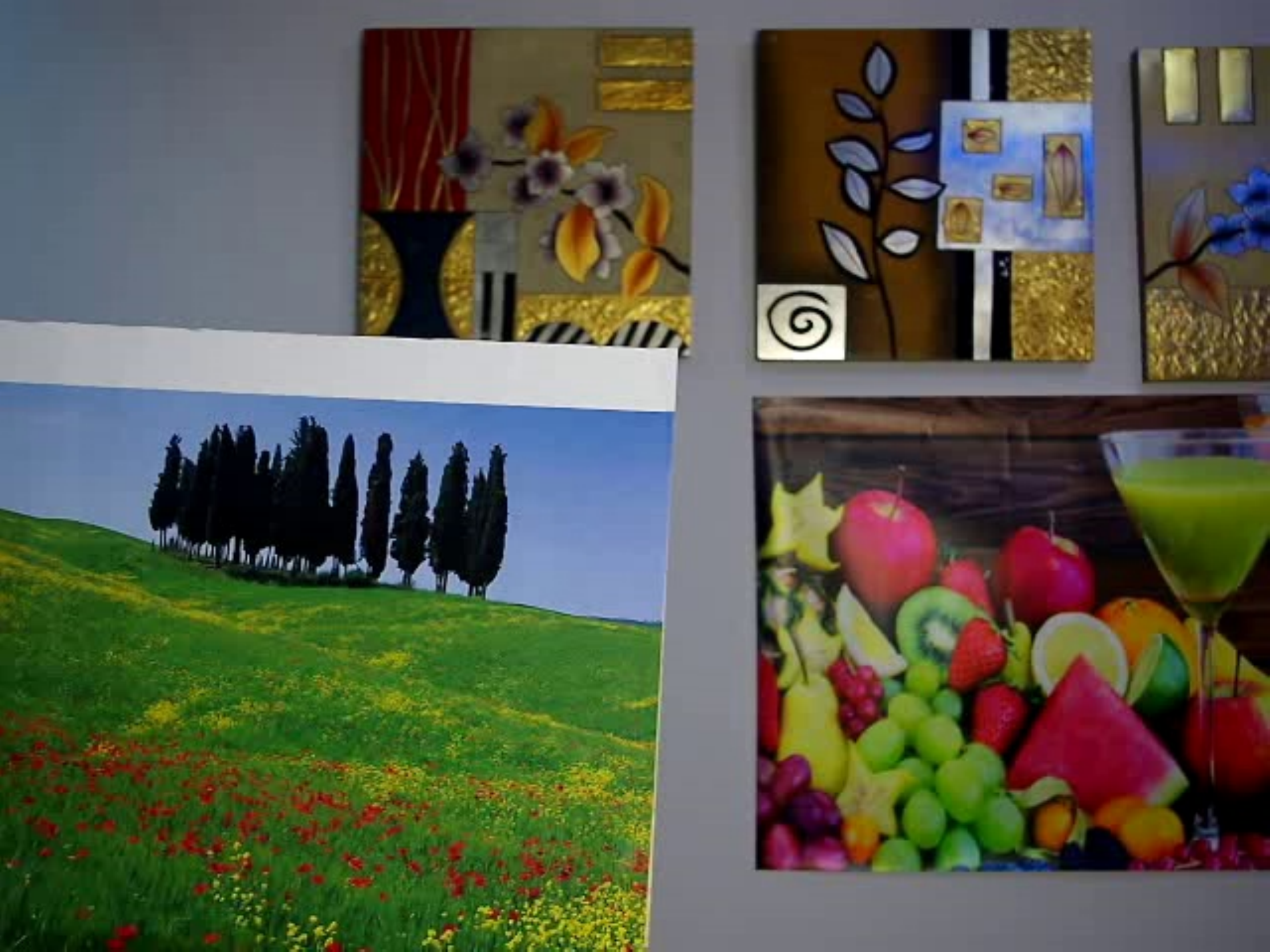}} 
	\subfigure[]{\label{Fig10.8} \includegraphics[width=0.23\textwidth]{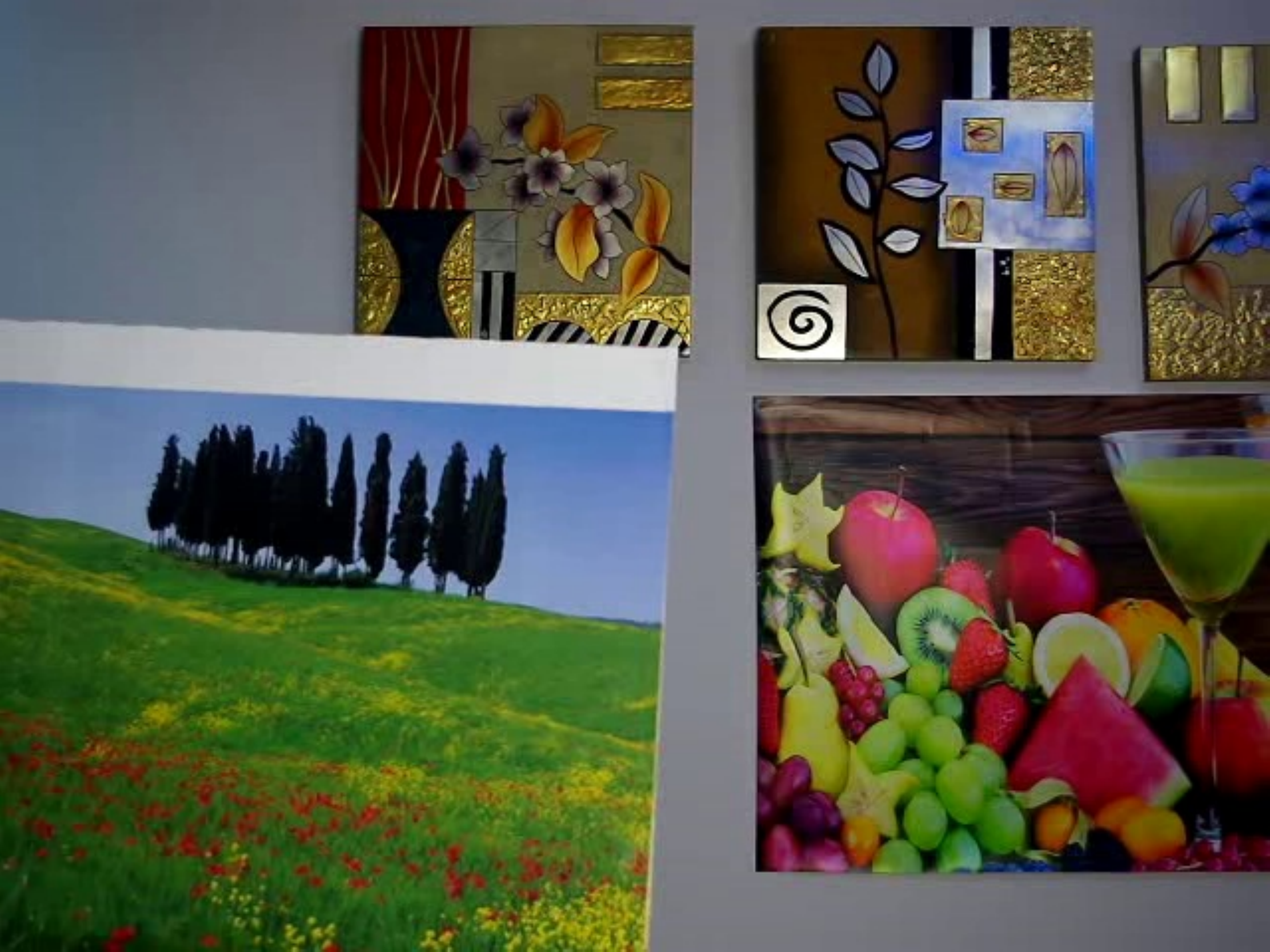}}
	\subfigure[]{\label{Fig10.9} \includegraphics[width=0.23\textwidth]{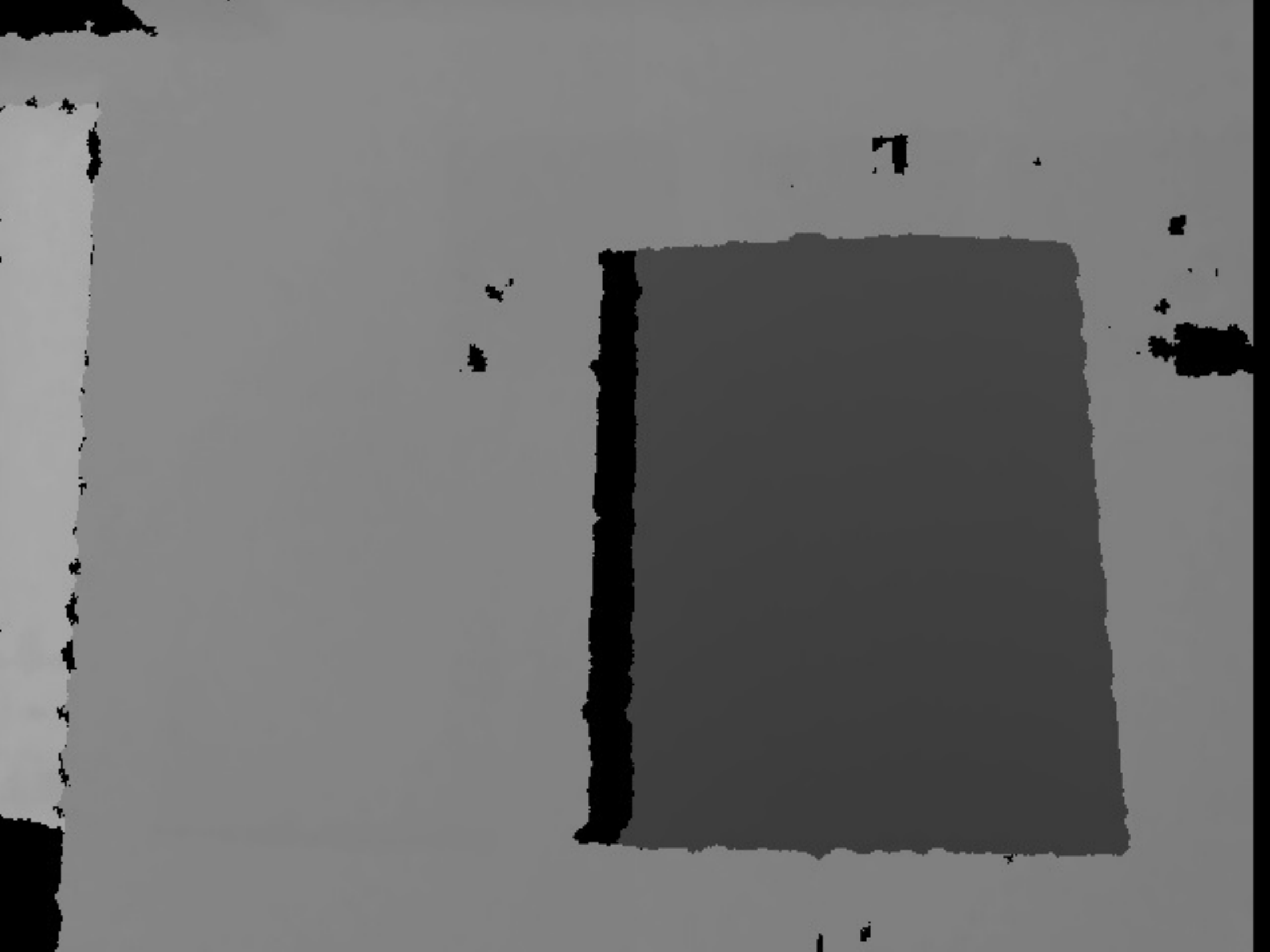}} 
	\subfigure[]{\label{Fig10.10} \includegraphics[width=0.23\textwidth]{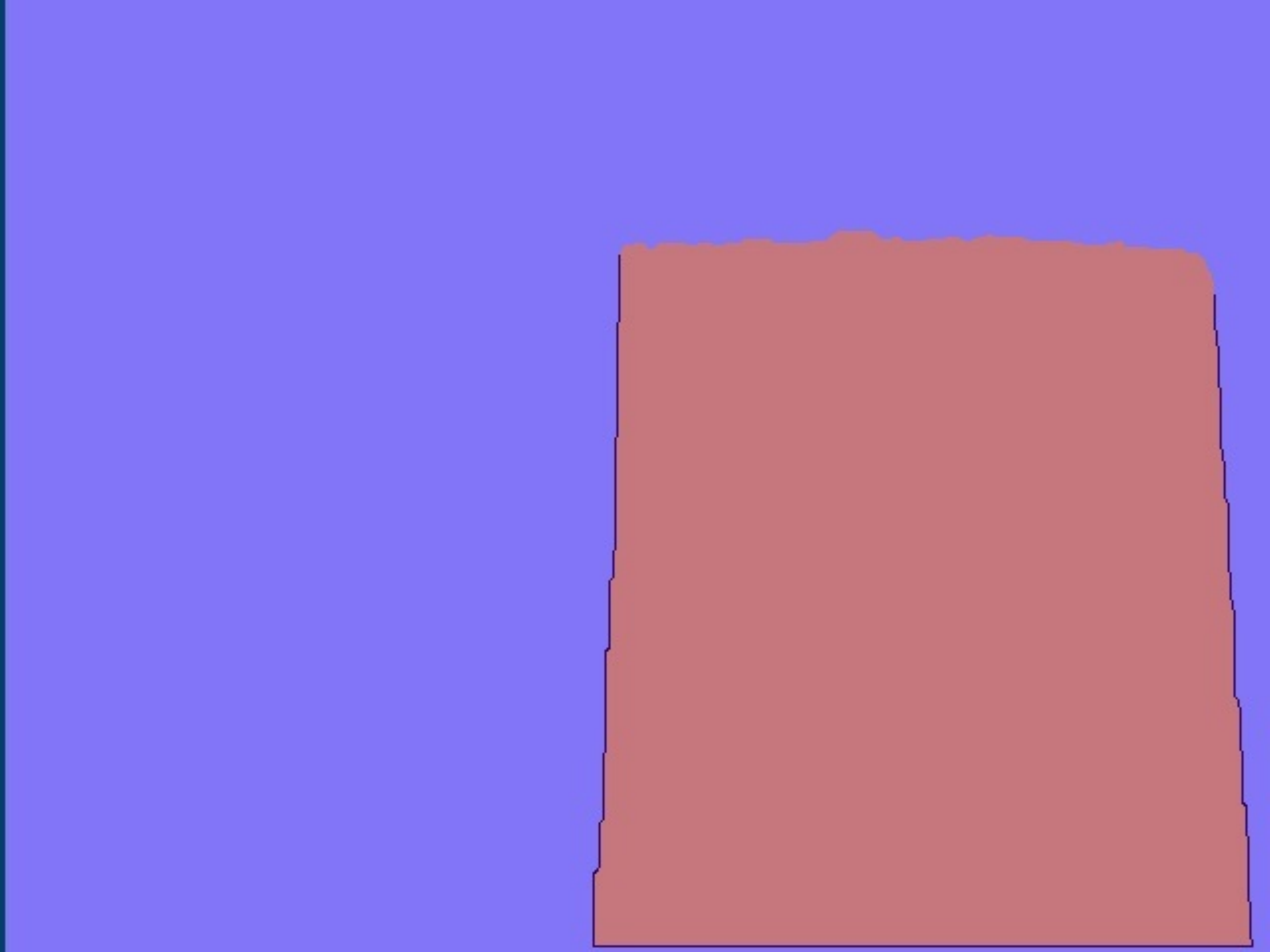}} 
	\subfigure[]{\label{Fig10.11} \includegraphics[width=0.23\textwidth]{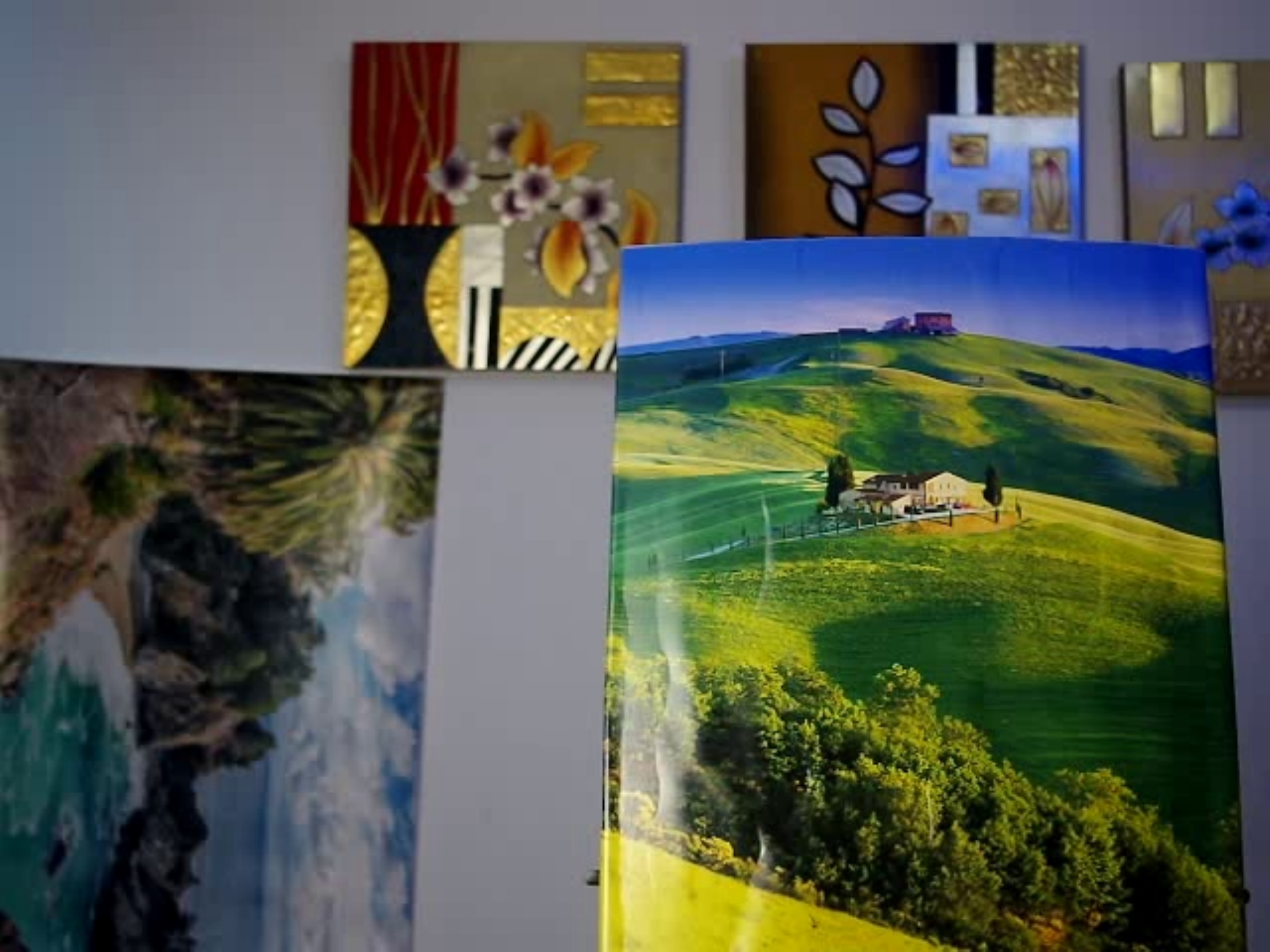}} 
	\subfigure[]{\label{Fig10.12} \includegraphics[width=0.23\textwidth]{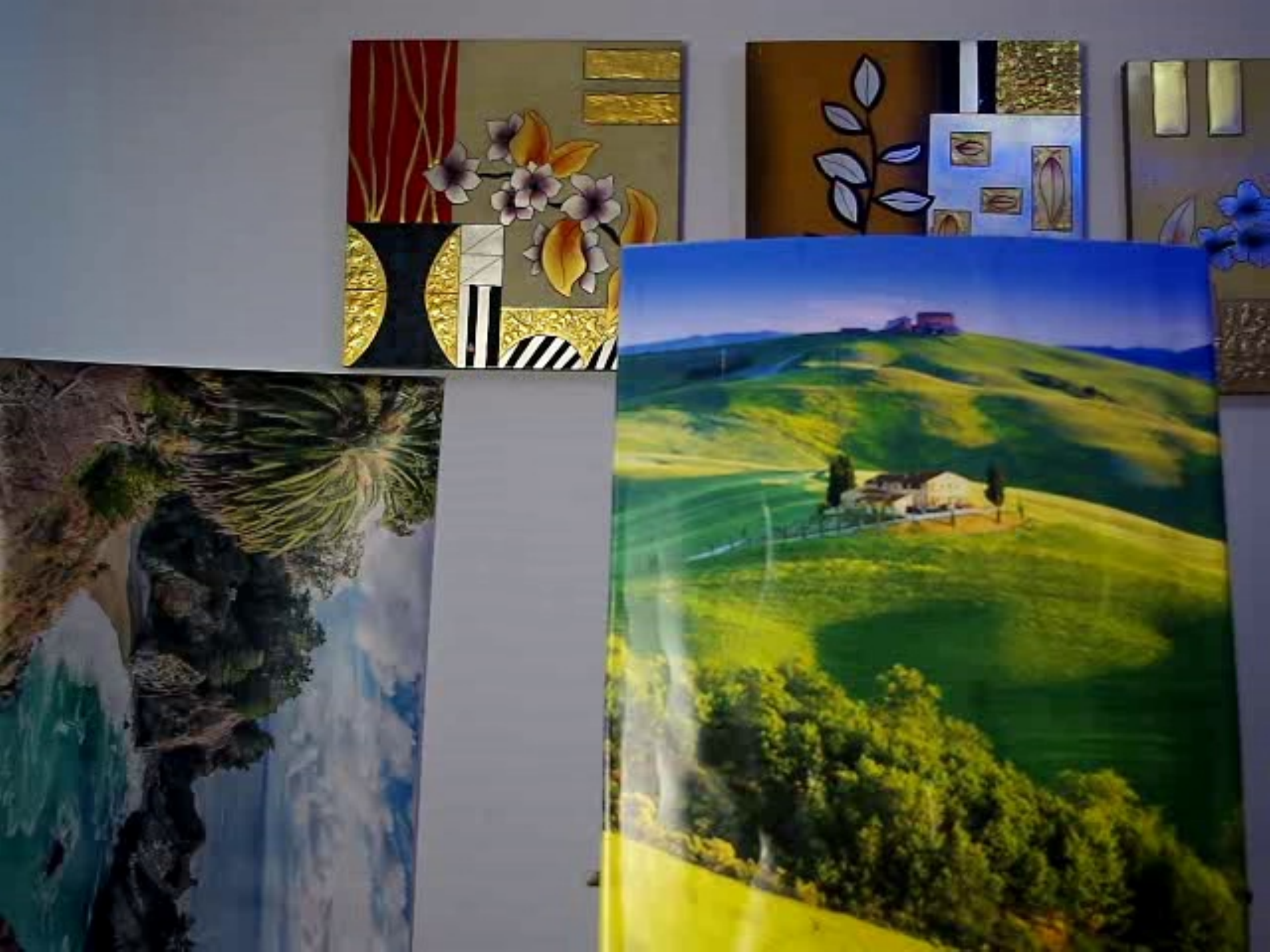}} 
	\subfigure[]{\label{Fig10.13} \includegraphics[width=0.23\textwidth]{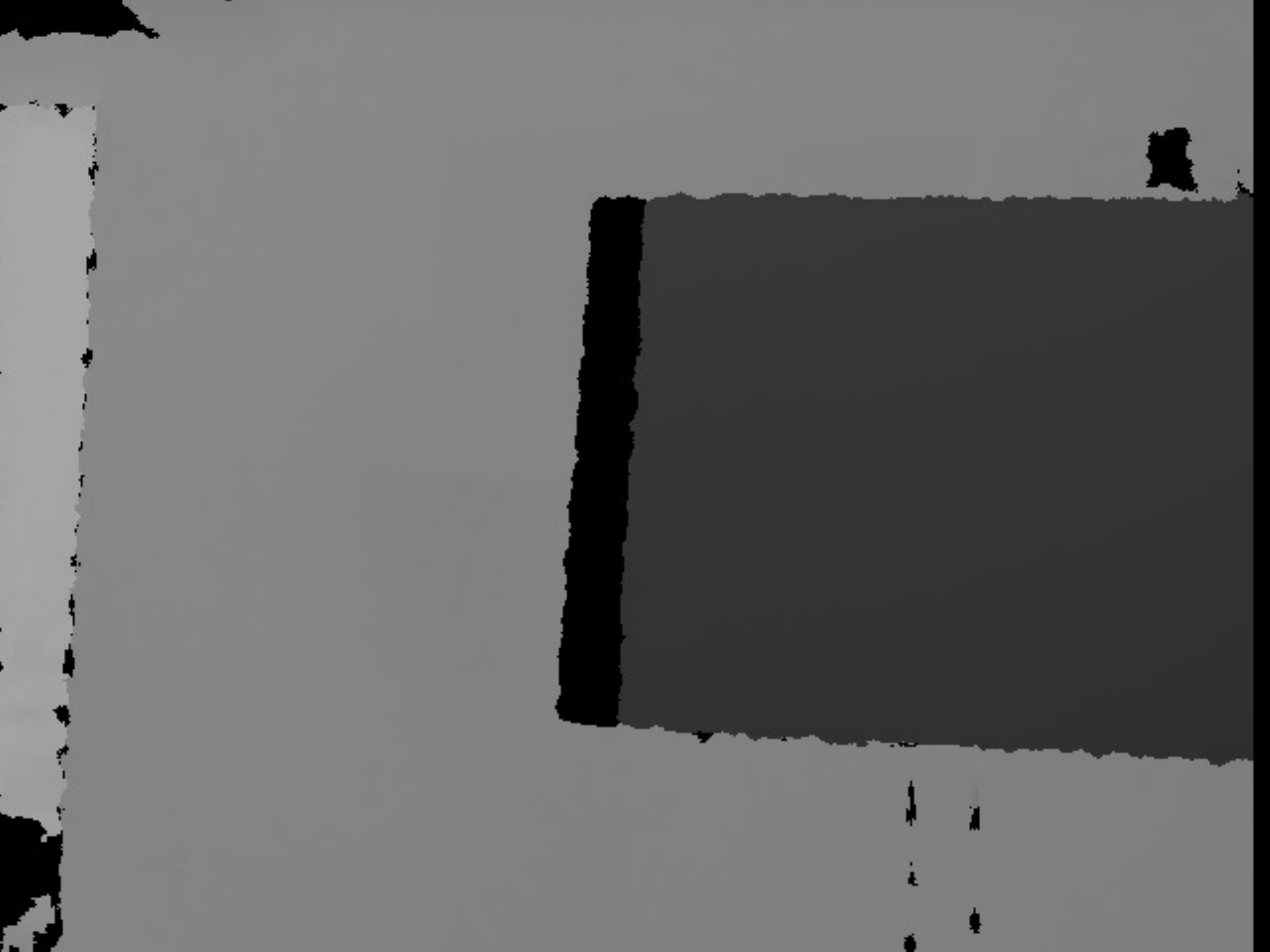}} 
	\subfigure[]{\label{Fig10.14} \includegraphics[width=0.23\textwidth]{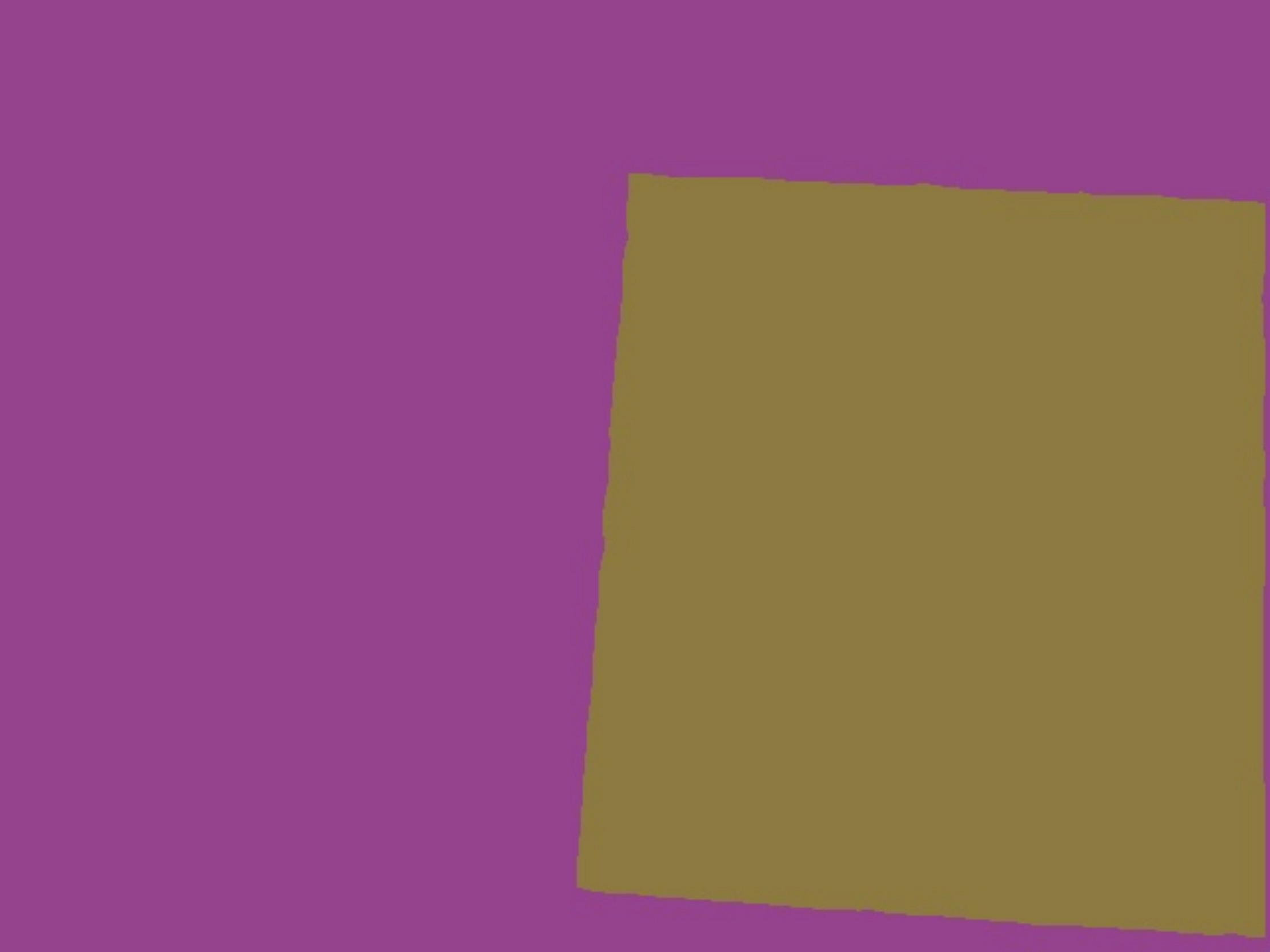}} 
	\subfigure[]{\label{Fig10.15} \includegraphics[width=0.23\textwidth]{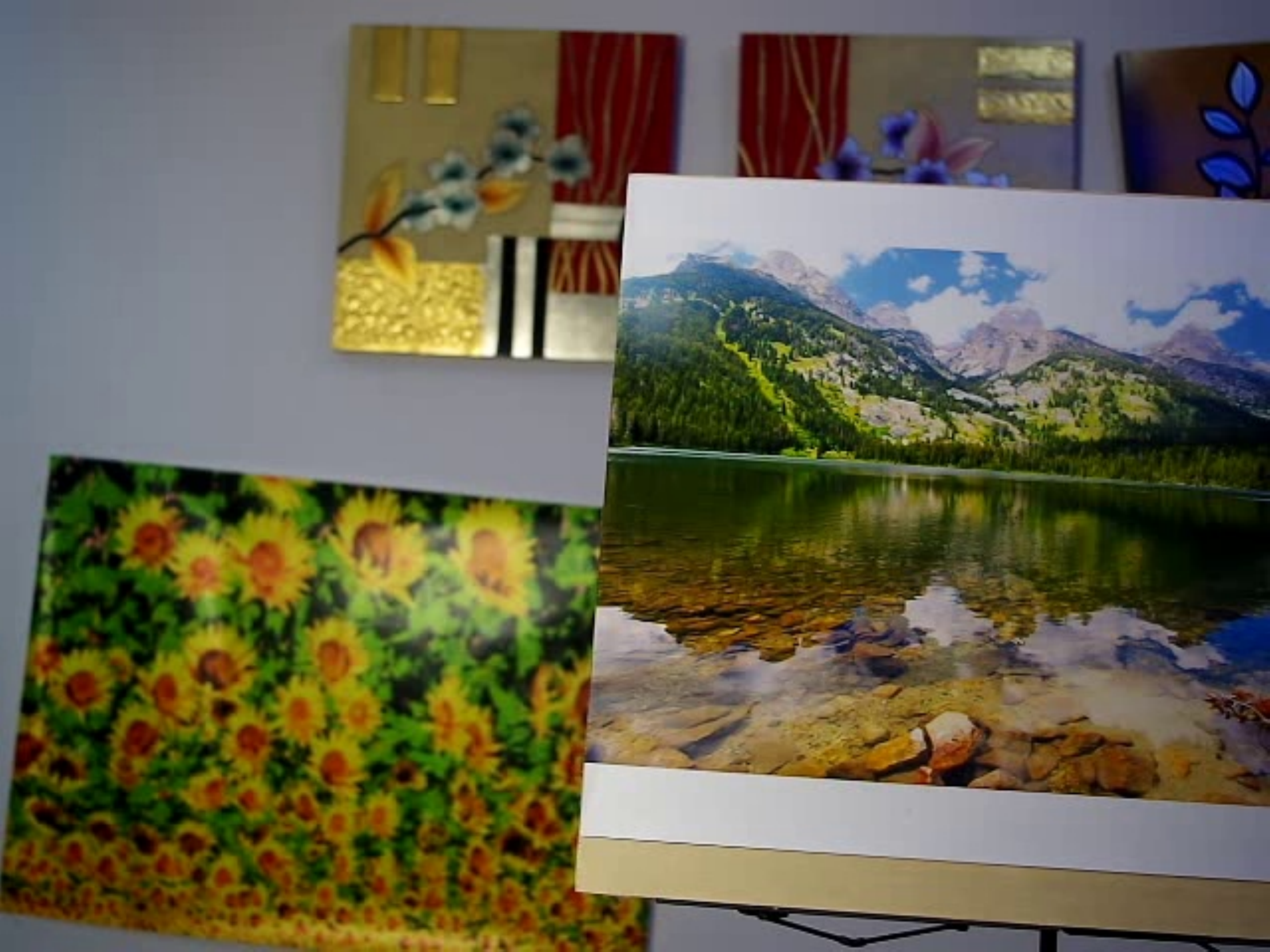}} 
	\subfigure[]{\label{Fig10.16} \includegraphics[width=0.23\textwidth]{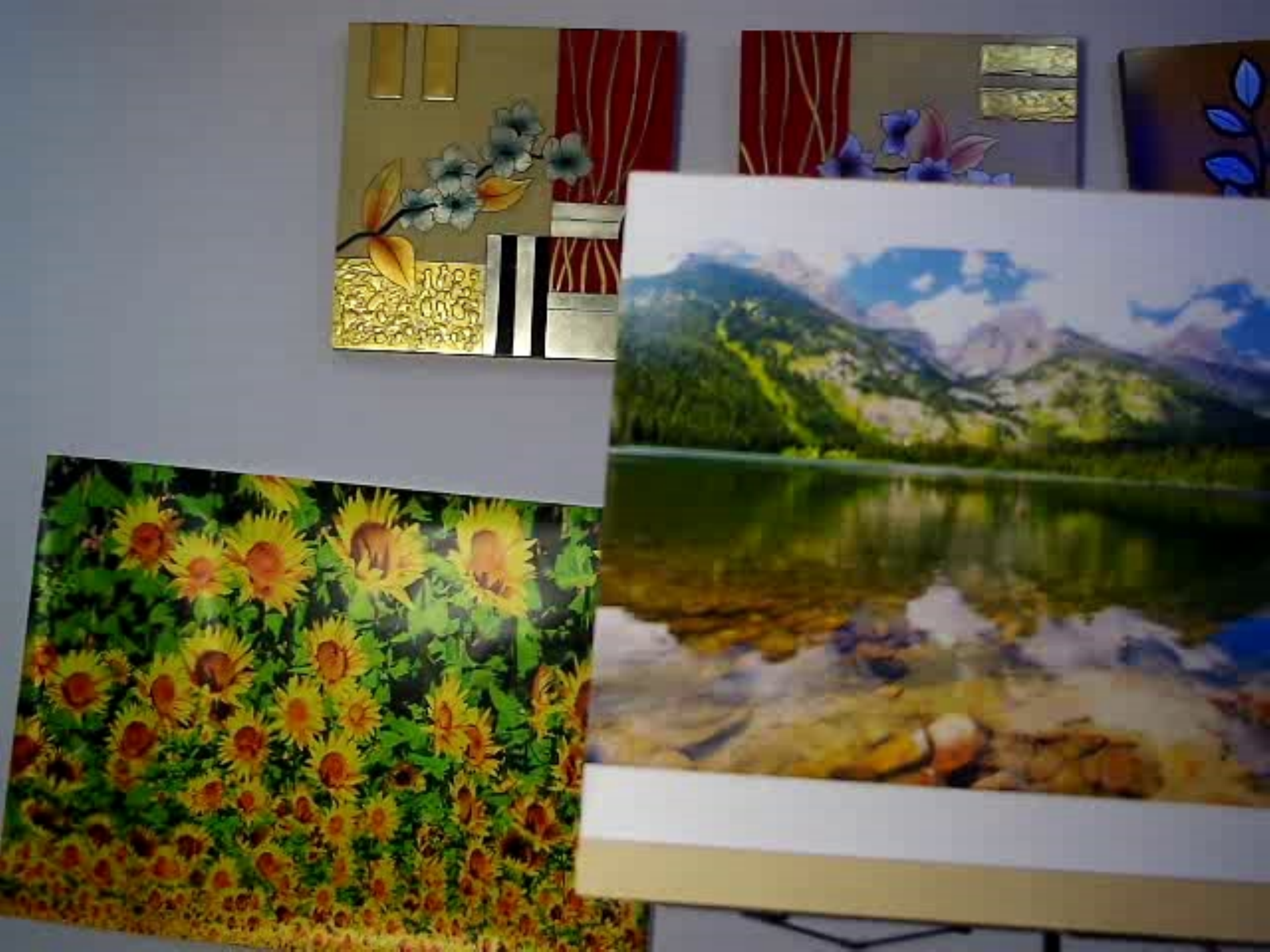}}
	\subfigure[]{\label{Fig10.17} \includegraphics[width=0.23\textwidth]{figs2_q-eps-converted-to.pdf}}
	\subfigure[]{\label{Fig10.18} \includegraphics[width=0.23\textwidth]{figs2_r-eps-converted-to.pdf}} 
	\subfigure[]{\label{Fig10.19} \includegraphics[width=0.23\textwidth]{figs2_s-eps-converted-to.pdf}} 
	\subfigure[]{\label{Fig10.20} \includegraphics[width=0.23\textwidth]{figs2_t-eps-converted-to.pdf}} 
	\makeatletter 
	\renewcommand{\thefigure}{S\@arabic\c@figure}
	\makeatother
	\caption{Multi-focus source images captured at five different scenes, the four images in each row belong to a same scene. In each row, the first image is the depth map of the scene, the second image is the segmentation result of the depth map, the front object is in best focus in the third image and the background objects is in best focus in the last image, (a)-(d) belong to the first scene, (e)-(h) belong to the second scene, (i)-(l) belong to the third scene, (m)-(p) belong to the fourth scene, (q)-(t) belong to the fifth scene.}
	\label{FigS2}
\end{figure*}

\begin{figure*}[h]
	\centering		
	\subfigure[]{\label{Fig11.1} \includegraphics[width=0.1884\textwidth]{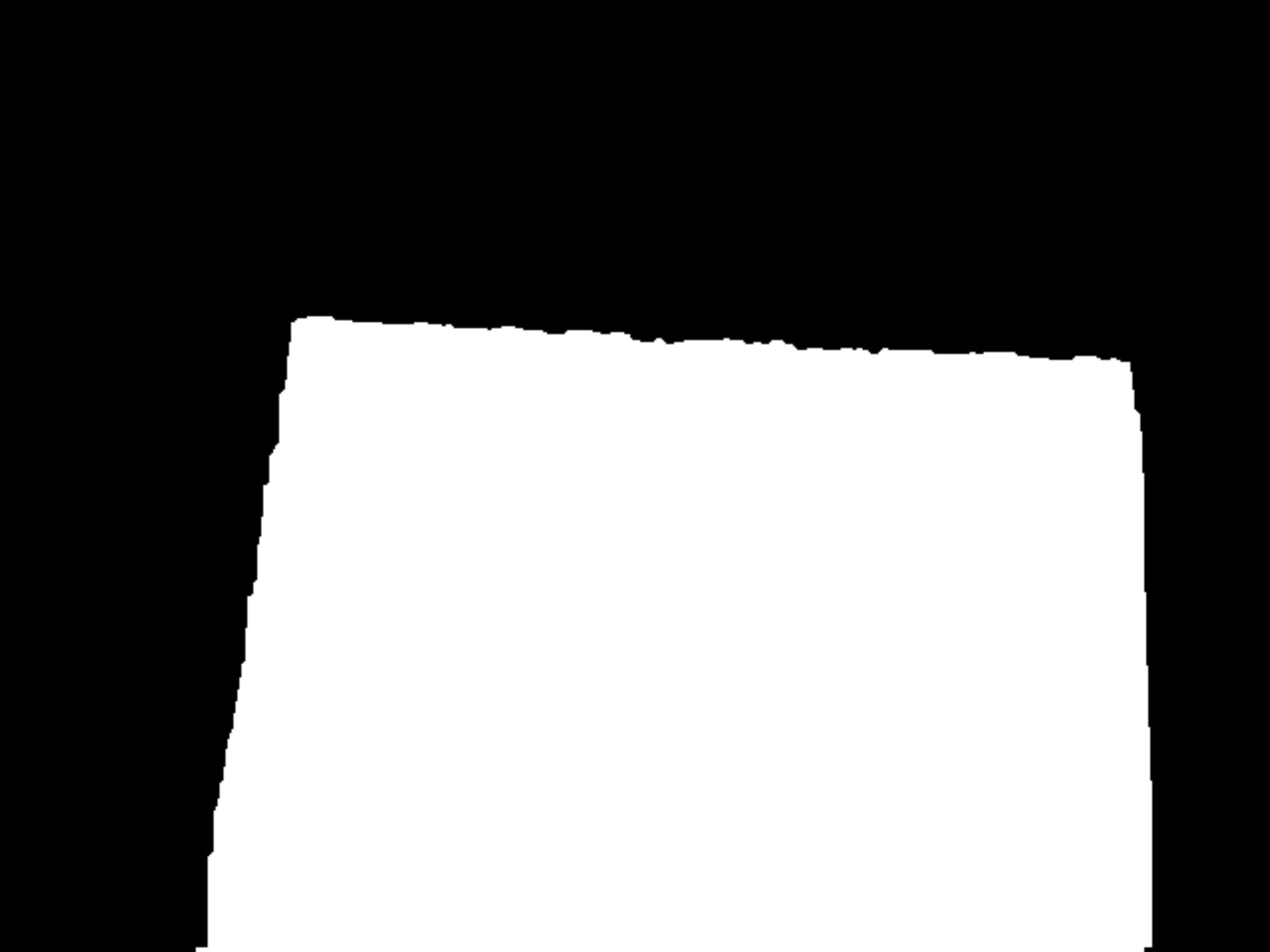}} 
	\subfigure[]{\label{Fig11.1} \includegraphics[width=0.1884\textwidth]{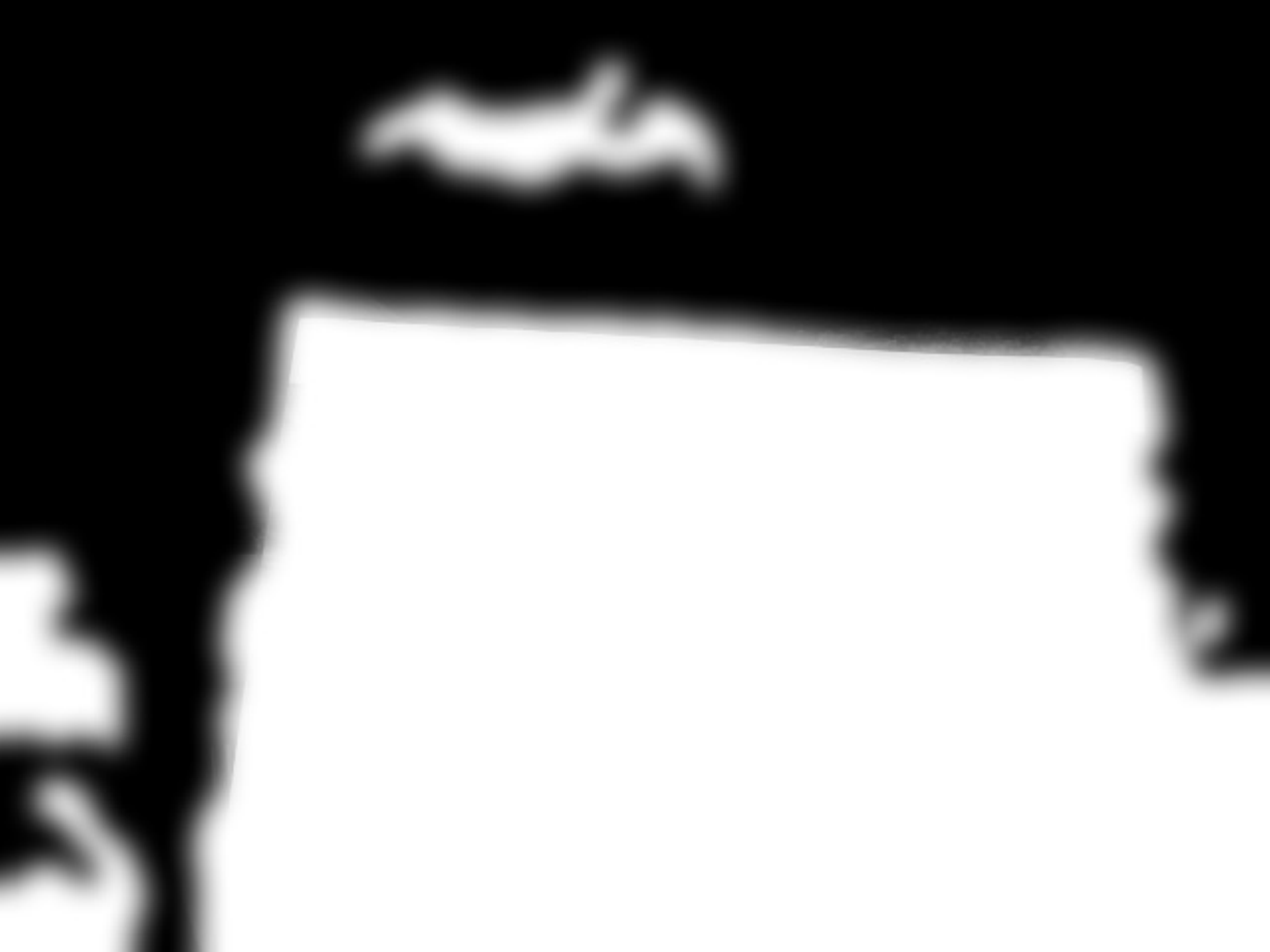}} 
	\subfigure[]{\label{Fig11.2} \includegraphics[width=0.1884\textwidth]{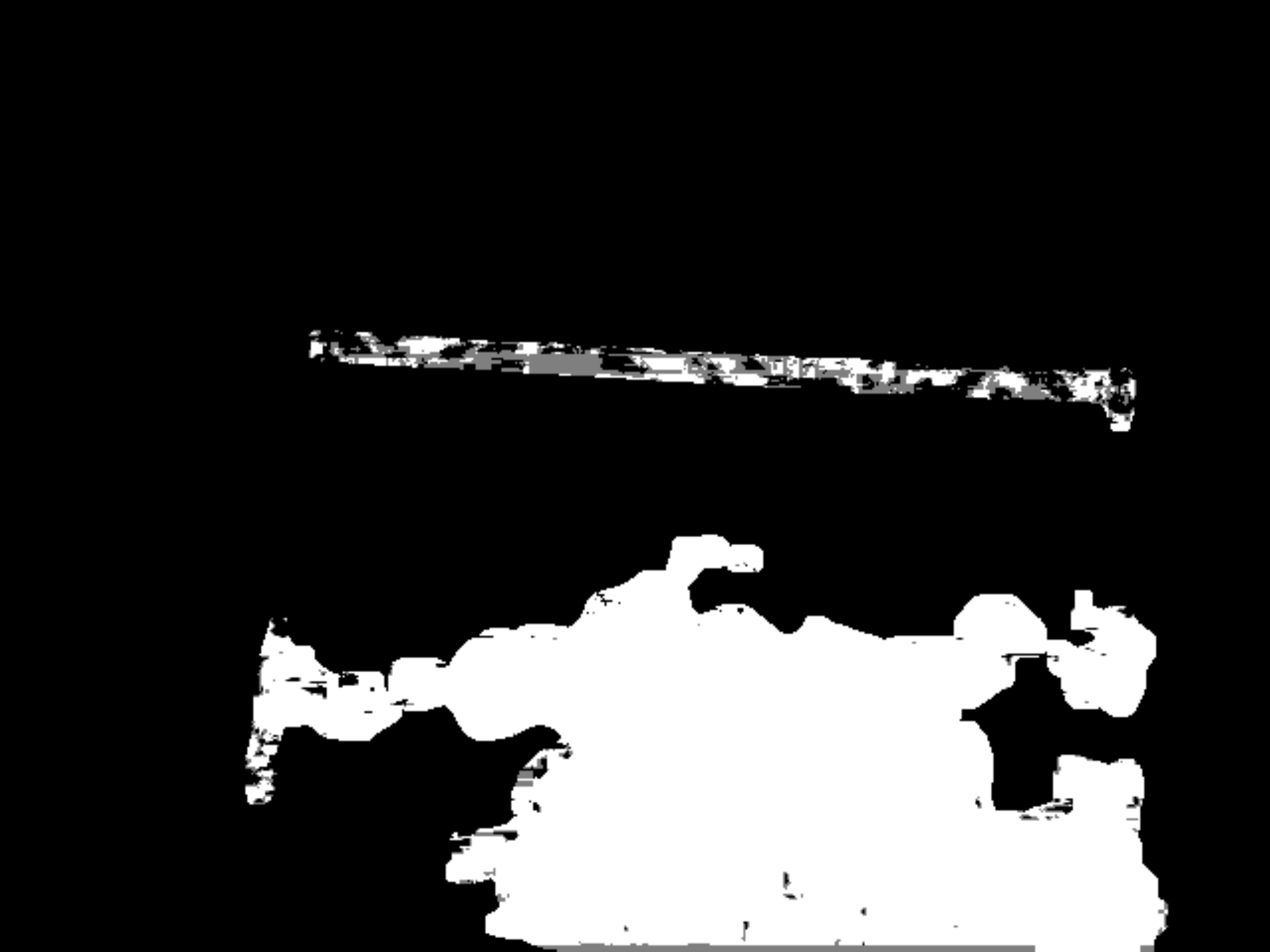}} 
	\subfigure[]{\label{Fig11.3} \includegraphics[width=0.1884\textwidth]{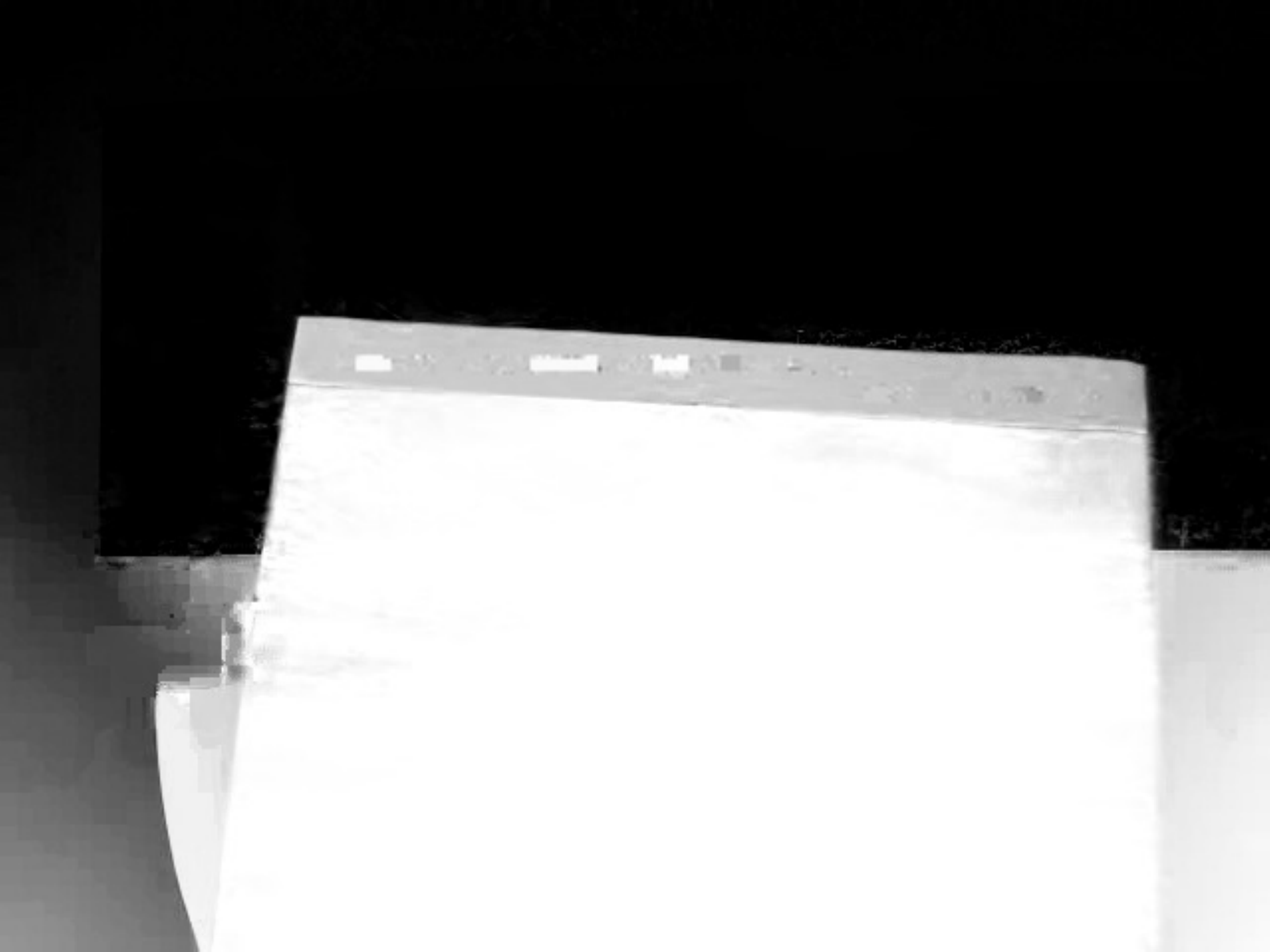}} 
	\subfigure[]{\label{Fig11.4} \includegraphics[width=0.1884\textwidth]{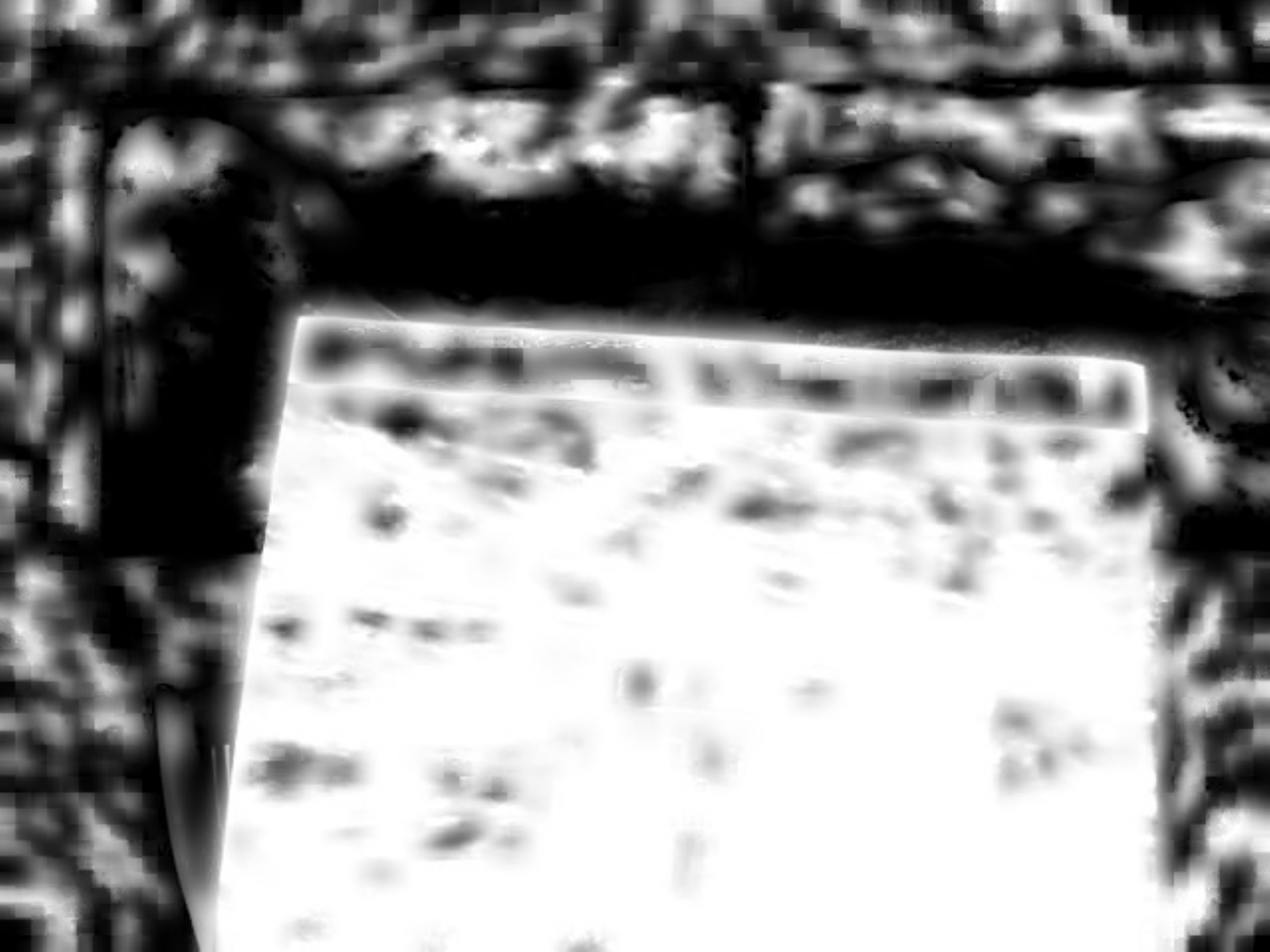}} 
	\subfigure[]{\label{Fig11.5} \includegraphics[width=0.1884\textwidth]{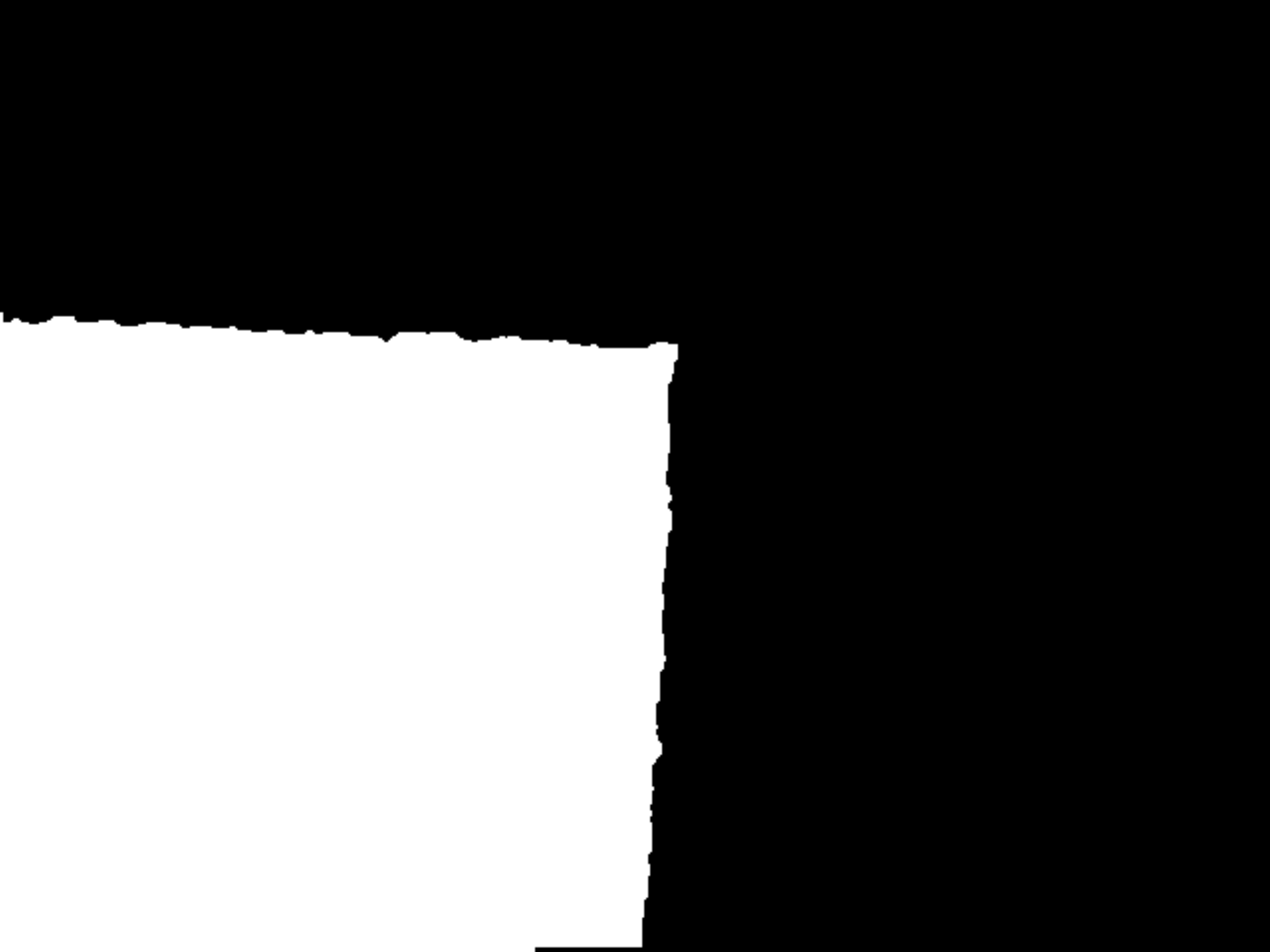}}  
	\subfigure[]{\label{Fig11.1} \includegraphics[width=0.1884\textwidth]{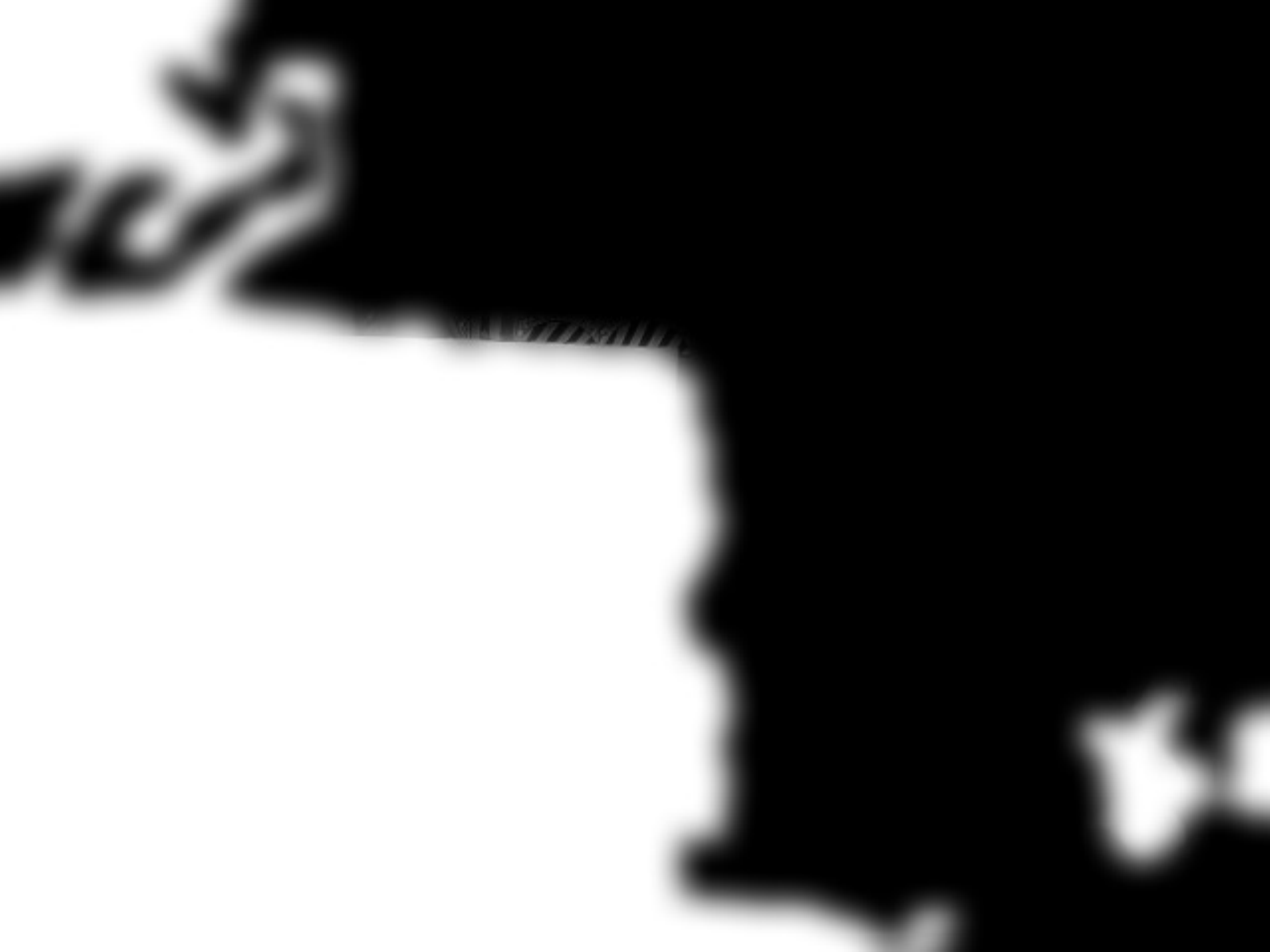}} 
	\subfigure[]{\label{Fig11.6} \includegraphics[width=0.1884\textwidth]{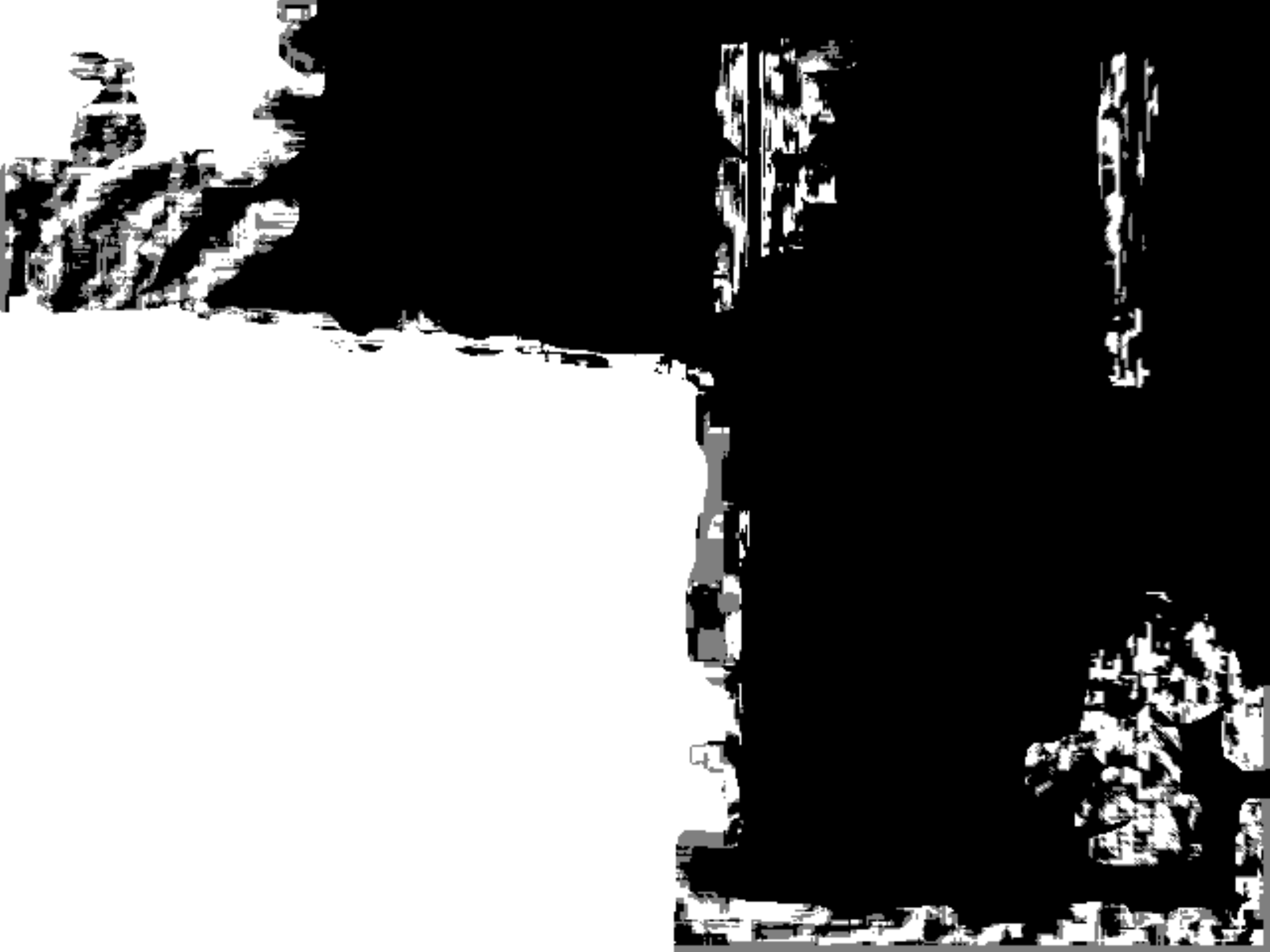}} 
	\subfigure[]{\label{Fig11.7} \includegraphics[width=0.1884\textwidth]{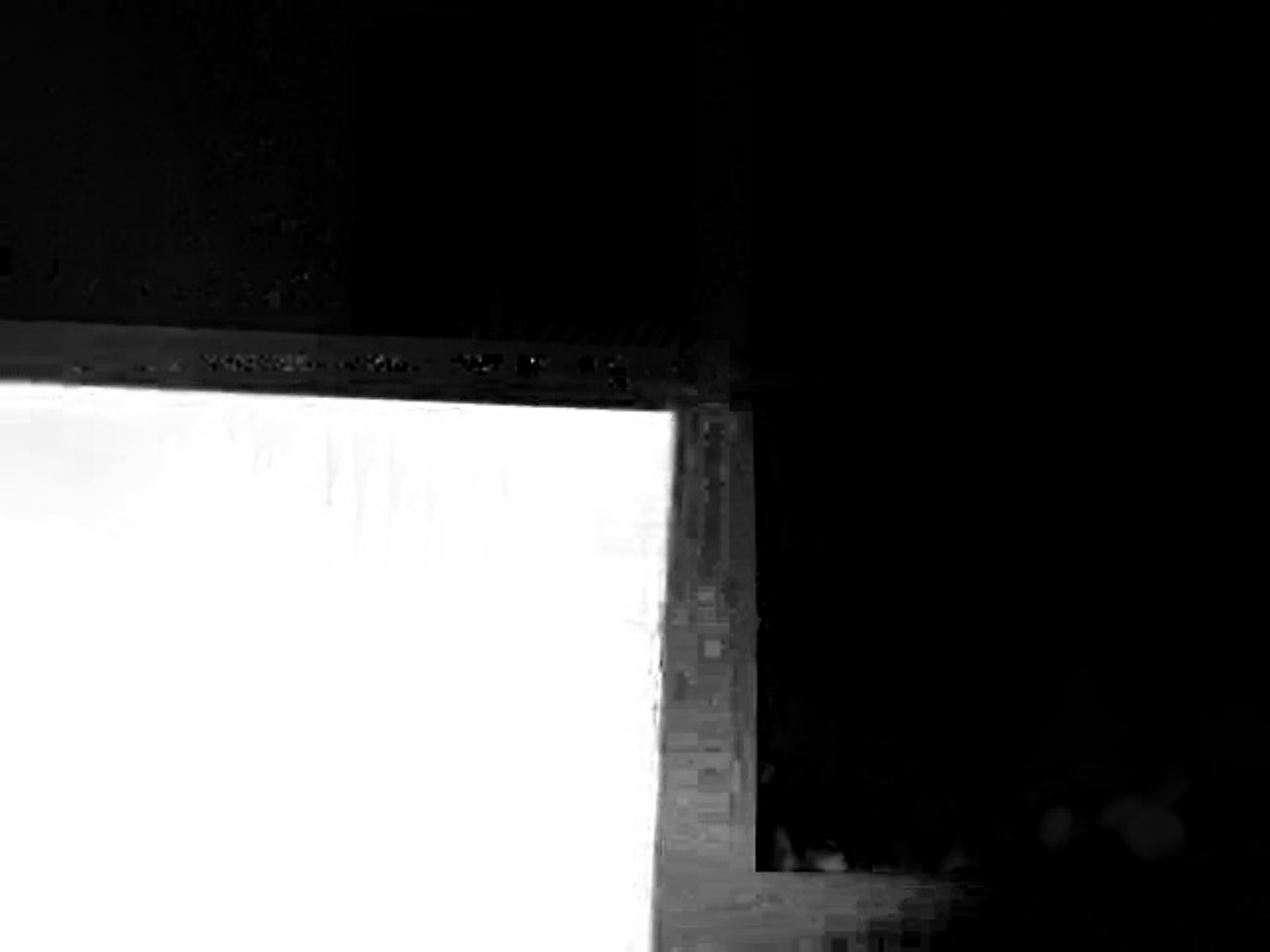}} 
	\subfigure[]{\label{Fig11.8} \includegraphics[width=0.1884\textwidth]{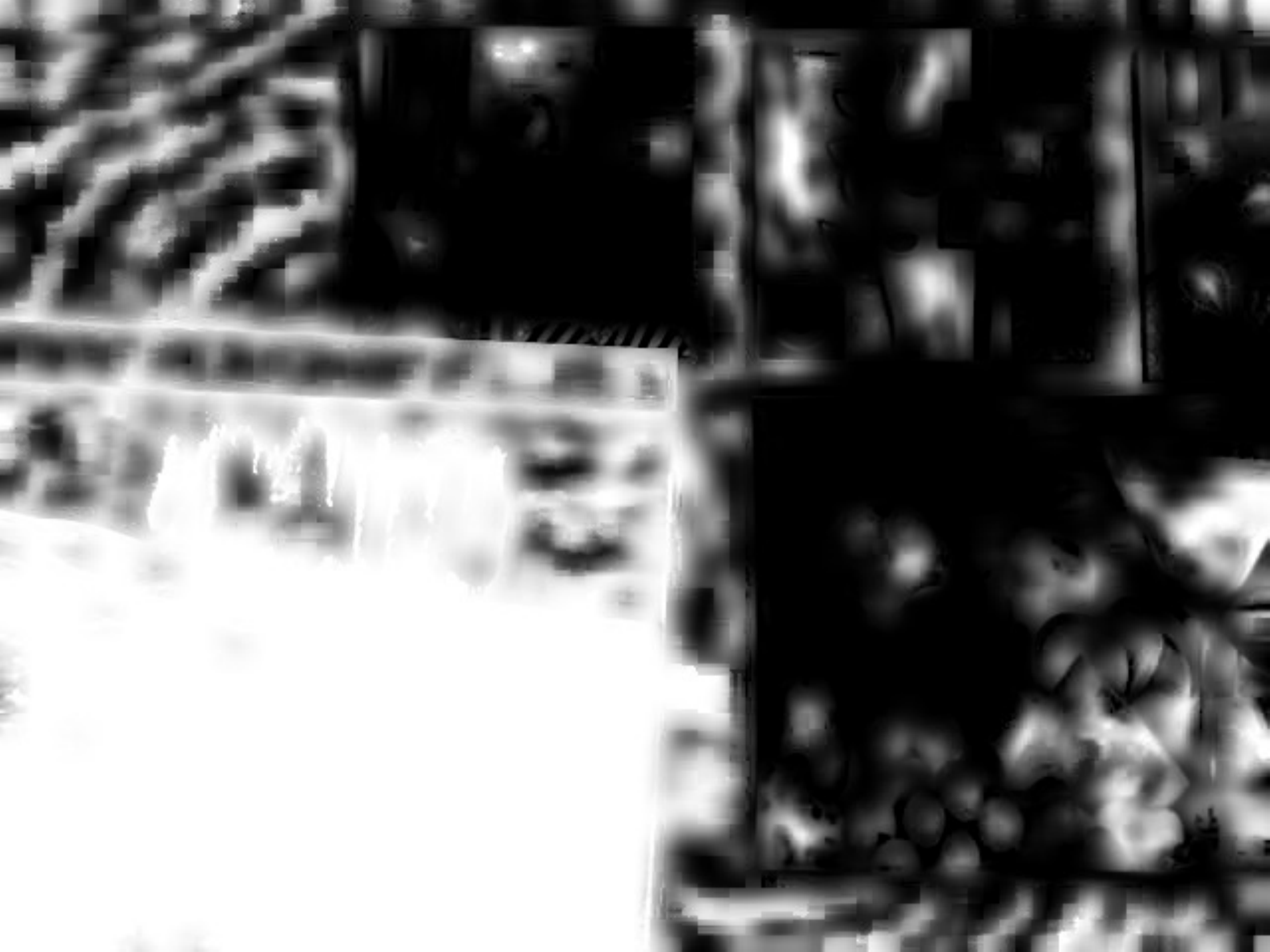}} 
	\subfigure[]{\label{Fig11.9} \includegraphics[width=0.1884\textwidth]{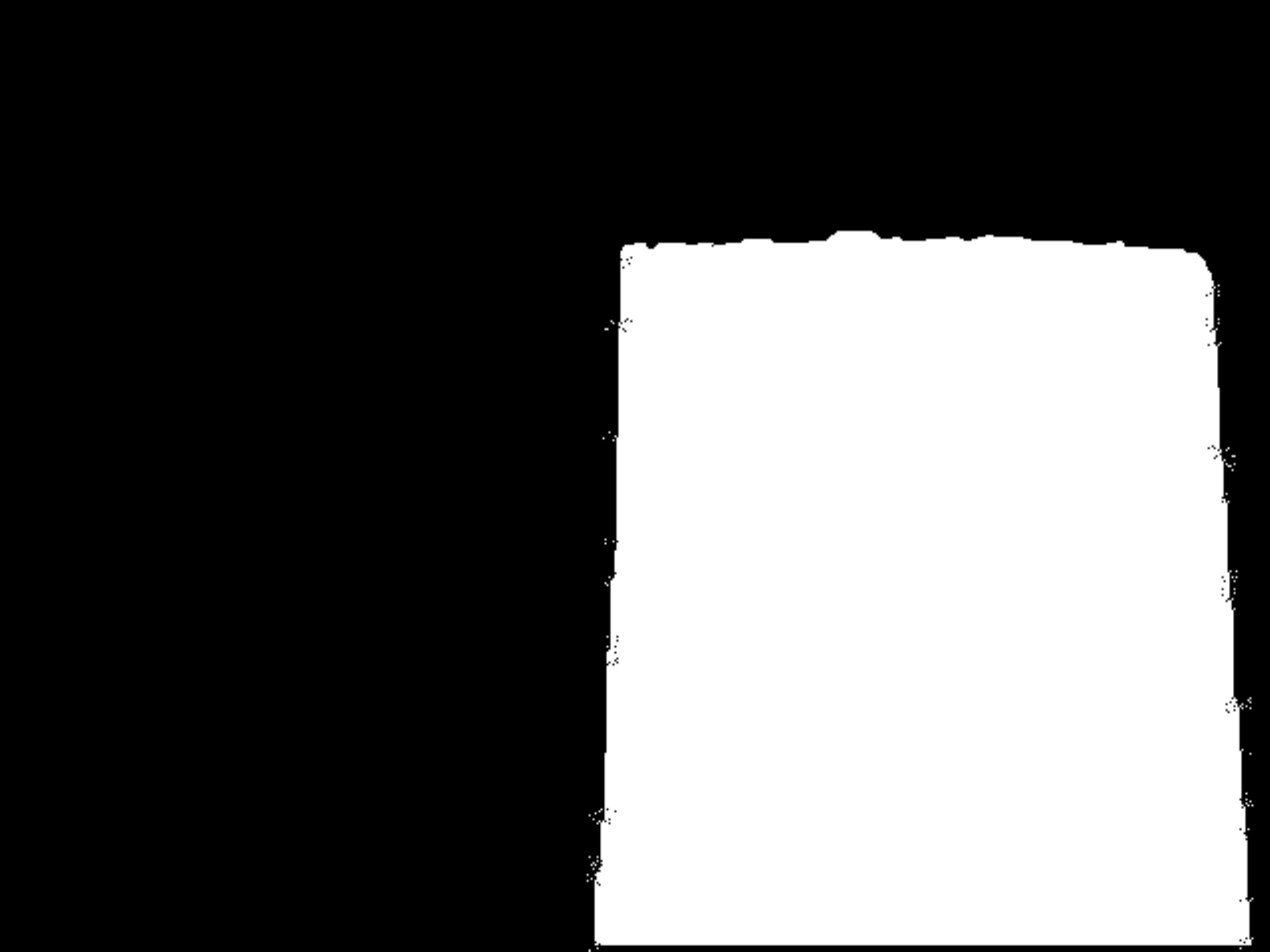}} 
	\subfigure[]{\label{Fig11.1} \includegraphics[width=0.1884\textwidth]{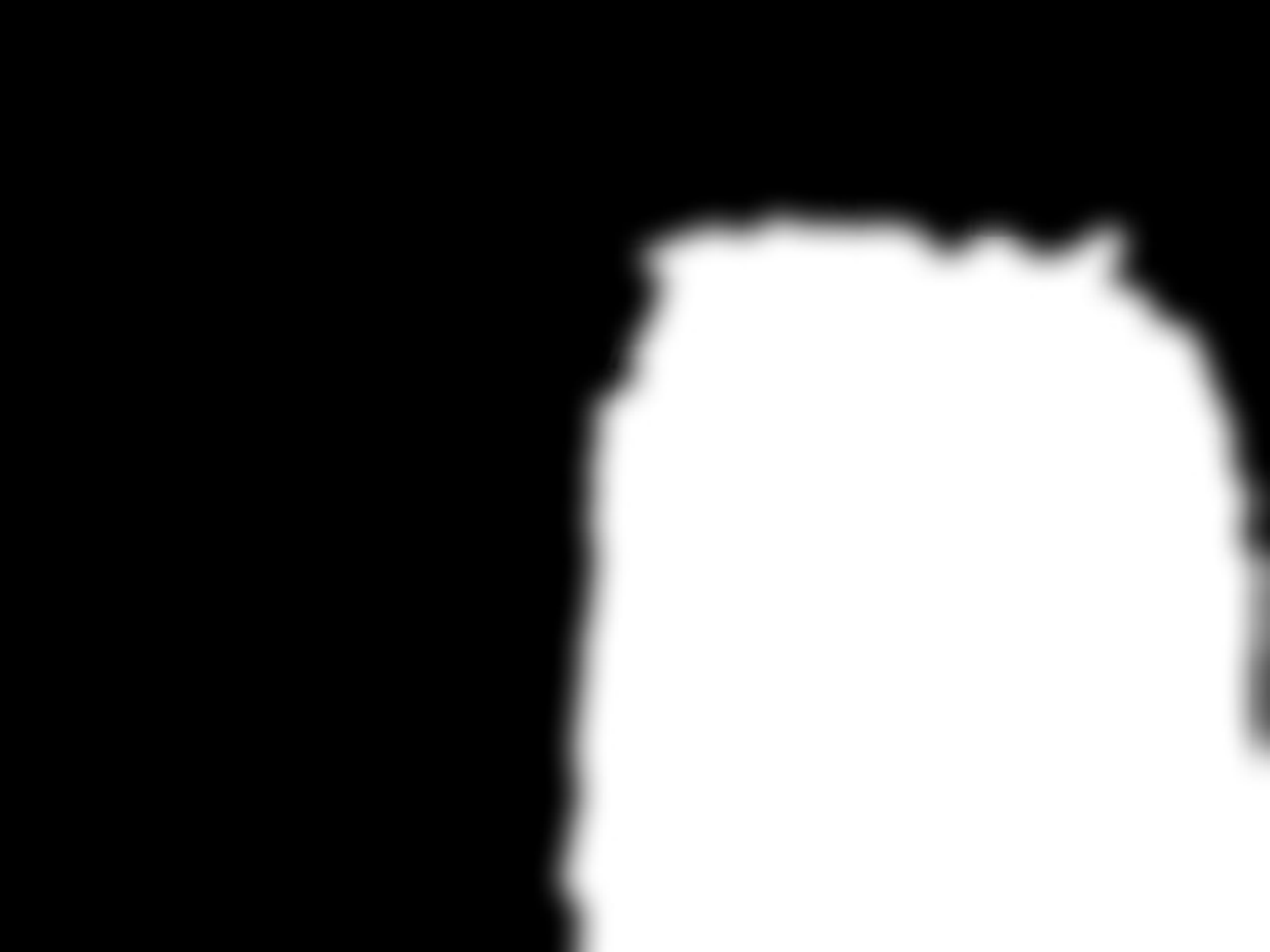}} 
	\subfigure[]{\label{Fig11.10} \includegraphics[width=0.1884\textwidth]{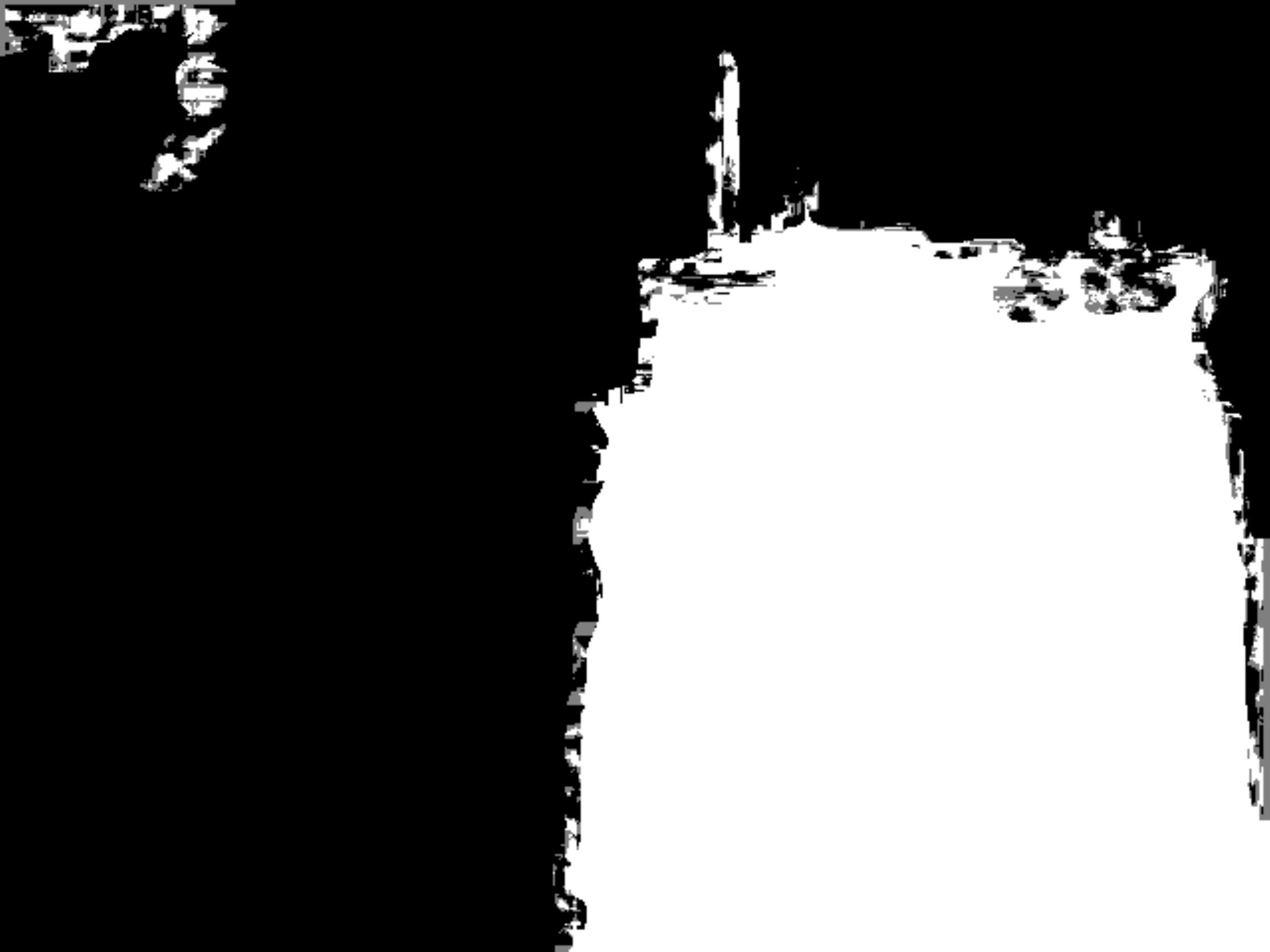}} 
	\subfigure[]{\label{Fig11.11} \includegraphics[width=0.1884\textwidth]{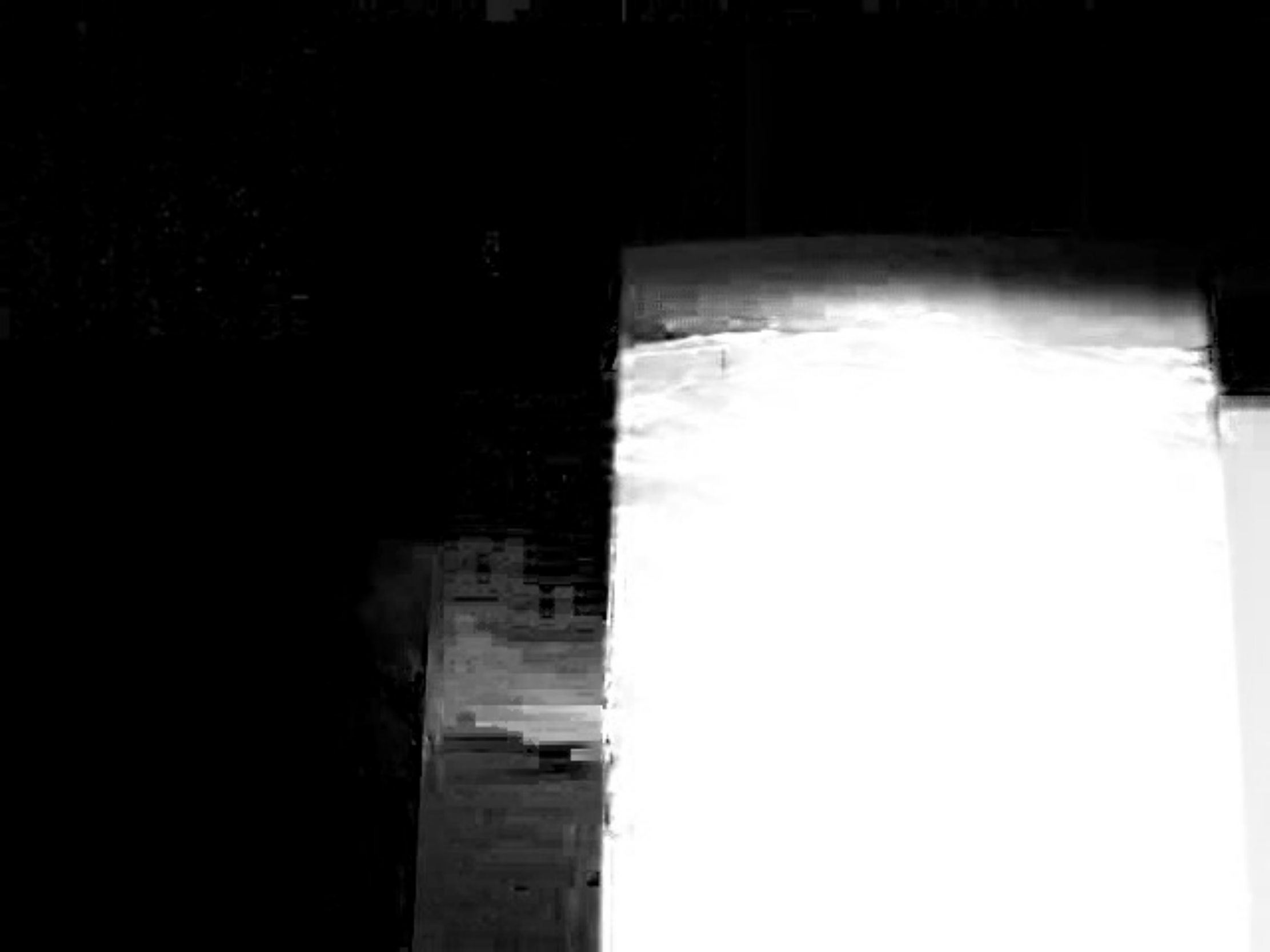}} 
	\subfigure[]{\label{Fig11.12} \includegraphics[width=0.1884\textwidth]{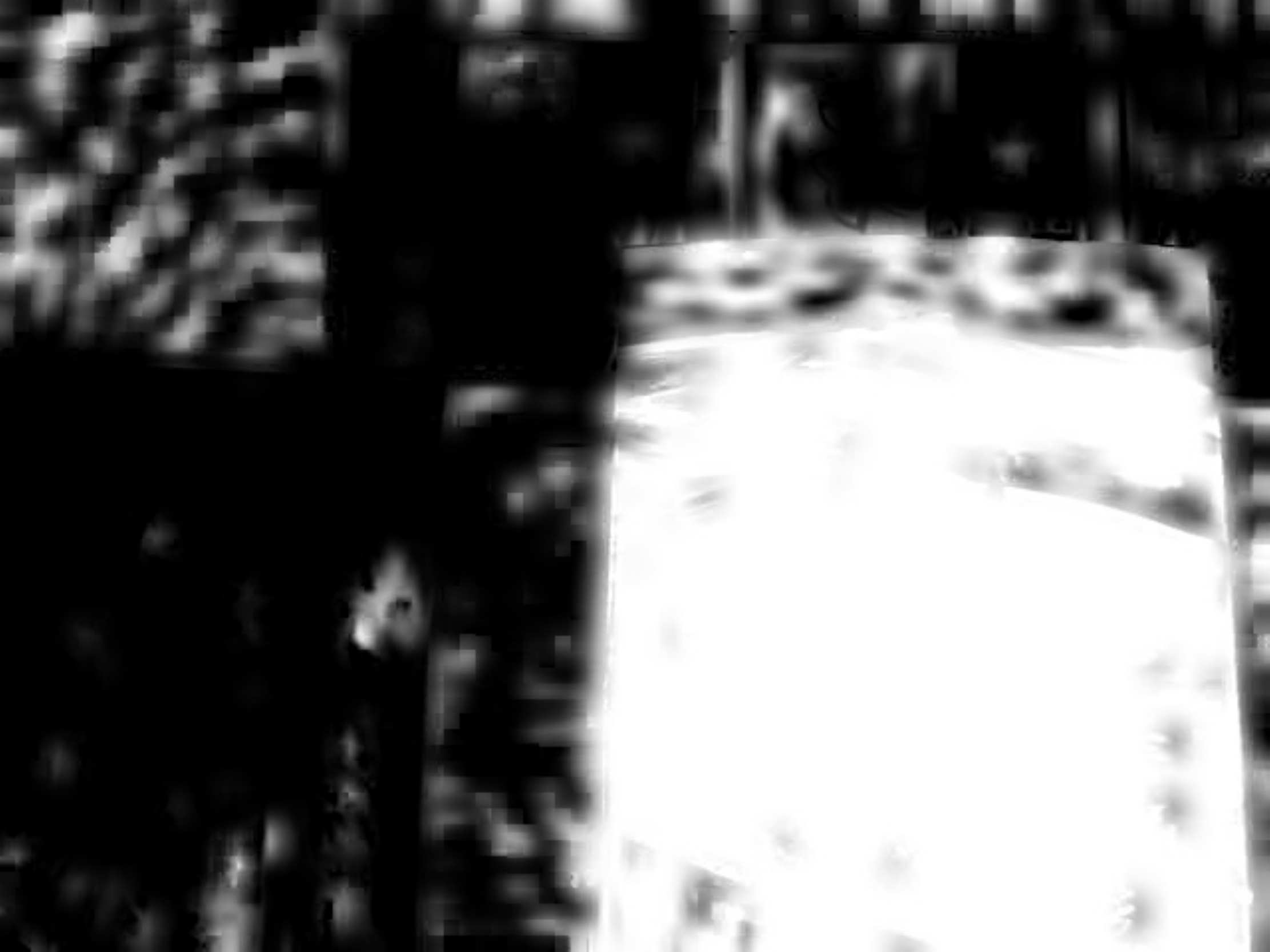}} 
	\subfigure[]{\label{Fig11.13} \includegraphics[width=0.1884\textwidth]{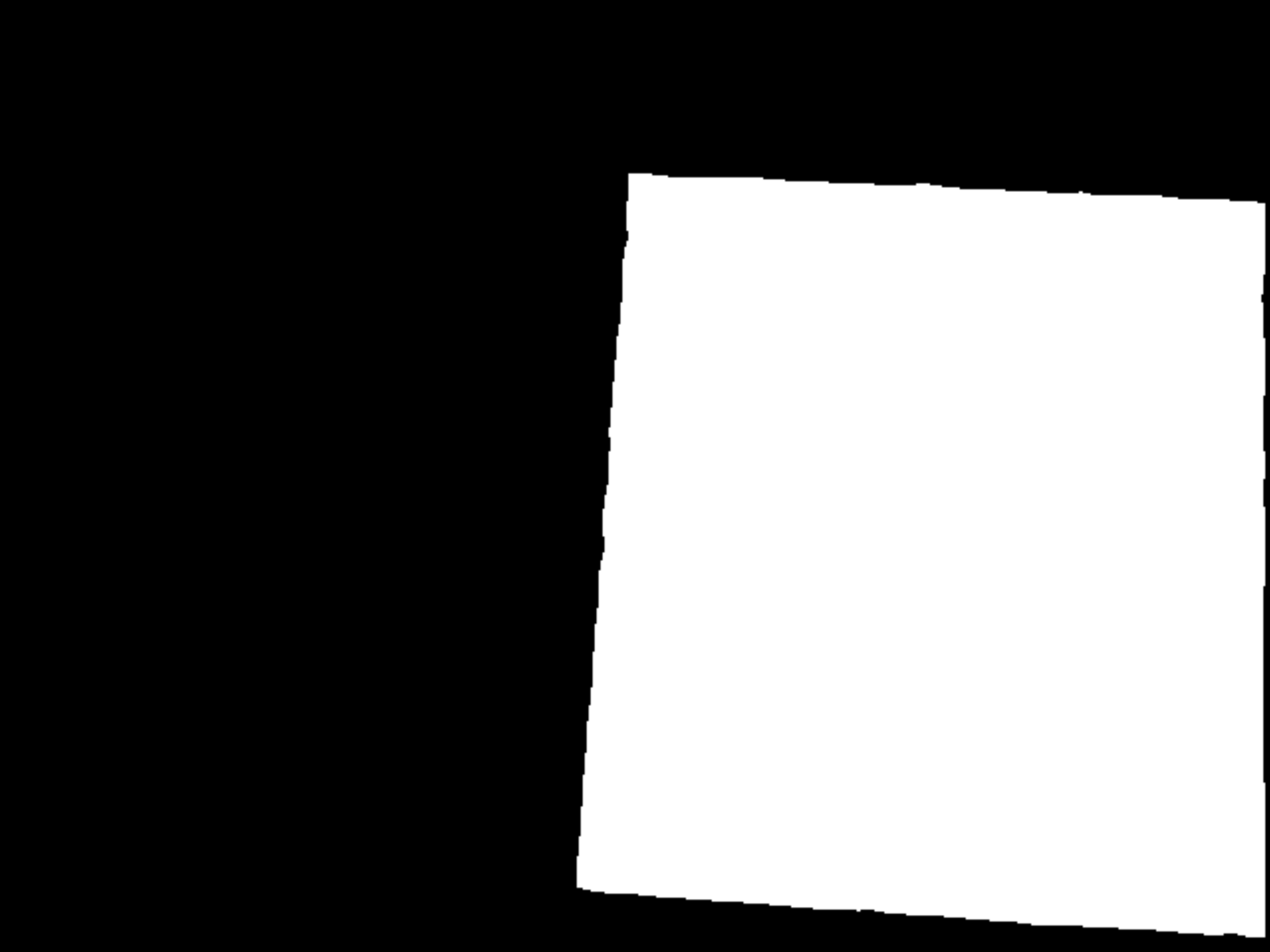}}
	\subfigure[]{\label{Fig11.1} \includegraphics[width=0.1884\textwidth]{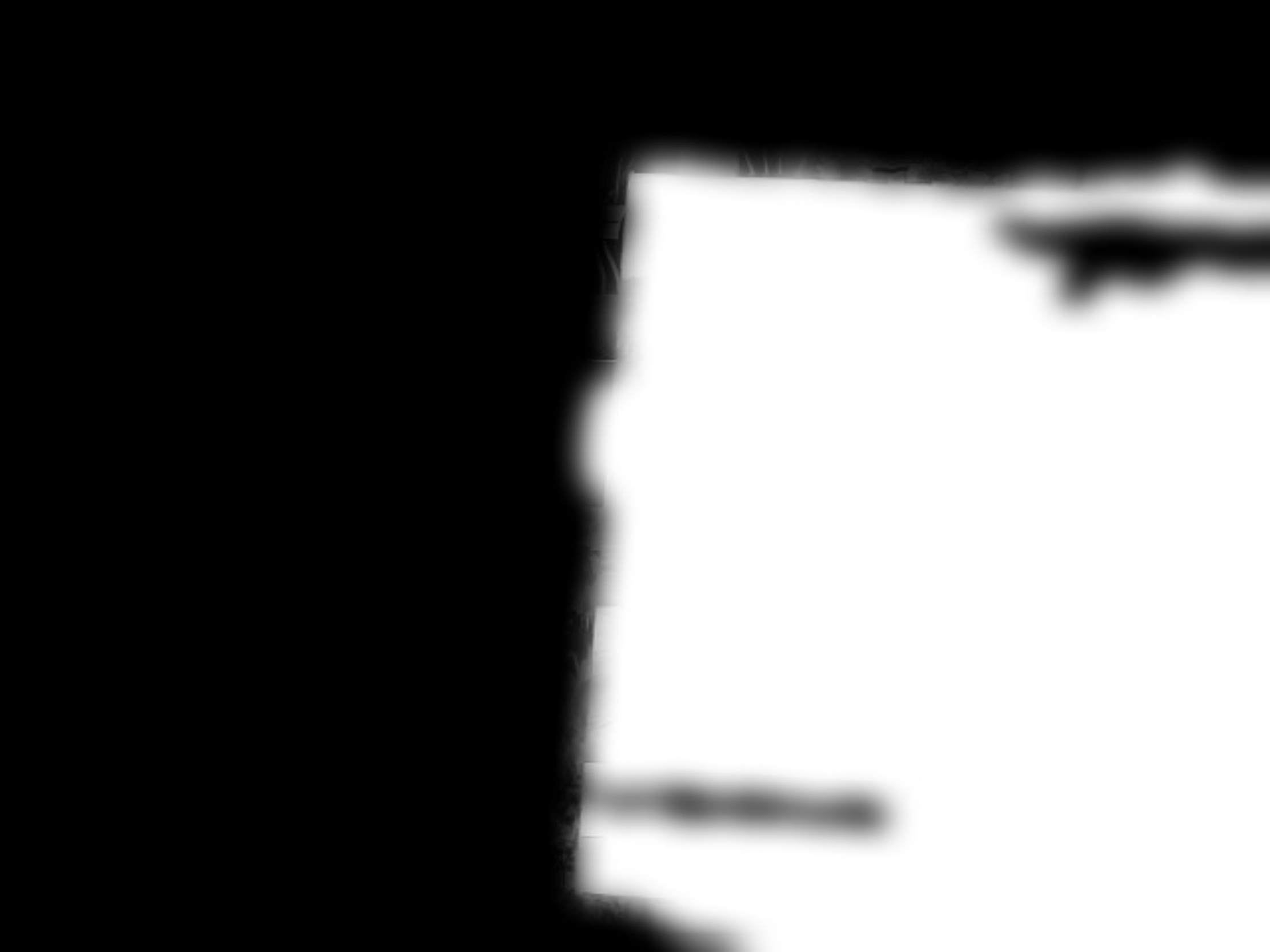}} 
	\subfigure[]{\label{Fig11.14} \includegraphics[width=0.1884\textwidth]{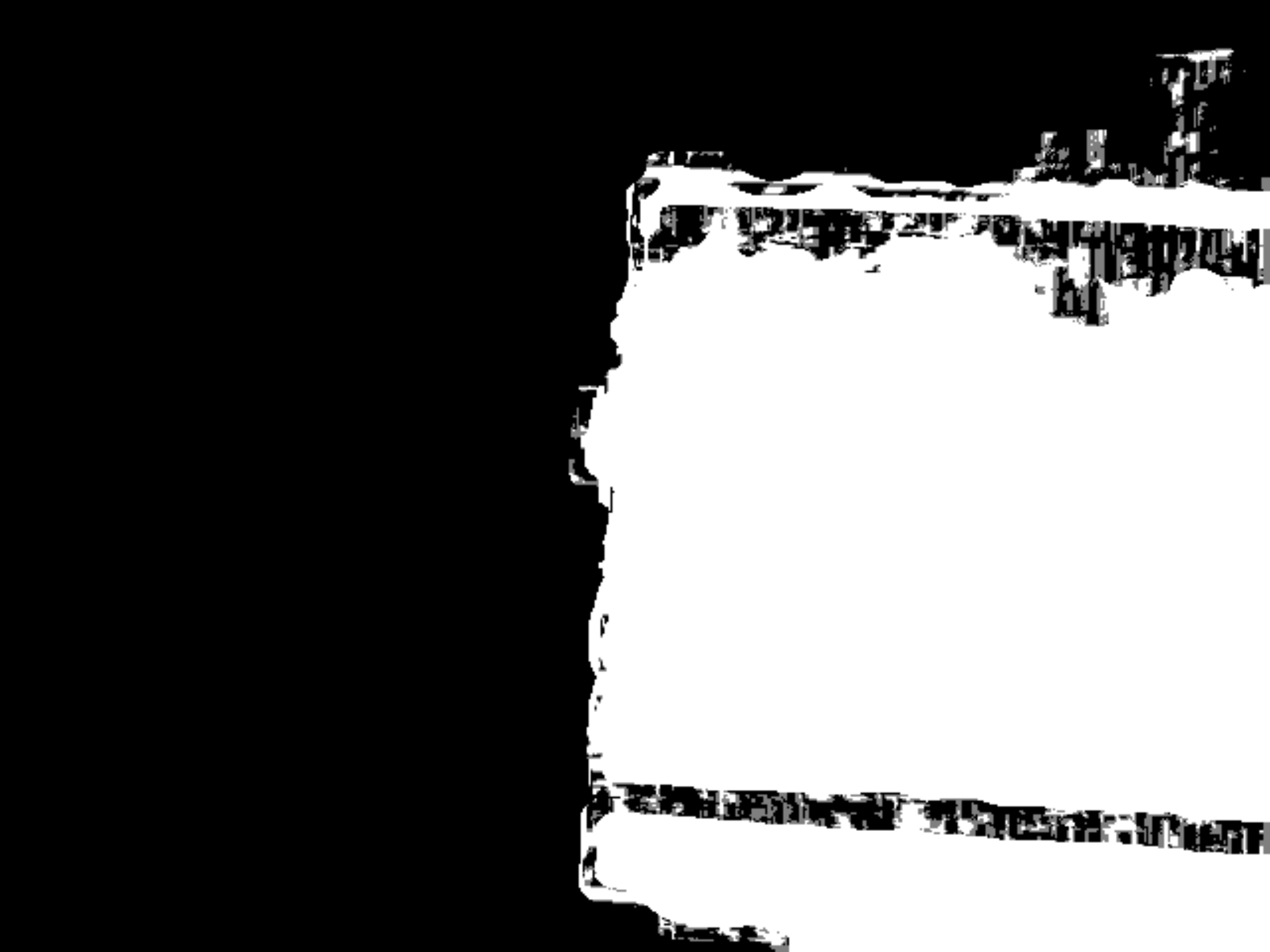}} 
	\subfigure[]{\label{Fig11.15} \includegraphics[width=0.1884\textwidth]{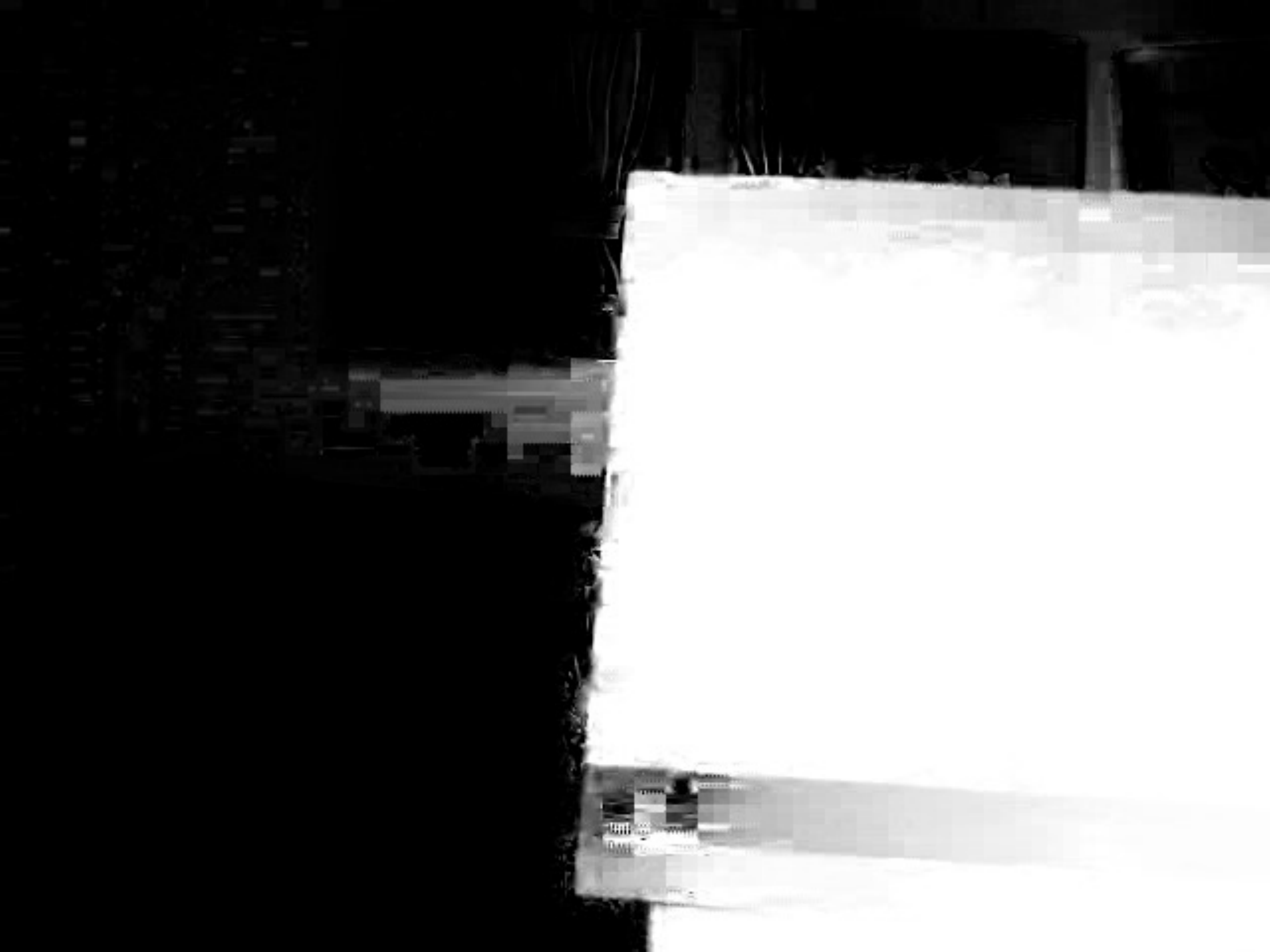}} 
	\subfigure[]{\label{Fig11.16} \includegraphics[width=0.1884\textwidth]{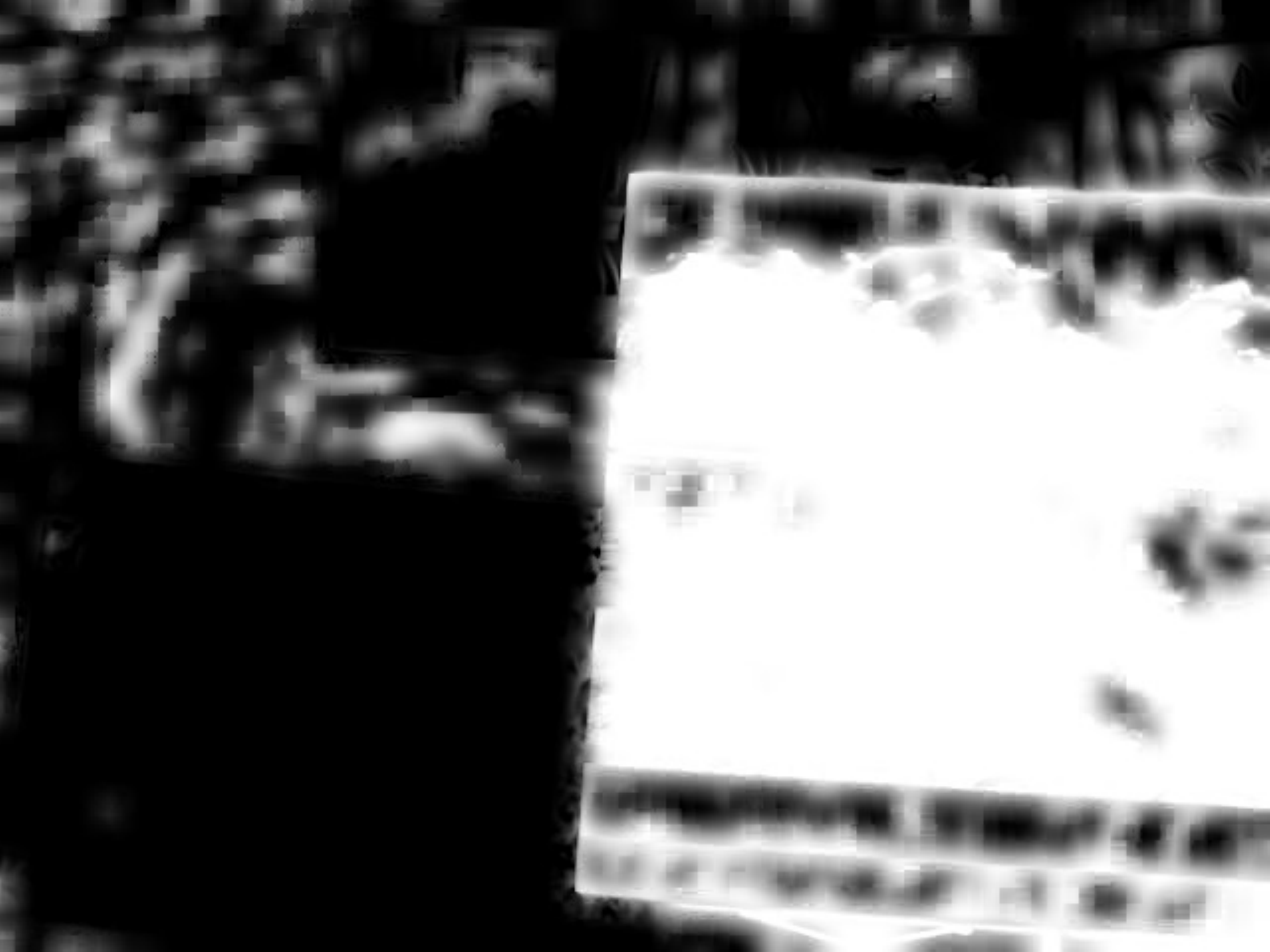}} 
	\subfigure[]{\label{Fig11.13} \includegraphics[width=0.1884\textwidth]{figs3_u-eps-converted-to.pdf}} 
	\subfigure[]{\label{Fig11.1} \includegraphics[width=0.1884\textwidth]{figs3_v-eps-converted-to.pdf}} 
	\subfigure[]{\label{Fig11.14} \includegraphics[width=0.1884\textwidth]{figs3_w-eps-converted-to.pdf}} 
	\subfigure[]{\label{Fig11.15} \includegraphics[width=0.1884\textwidth]{figs3_x-eps-converted-to.pdf}} 
	\subfigure[]{\label{Fig11.16} \includegraphics[width=0.1884\textwidth]{figs3_y-eps-converted-to.pdf}} 
	\renewcommand{\thefigure}{S3}
	\caption{Weight maps generated by different methods when fusing different groups of multi-focus source images in Fig. S1. In each row, the weight map from left to right is generated by the proposed method, DCNN, DSIFT, IM and GF respectively. (a)-(e) belong to the first scene, (f)-(j) belong to the second scene, (k)-(o) belong to the third scene, (p)-(t) belong to the fourth scene, (u)-(y) belong to the fifth scene.
	}
	\label{Fig8}
\end{figure*}

\bibliographystyle{abbrv}
\bibliography{supplebib}